\documentclass[11pt,a4paper,logo]{googledeepmind}

\setrightlogo[180pt]{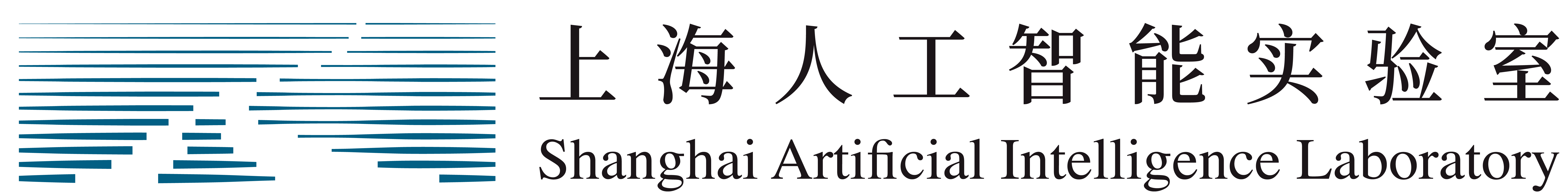}

\usepackage[T1]{fontenc}
\usepackage{pifont}
\usepackage[
    natbib=true,
    backend=biber,
    style=numeric, 
    sorting=none, 
    maxbibnames=99, minbibnames=99 
]{biblatex}
\AtEveryBibitem{\clearfield{month}}
\AtEveryBibitem{\clearfield{day}}
\usepackage{csquotes}
\usepackage{marvosym}

\newcommand{\ProjectName}{Princigram} 
\newcommand{\BenchName}{VeriphyT2IBench} 
\newcommand{\SPCoT}{\mbox{SP-CoT}} 

\title{Towards Physics-Faithful Generation of Scientific Diagrams}

\correspondingauthor{
$\spadesuit$: Co-first Author
\Letter: Corresponding Author. 
$\clubsuit$: Project Leader.
}

\author[1,2 $\spadesuit$]{Minghui Zhang}
\author[1 $\spadesuit$]{Jinxin Shi}
\author[1]{Yifan Chang}
\author[1]{Liangliang Zhao}
\author[1]{Yuandong Pu}
\author[1]{Qian Yu}
\author[1]{Ming Hu}
\author[2]{Hanxiao Zhang}
\author[2]{Yun Gu}
\author[1]{Yirong Chen}
\author[1]{Yu Qiao}
\author[1]{Bo Zhang}
\author[1(\Letter)]{Xiangchao Yan}
\author[1(\Letter, $\clubsuit$)]{Bin Fu}
\author[1(\Letter, $\clubsuit$)]{Yihao Liu}
\affil[1]{Shanghai Artificial Intelligence Laboratory, Shanghai, China}
\affil[2]{Shanghai Jiao Tong University, Shanghai, China}

\usepackage{pdflscape}

\usepackage{textcomp}
\usepackage{rotating}

\usepackage{setspace}
\usepackage{microtype}

\usepackage{graphicx}
\usepackage{subcaption}
\usepackage{caption}

\usepackage{booktabs}
\usepackage[table]{xcolor}

\usepackage{array}
\usepackage{tabularx} 
\usepackage{threeparttable}
\usepackage{multirow}

\usepackage{amsmath}
\usepackage{amssymb}
\usepackage{algorithm}
\usepackage{algpseudocode}
\usepackage{siunitx}

\usepackage{enumitem}
\usepackage{float}
\usepackage{seqsplit}
\usepackage{framed}

\usepackage{tikz}

\usepackage{ragged2e}
\usepackage[most]{tcolorbox}

\definecolor{promptbar}{HTML}{8C8C8C}   
\definecolor{prompttext}{HTML}{1F3864}  
\definecolor{promptcode}{HTML}{7B3F00}  

\newtcolorbox{promptbox}[2][]{%
  enhanced, breakable,
  colback=white, colframe=black, boxrule=0.5pt,
  arc=0pt, outer arc=0pt,
  left=5pt, right=5pt, top=6pt, bottom=6pt,
  colbacktitle=promptbar,
  coltitle=white, fonttitle=\bfseries\sffamily\small,
  title={Prompt}, titlerule=0pt,
  colupper=prompttext,
  before upper={\prompthead{#2}},
  #1
}
\newcommand{\prompthead}[1]{{\color{black}\bfseries #1\par}\smallskip}
\usepackage{fvextra} 
\DefineVerbatimEnvironment{promptjson}{Verbatim}%
  {fontsize=\footnotesize, formatcom=\color{promptcode}, xleftmargin=0pt,
   breaklines=true, breakanywhere=true, breakautoindent=true,
   breaksymbolleft={\textcolor{promptbar}{\ensuremath{\hookrightarrow}}}}

\definecolor{samplebar}{HTML}{1BAF7A}   
\definecolor{sampleeye}{HTML}{7A7B73}   
\newtcolorbox{samplebox}[1]{%
  enhanced, breakable,
  colback=white, colframe=samplebar,
  boxrule=0pt, leftrule=3pt, toprule=0pt, bottomrule=0pt, rightrule=0pt,
  sharp corners,
  left=9pt, right=3pt, top=4pt, bottom=6pt,
  fonttitle=\bfseries\sffamily\small, coltitle=black,
  colbacktitle=white, title={#1}, titlerule=0pt, toptitle=3pt, bottomtitle=1pt
}
\newcommand{\sampleeyebrow}[1]{{\sffamily\scriptsize\color{sampleeye}#1\par}\vspace{2pt}}
\newcommand{\stepline}[2]{{\footnotesize\textbf{\sffamily #1.}\ #2\par}\vspace{1.5pt}}

\definecolor{annstepc}{HTML}{1F3A5F}   
\definecolor{annlabc}{HTML}{B4531F}    
\definecolor{annkeyc}{HTML}{8A8B83}    
\definecolor{annband}{HTML}{ECF1F6}    
\newtcolorbox{anncase}[1]{%
  enhanced, breakable,
  colback=white, colframe=samplebar,
  boxrule=0pt, leftrule=3pt, toprule=0pt, bottomrule=0pt, rightrule=0pt,
  sharp corners, left=8pt, right=4pt, top=3pt, bottom=6pt,
  fonttitle=\bfseries\sffamily\small, coltitle=black,
  colbacktitle=white, title={#1}, titlerule=0pt, toptitle=3pt, bottomtitle=1pt}
\newcommand{\annstep}[2]{\par\smallskip\noindent
  \colorbox{annband}{\parbox{\dimexpr\linewidth-2\fboxsep\relax}{\sffamily\footnotesize
    \textbf{\color{annstepc}#1}\quad{\ttfamily\scriptsize\color{annkeyc}#2}}}\par\smallskip}
\newcommand{\annlab}[1]{{\sffamily\scriptsize\bfseries\color{annlabc}\MakeUppercase{#1}}}
\newlength{\annind}
\newcommand{\annrow}[2]{\par\addvspace{0.8pt}%
  \hangindent=\dimexpr\annind+1.4em\relax\hangafter=1
  \noindent\hspace*{\annind}\annlab{#1}\hspace{0.4em}{\footnotesize #2}\par}
\newcommand{\annline}[1]{\par\addvspace{0.4pt}%
  \hangindent=\dimexpr\annind+1.0em\relax\hangafter=1
  \noindent\hspace*{\annind}{\footnotesize #1}\par}
\newcommand{\annempty}{{\itshape\color{annkeyc}empty}}
\newcommand{\annitem}[1]{\par\addvspace{1.4pt}\noindent\hspace*{\annind}%
  {\sffamily\footnotesize\bfseries\color{annstepc}$\triangleright$~#1}\par}
\newcommand{\annsep}{\;\textcolor{annkeyc}{\textperiodcentered}\;}

\usepackage{makecell}
\usepackage{tabularray}

\usepackage[symbol]{footmisc}
\newcolumntype{Y}{>{\RaggedRight\arraybackslash}X}

\usepackage{hyperref}
\hypersetup{
    colorlinks=true,
    linkcolor=blue, 
    citecolor=blue,  
    filecolor=black,
    urlcolor=blue    
}
\usepackage{cleveref} 

\usepackage{tocloft}

\usepackage{etoolbox}
\usepackage{arydshln}
\makeatletter
\patchcmd{\@tocline}
    {\hfil}
    {\leaders\hbox{\hfil}\hfil}
    {}{}
\makeatother

\begin{abstract}
Text-to-image generation has reached photorealistic quality, yet state-of-the-art systems remain unreliable when asked to produce \emph{scientific diagrams}, whose value depends not on appearance but on \emph{physical faithfulness}: correct force directions, valid coordinate systems, consistent thermodynamic states, and equations that match the depicted scenario. Generic models, trained on web imagery with physically shallow captions, routinely produce diagrams that look plausible but are physically wrong, which is actively harmful in education and scientific communication. Here we present \ProjectName{}, a physics-faithful scientific-diagram generator, together with the data pipeline behind it. Our central advance is \emph{Structured Physical Chain-of-Thought} (\SPCoT{}): a physics-grounded, per-subdiscipline schema that decomposes a physics diagram into an explicit multi-step reasoning chain (scene and object identification, state analysis, force or process analysis, coordinate systems and governing laws, and a synthesis of key relationships) across six physics subdisciplines (mechanics, electromagnetism, physical optics, thermodynamics, acoustics and quantum mechanics). Unlike free-form chain-of-thought, \SPCoT{} follows a fixed schema with strict fidelity rules that separate visually grounded facts from physically inferred reasoning and type all mathematics symbolically, and it is used both as dense training supervision and, at inference, as a structured ``thinking'' prompt the generator conditions on. Using it we curate and structurally annotate a corpus of $4.3$ million physics images spanning the six subdisciplines, within which a high-quality subset of $115{,}037$ images carries expert-level annotation, and on this corpus we adapt a unified multimodal backbone. We further introduce \BenchName{}, a benchmark whose questions are derived from each held-out diagram's own structured annotation rather than authored separately: every diagram is compiled into an item-specific bank of binary questions covering its objects, forces, states and their attributes, which a judge model answers so that a score decomposes into named physical facts rather than a single holistic number. Evaluated on the physics subset of the public GenExam benchmark and on \BenchName{} against a broad suite of open and closed models, \ProjectName{} demonstrates that explicit physics-structured supervision improves the physical faithfulness of generated scientific diagrams.
\end{abstract}

\begin{document}
\sloppy 
\maketitle

\thispagestyle{firststyle}

\section{Introduction}
\label{sec:intro}

Text-to-image (T2I) generation has progressed from producing recognizable objects to synthesizing photorealistic scenes with fine-grained control over style, layout, and text rendering. The trajectory runs from autoregressive token models~\cite{dalle} through denoising diffusion~\cite{ddpm,scoresde,glide} and its latent~\cite{ldm} and transformer-based~\cite{dit} formulations, to cascaded and large-scale systems~\cite{imagen,dalle2,sdxl,pixart,bai2026scaling} and, most recently, to rectified-flow and flow-matching generators~\cite{flowmatching,sd3,flux,qwenimage,hidream}. In parallel, unified multimodal models have begun to couple understanding and generation in a single network, whether by early-fusion token modelling~\cite{chameleon,emu3}, by combining next-token prediction with diffusion~\cite{transfusion,showo,showo2,luminadimoo}, or by decoupling the visual encoders that serve the two tasks~\cite{januspro,bagel}. Contrastive image-text pretraining~\cite{clip} and web-scale corpora~\cite{laion5b} supplied the representations and the data that made this progression possible, and capabilities once considered out of reach, such as rendering legible text inside an image, are now addressed directly~\cite{textdiffuser,anytext}. For the vast majority of natural-image prompts, the dominant failure mode is now subtle: a slightly wrong count, an awkward hand, an imperfect reflection~\cite{geneval,t2icompbench,attendexcite}.

\begin{figure}[!t]
\centering
\includegraphics[width=\linewidth,height=0.9\textheight,keepaspectratio]{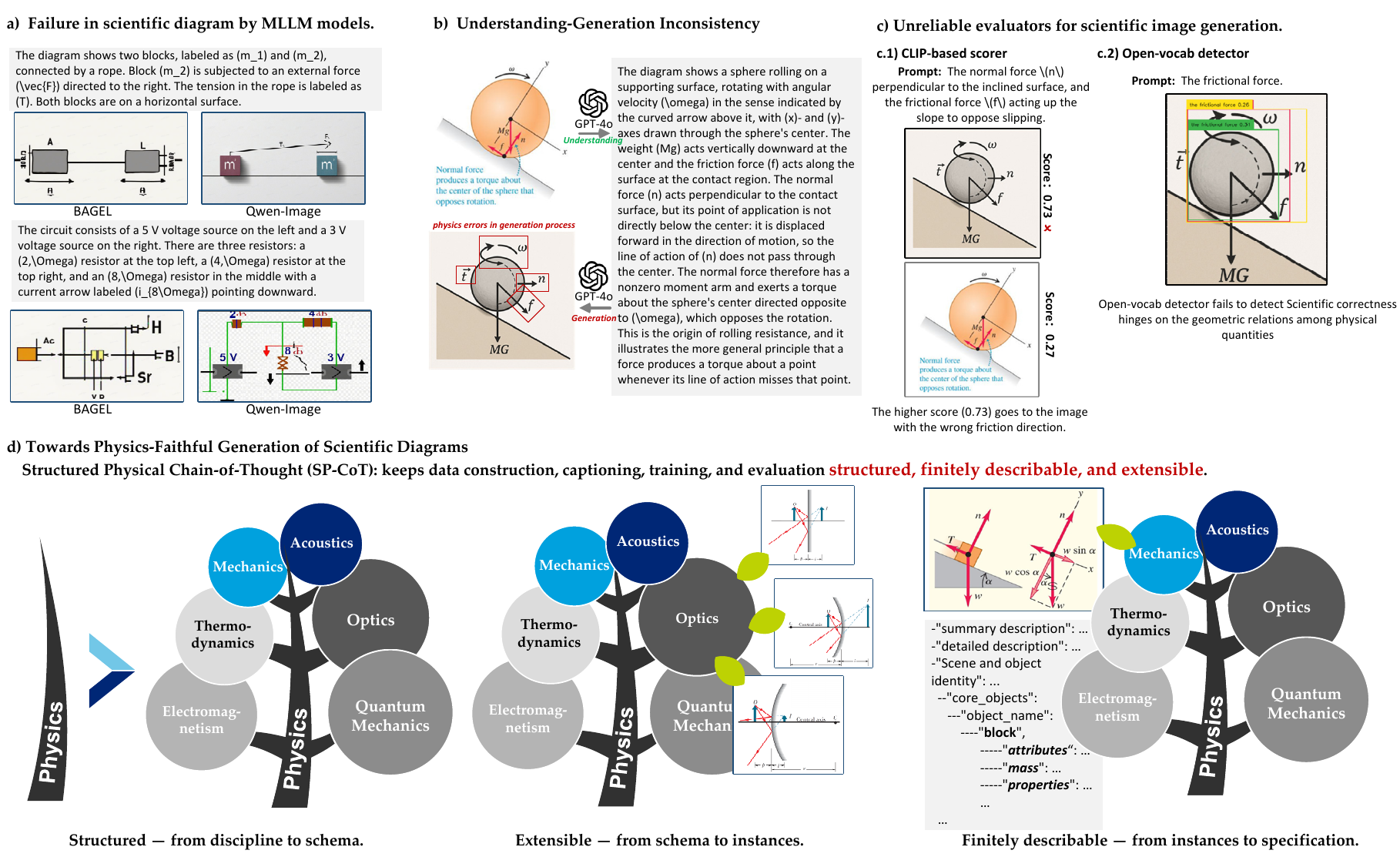}
\caption{
Several challenges to physics-faithful scientific-diagram generation, and our approach. \textbf{(a)}~State-of-the-art multimodal generators produce diagrams that violate basic physics, such as mislabeled forces or incorrect circuit topology. \textbf{(b)}~The failure is generative rather than perceptual: a model that correctly \emph{describes} the physics of a scene still generates it incorrectly. \textbf{(c)}~Appearance-based evaluation is unreliable in this setting, a CLIP-style scorer prefers the physically incorrect image (c.1), and an open-vocabulary detector cannot verify the geometric relations that determine correctness (c.2). \textbf{(d)}~Structured Physical Chain-of-Thought (\SPCoT{}) addresses both problems by keeping data construction, annotation, training, and evaluation structured, finitely describable, and extensible.
}
\label{fig:teaser}
\end{figure}

Scientific diagrams are different in kind. A free-body diagram, a pressure-volume ($PV$) diagram, an optical ray diagram, or a Doppler illustration is not judged by how real it looks but by whether it is \emph{physically correct}. A diagram is wrong if a force points the wrong way or a reaction force is missing, if an isobaric process is not drawn as a horizontal line, if refraction bends light the wrong way at an interface, or if wavefronts fail to compress toward the direction of motion. These are not stylistic preferences but constraints imposed by the underlying physics, and a diagram that violates them is not merely lower quality but misinformation, harmful in the educational and scientific-communication settings where such figures are used. This is a different target from the one current evaluation optimizes for. Prompt-adherence benchmarks ask whether the requested objects, attributes, counts, and spatial relations are present~\cite{geneval,t2icompbench,tifa}, and preference models score how appealing an image is to a human rater~\cite{imagereward}. Neither asks whether the depicted physics is self-consistent. Automatic scorers offer no remedy either: a CLIP-style alignment scorer~\cite{Science-t2i} can rate a physically incorrect diagram as well matched to its prompt, and an open-vocabulary detector~\cite{liu2024grounding} localizes the named objects yet cannot verify the geometric relations (directions, incidences, and orderings) that decide physical correctness (\Cref{fig:teaser}). The scientific-figure literature has so far concentrated on the inverse direction, teaching models to \emph{read} diagrams, charts, and figures rather than to draw them~\cite{ai2d,scicap,chartqa,multimodalarxiv,mathvista}, which leaves the generation of physically faithful diagrams comparatively unexamined.

Two compounding gaps explain the weak performance of generic systems on scientific diagrams. First, a \emph{data gap}: large web-scale corpora~\cite{laion5b} contain relatively few high-quality, physically accurate diagrams, and their alt-text captions are short and physically shallow (``a diagram of a pendulum''), carrying none of the constraints (forces, directions, governing equations) that make the diagram correct. Second, a \emph{supervision gap}: even when accurate diagrams are present, standard caption supervision never encodes \emph{why} the diagram looks the way it does, so a model learns the surface appearance of physics figures without learning the rules that generate them. That caption quality, and not only image quality, governs what a generator learns is by now well established: replacing web alt-text with dense synthetic descriptions is what drove much of the recent gain in prompt adherence~\cite{dalle3,pixart,recapdatacomp}. Those descriptions, however, remain descriptions of \emph{appearance}. What a physics diagram needs is supervision on the reasoning behind the appearance, in the spirit of the explicit intermediate steps that unlocked reasoning in language models~\cite{cot}, but tied to the marks actually drawn on the page. Recent multidisciplinary T2I benchmarks make this concrete: on the physics portion of GenExam~\cite{genexam}, even strong closed models score in the single digits under strict, fine-grained scoring, and open models are near zero.

We argue that closing the supervision gap is the key lever, and that it can be achieved with data rather than architectural change. Here we present \ProjectName{}, a physics-faithful scientific-diagram generator, and the data pipeline behind it. The name states the goal: to render the figure that a set of physical principles \emph{entails}, rather than one that merely resembles figures of its kind. The core of our approach is \emph{Structured Physical Chain-of-Thought} (\SPCoT{}) (\Cref{sec:res-framework}): for each physics subdiscipline we define a JSON schema that lays out the physical reasoning behind a diagram as an explicit chain of steps: identifying objects and attributes, characterizing the system state, performing the force or process analysis, declaring the coordinate system and governing laws, and synthesizing the key physical relationship. In the spirit of the intermediate reasoning steps that unlocked language-model reasoning~\cite{cot}, but constrained to a fixed physics schema rather than free-form text, this chain is used two ways: as dense supervision when a diagram is annotated with its populated schema, and, at inference, as a structured ``thinking'' prompt that a language model fills in before the generator draws. Crucially, the schema enforces strict fidelity rules that distinguish what must be read directly off the image (objects, drawn arrows, labeled states) from what may be inferred by physical reasoning (motion regimes, governing equations, idealizing assumptions), and types all mathematics symbolically. This turns a flat image-caption pair into an image paired with an auditable chain of physical reasoning.

Using this framework, we assemble a large-scale corpus of physics imagery ($4.3$~million images), structurally annotate all of it across mechanics, electromagnetism, physical optics, thermodynamics, acoustics, and quantum mechanics, and raise a high-quality subset of $115{,}037$ image--annotation pairs to expert-level annotation (\Cref{sec:res-data}). We adapt unified multimodal backbones~\cite{bagel,luminadimoo} to physics-faithful generation by training on these structured annotations, so that the same model that reasons over the physics also renders the diagram (\Cref{sec:res-model}). We report results on two such backbones to show that the gains follow the supervision rather than a particular architecture. To measure physical faithfulness at the level of individual physical facts, we build \BenchName{} (\Cref{sec:res-inhouse}), a benchmark of held-out annotated diagrams in which each item's structured analysis is compiled into an item-specific bank of binary questions about its objects, forces, states and their attributes. A judge model answers these questions about each generation, so the score decomposes per attribute rather than collapsing to a single holistic number. On the physics subset of GenExam and on \BenchName{}, we compare \ProjectName{} against a broad suite of open and closed models (\Cref{sec:res-genexam,sec:res-inhouse}), and introduce an automatic, structured-key-value evaluator that reuses the annotation schema to score physical correctness at scale (\Cref{sec:res-autoeval}). Full data, model, training, and evaluation details are provided in the Methods (Supplementary Information, \Cref{sec:methods}).

\section{Results}
\label{sec:results}

We organize the results around the four components of the work: Structured Physical Chain-of-Thought (\SPCoT{}, \Cref{sec:res-framework}), the corpus built with it (\Cref{sec:res-data}), the \ProjectName{} generator trained on it (\Cref{sec:res-model}), and the evaluation of physical faithfulness on public and in-house benchmarks (\Cref{sec:res-genexam,sec:res-inhouse,sec:res-autoeval}). Full procedural detail for every component is deferred to the Methods (\Cref{sec:methods}).

\subsection{Structured Physical Chain-of-Thought (\SPCoT{})}
\label{sec:res-framework}

The central result of this work is a representation rather than a network: \emph{Structured Physical Chain-of-Thought} (\SPCoT{}), a single schema that makes the physics of a diagram explicit and auditable. A flat caption such as ``a block on an inclined plane with a spring'' records \emph{what} appears but not \emph{why} it is arranged that way. The physics (the direction of the normal force, the decomposition of gravity along and perpendicular to the incline, the choice of coordinate axes) is exactly what distinguishes a correct free-body diagram from a wrong one, and exactly what flat captions omit. We therefore replace flat captions with an \SPCoT{}: a JSON object that lays out the diagram's physical content as an explicit, multi-step reasoning chain. Each populated schema, the \emph{structured annotation} of a diagram, is one instance of this chain of thought.

Although each subdiscipline has its own vocabulary, the reasoning structure of a physics diagram is shared. We exploit this to define a single five-step template, instantiated per subdiscipline (\Cref{tab:schema}): (1)~\emph{Scenario}, scene and object/system identification; (2)~\emph{Parameters}, state or parameterization; (3)~\emph{Structure}, interaction or structural analysis (force analysis in mechanics, process and energy-transfer analysis in thermodynamics, geometry-and-wave analysis in acoustics, circuit topology or field distribution in electromagnetism, optical-path and phase analysis in physical optics, and boundary conditions or basis states in quantum mechanics); (4)~\emph{Laws}, coordinate system and governing laws; and (5)~\emph{Synthesis} of the key physical relationship and idealizing assumptions.

The defining feature of the framework is that it explicitly separates what can be \emph{seen} from what must be \emph{reasoned}. Each subdiscipline designates exactly two steps as strictly faithful to the image, chosen according to where its physics is visually carried (\Cref{tab:schema}): in mechanics, object identification (Step~1) and force analysis (Step~3); in thermodynamics, state identification (Step~2) and the visualized process path (Step~3); in acoustics, scene identification (Step~1) and the system components (Step~2); and in electromagnetism, physical optics, and quantum mechanics, component or source identification (Step~2) together with the structural layout of Step~3 (circuit topology or field distribution, the optical path, and the spatial regions or basis states, respectively). For these steps the annotator may record only elements that are explicitly drawn, and may not invent any object, force, or state that is not visible. The remaining steps may be inferred by physical reasoning when they can be logically deduced from the drawn elements. When information is missing it must be encoded as an empty string or list rather than guessed. All mathematics is required to be valid \LaTeX{}, giving a consistent, renderable, machine-parseable representation that the automatic evaluator of \Cref{sec:res-autoeval} reuses directly. This visible/inferred split is what makes the annotations trustworthy: grounded fields can be checked against the pixels, while inferred fields capture the physics the diagram is meant to teach. We state the schema precisely in \Cref{sec:formal-annotation}: each field $f=(\kappa_f,\nu_f)$ carries a type $\operatorname{type}(f)\in\{\textsc{ent},\textsc{rel},\textsc{val}\}$ (entity, relation, or value) and a grounding $g(f)\in\{\mathsf{vis},\mathsf{inf}\}$, and a valid annotation of an image $x$ obeys the fidelity rule
\begin{equation}
  g(f)=\mathsf{vis}\ \Longrightarrow\ \nu_f\in\mathrm{drawn}(x),
  \label{eq:m-fidelity}
\end{equation}
so every visible field must name an element actually drawn. The complete per-subdiscipline templates are given in \Cref{sec:methods-templates}.

\begin{figure}[!t]
\centering
\includegraphics[width=0.95\linewidth]{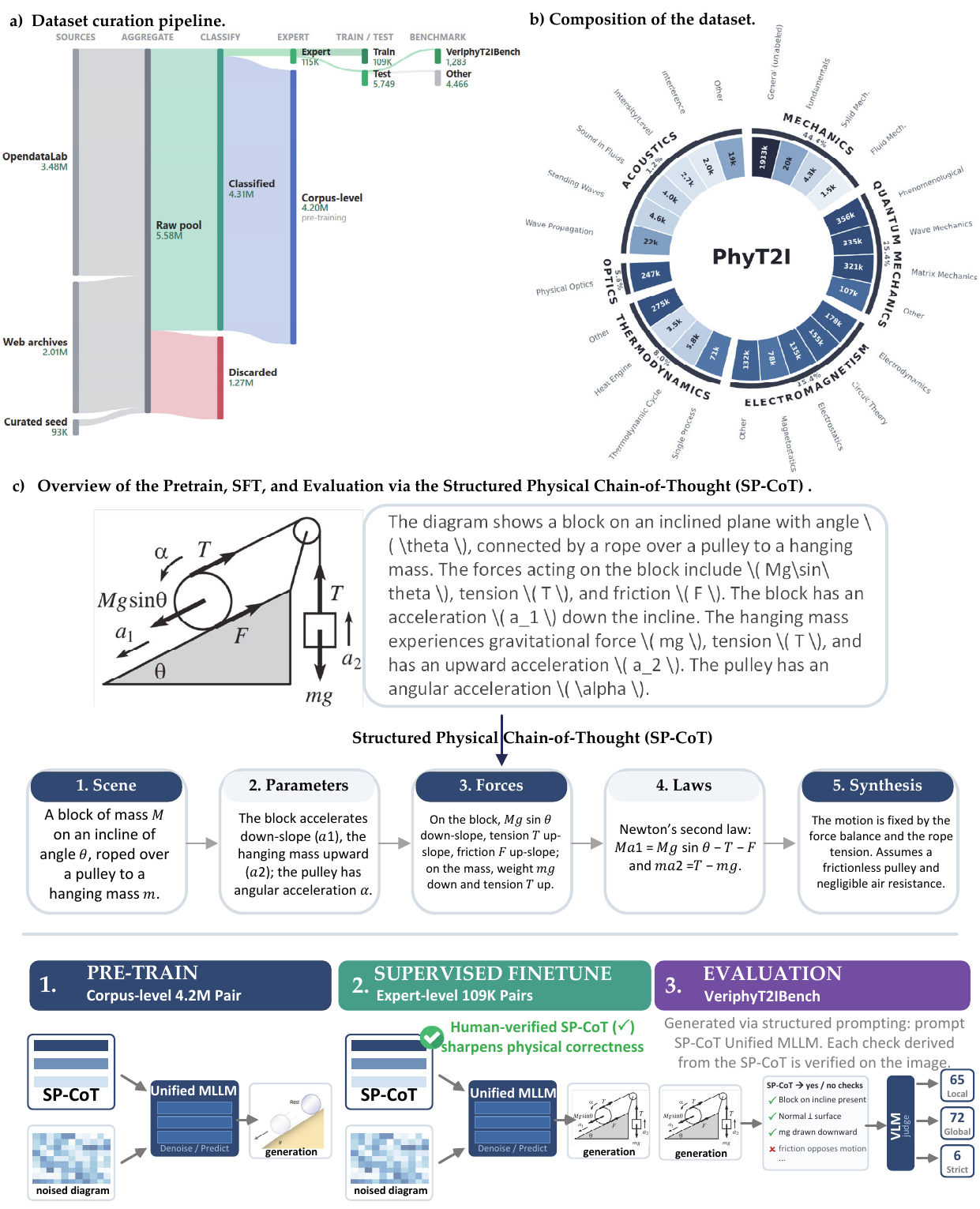}
\caption{Overview of \ProjectName{}: data construction, structured supervision, and training--evaluation.
}
\label{fig:overview}
\end{figure}

\begin{figure}[!t]
\centering
\includegraphics[width=1.0\linewidth]{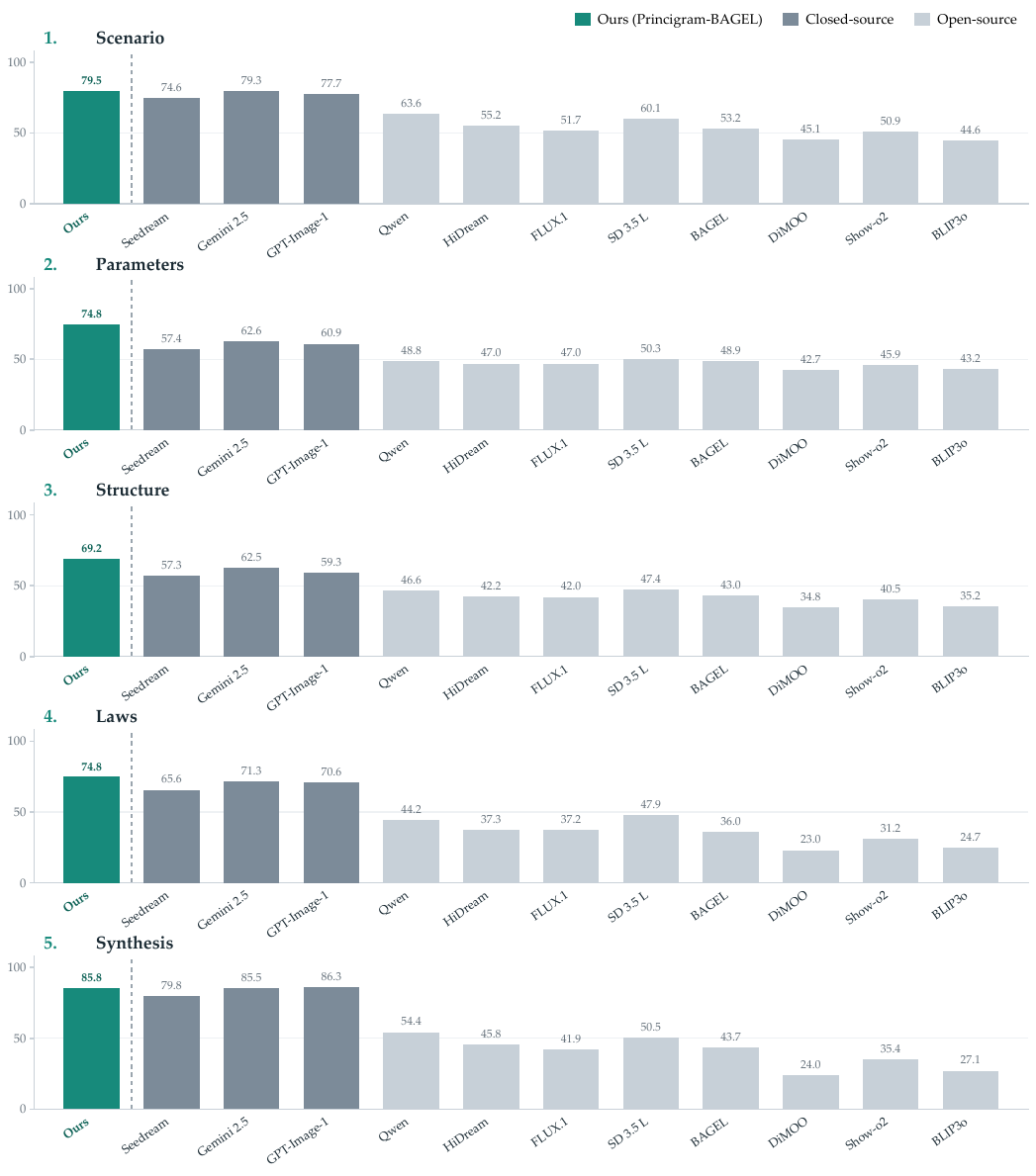}
\caption{Per-step faithfulness across the five \SPCoT{} schema steps (Scenario, Parameters, Structure, Laws, Synthesis). \ProjectName{}-BAGEL (teal) is compared against closed- and open-source models.}
\label{fig:step-scores}
\end{figure}

\begin{table}[t]
\centering
\caption{The unified five-step schema, instantiated across the six physics subdisciplines. The column headings abbreviate the five steps: \emph{Scenario}, identification of the scene, system, or phenomenon; \emph{Parameters}, the state or parameterization; \emph{Structure}, the interaction or structural analysis; \emph{Laws}, the coordinate system and governing laws; and \emph{Synthesis}. Cells marked $^\dagger$ are constrained to be \emph{strictly faithful to the image}: only elements explicitly drawn may be recorded there. The unmarked cells may be physically inferred from the drawn elements. Every subdiscipline grounds exactly two steps, but \emph{which} two depends on where its physics is visually carried: mechanics grounds the drawn force arrows, thermodynamics the plotted states and process path, and the field subdisciplines the components and the structural or geometric layout that relates them.}
\label{tab:schema}
\small
\setlength{\tabcolsep}{4pt}
\begin{tabularx}{\linewidth}{l Y Y Y Y Y}
\toprule
\textbf{Subdiscipline} & \textbf{1. Scenario} & \textbf{2. Parameters} & \textbf{3. Structure} & \textbf{4. Laws} & \textbf{5. Synthesis} \\
\midrule
Mechanics & Objects \& attributes, constraints$^\dagger$ & Motion state, kinematic variables & Force analysis$^\dagger$ & Coordinate system, Newton's laws & Key relationship, assumptions \\
\addlinespace[2pt]
Electromagnetism & Sub-domain, problem type & Components \& sources, parameters$^\dagger$ & Circuit topology / field distribution, state$^\dagger$ & Conventions, Kirchhoff / Maxwell laws & Key relationship, assumptions \\
\addlinespace[2pt]
Physical optics & Phenomenon type (interference, \dots) & Components \& parameters ($\lambda,d,a,L$)$^\dagger$ & Optical path, path \& phase difference$^\dagger$ & Principles, condition equations & Key relationship, assumptions \\
\addlinespace[2pt]
Thermodynamics & System definition, problem type & Thermodynamic states $P,V,T$$^\dagger$ & Process path \& energy transfer$^\dagger$ & Governing laws, equations & Key findings, assumptions \\
\addlinespace[2pt]
Acoustics & Scene \& phenomenon (e.g.\ Doppler)$^\dagger$ & Components \& parameters ($f_s,v_o,c$)$^\dagger$ & Geometry \& wave analysis & Conventions, principles, equations & Key relationship, assumptions \\
\addlinespace[2pt]
Quantum mechanics & Formalism, scenario, degrees of freedom & Hamiltonian / landscape, parameters$^\dagger$ & Boundary conditions / basis \& eigenstates$^\dagger$ & Evolution \& measurement equations & Key phenomena, assumptions \\
\bottomrule
\end{tabularx}
\end{table}

\subsection{A large-scale corpus of physics diagrams}
\label{sec:res-data}

We scope the first version of \ProjectName{} to physics, organized into six subdisciplines: mechanics, electromagnetism, physical optics, thermodynamics, acoustics, and quantum mechanics. Physics is an ideal first domain: its diagrams are governed by a small set of well-understood laws, which makes ``faithfulness'' precisely definable and gives a clear target for both annotation and evaluation.

We combine four complementary sources, trading scale against precision: a manually curated gold seed of high-quality, unambiguous diagrams selected against an explicit source-selection principle; large-scale mining from OpenDataLab~\cite{opendatalab}; targeted crawling of physics references from public web archives; and a procurement pipeline for high-resolution original English-language physics textbooks (high-school through graduate level), governed by a formal requirements-and-acceptance specification covering format, scope, and language. After de-duplication, resolution filtering, and a subdiscipline classifier that routes each image to one of the six domains and discards non-physics content, these sources yield roughly $4.3$~million physics images, every one of which is structurally annotated with the framework of \Cref{sec:res-framework} (\Cref{fig:overview}a, \Cref{tab:data}). The corpus is distributed very unevenly across subdisciplines, reflecting how often each is drawn: mechanics and quantum mechanics together account for more than two thirds of it, whereas optics is comparatively rare (\Cref{fig:overview}b, Methods, \Cref{sec:methods-data}).

We annotate at two levels, because the two uses of the data have different quality requirements. Annotation at corpus scale gives the breadth needed to associate physics-structured text with faithful pixels across every subdiscipline. On top of it, a high-quality subset of $115{,}037$ images receives \emph{expert-level} annotation, which is what supervision on physical correctness, and any evaluation of it, must rest on (\Cref{tab:data}). This second tier deliberately re-weights the corpus toward the subdisciplines with the richest diagrammatic conventions: electromagnetism and mechanics, rather than the corpus leaders, contribute the most expert-annotated pairs, and optics, the rarest subdiscipline in the corpus, is lifted from $1\%$ of the corpus to $7\%$ of the expert tier. Within the expert tier we hold out a test split of $5{,}749$ pairs and use the remaining $109{,}288$ for training. A balanced $1{,}283$-item subset of the held-out split forms \BenchName{}, the benchmark evaluated in \Cref{sec:res-inhouse}.

\begin{table}[t]
\centering
\caption{\textbf{Corpus and annotation statistics per subdiscipline.} The corpus is annotated at two levels. \emph{Full corpus} counts every image retained after de-duplication, resolution filtering, and subdiscipline classification. All of these carry a structured annotation produced with the framework of \Cref{sec:res-framework}. \emph{Expert-level} counts the subset whose annotations are additionally expert-grade, and which is partitioned into a training set (\emph{Train}) and a held-out \emph{Test} set. \BenchName{} is a balanced subset of \emph{Test} used as the evaluation benchmark (\Cref{sec:res-inhouse}). Counts are numbers of images.}
\label{tab:data}
\small
\setlength{\tabcolsep}{4pt}
\begin{tabular}{lrrrrr}
\toprule
& \textbf{Full corpus} & \multicolumn{4}{c}{\textbf{Expert-level subset}} \\
\cmidrule(lr){2-2}\cmidrule(lr){3-6}
\textbf{Subdiscipline} & \textbf{Structured} & \textbf{Total} & \textbf{Train} & \textbf{Test} & \textbf{\BenchName{}} \\
\midrule
Mechanics         & 1{,}938{,}649 & 27{,}583 & 26{,}204 & 1{,}379 & 300 \\
Electromagnetism  &   240{,}025 & 41{,}927 & 39{,}831 & 2{,}096 & 173 \\
Physical optics   &    48{,}282 &  8{,}419 &  7{,}999 &   420 & 168 \\
Thermodynamics    &   346{,}904 &  9{,}026 &  8{,}575 &   451 & 163 \\
Acoustics         &   641{,}477 &  6{,}489 &  6{,}165 &   324 & 320 \\
Quantum mechanics & 1{,}098{,}529 & 21{,}593 & 20{,}514 & 1{,}079 & 159 \\
\midrule
\textbf{Total}    & \textbf{4{,}313{,}866} & \textbf{115{,}037} & \textbf{109{,}288} & \textbf{5{,}749} & \textbf{1{,}283} \\
\bottomrule
\end{tabular}
\end{table}

\subsection{\ProjectName{} generates physics-faithful diagrams}
\label{sec:res-model}

\ProjectName{} is built on a unified multimodal backbone~\cite{bagel} that handles both multimodal understanding and image generation within a single model, so that text and image tokens are processed jointly. We adopt this unified design because structured-annotation supervision is most effective when the same model that reasons over the physics also renders the diagram. The model is trained in two stages, which draw on the two annotation levels of \Cref{tab:data}: pre-training on (structured-annotation, image) pairs from the full corpus, to associate physics-structured text with faithful pixels at breadth, and supervised fine-tuning on the expert-level subset, to align the model with natural-language use and sharpen physical correctness. Both stages optimize the same subdiscipline-balanced objective on the serialized annotation, and the backbone enters only through its generative loss (a rectified-flow loss for BAGEL, a masked-diffusion loss for Lumina-DiMOO~\cite{luminadimoo}), which is precisely why the same supervision transfers across backbones (\Cref{sec:formal-annotation}). At inference time we close the gap between short user prompts and the structured conditioning the model was trained on with an \SPCoT{} (``thinking'') pipeline: a language model first expands the raw prompt into the subdiscipline schema, and \ProjectName{} then generates conditioned on the populated schema. Architecture, training recipe, and the inference pipeline are detailed in \Cref{sec:methods-model,sec:methods-training,sec:methods-extension}.

Qualitatively, \ProjectName{} produces diagrams whose drawn elements are consistent with the stated physics (forces with correct directions and sources, process paths with the correct shape on a $PV$ diagram, and wavefronts consistent with the depicted phenomenon), where generic baselines produce plausible-looking but physically inconsistent figures. Side-by-side comparisons against a broad suite of open and closed generators are shown per subdiscipline in \Cref{sec:qual-analysis}, and an ablation that isolates the structured supervision in \Cref{sec:qual-ablation}.

\subsection{Benchmarking physical faithfulness on GenExam}
\label{sec:res-genexam}

We evaluate on the physics subset of GenExam~\cite{genexam}, a multidisciplinary T2I ``exam'' benchmark that pairs each prompt with a ground-truth reference and fine-grained scoring points and uses an MLLM judge to produce strict and relaxed scores. The subset is organized by its own subject taxonomy: circuits, electromagnetism, mechanics, optics, quantum mechanics, and thermodynamics. We compare against a broad suite of open and closed models, and report \ProjectName{} on two backbones, BAGEL and Lumina-DiMOO (which we abbreviate DiMOO). On the full GenExam benchmark, open models score near zero under strict scoring and only modestly under relaxed scoring~\cite{genexam}, which sets the difficulty of the physics subset in context. \Cref{tab:results-genexam-subjects} reports per-subject and aggregate scores under the structured-prompt protocol (Methods, \Cref{sec:methods-eval}).

\begin{table}[t]
\centering
\caption{
\textbf{GenExam Physics} subjects results.
Per-subject columns report the relaxed score (\%) for each of the six
physics topics: Circuits (circuit analysis),
E\&M (electromagnetism),
Mech.\ (mechanics),
Optics,
QM (quantum mechanics), and
Thermo.\ (thermodynamics).
Strict and Relaxed are the two aggregate scoring
modes of the GenExam MLLM judge: strict requires all scoring points
to be satisfied, while relaxed awards partial credit. All scores are percentages, higher is better.
Per column, the best value is in \textbf{bold} and the second best is \underline{underlined}.}
\label{tab:results-genexam-subjects}
\small
\begin{tabular}{lcccccc cc}
\toprule
& \multicolumn{6}{c}{\textbf{Per-subject (relaxed)}} & \multicolumn{2}{c}{\textbf{Overall}} \\
\cmidrule(lr){2-7}\cmidrule(lr){8-9}
\textbf{Model} & Circuits & E\&M & Mech. & Optics & QM & Thermo. & Strict & Relaxed \\
\midrule
Seedream 4.0~\cite{seedream}             & \underline{36.8} & \underline{60.0} & 47.4 & 66.2 & \textbf{60.1} & \textbf{50.6} & 3.5 & 49.0 \\
FLUX.1 Kontext max~\cite{flux}           & 14.5 & 29.3 & 33.1 & 38.3 & 35.7 & 17.7 & 0.0 & 25.6 \\
Qwen-Image~\cite{qwenimage}              & 14.7 & 35.1 & 37.7 & 28.8 & 28.6 & 22.9 & 0.0 & 26.3 \\
HiDream-I1-Full~\cite{hidream}           & 8.8  & 26.0 & 23.6 & 24.8 & 25.1 & 8.2  & 0.0 & 17.7 \\
FLUX.1 dev~\cite{flux}                   & 9.7  & 19.1 & 11.4 & 36.0 & 20.7 & 14.0 & 0.0 & 14.4 \\
SD 3.5 Large~\cite{sd3}                  & 5.0  & 19.1 & 16.8 & 21.5 & 26.9 & 3.4  & 0.0 & 13.2 \\
BAGEL~\cite{bagel}            & 7.2  & 24.6 & 14.0 & 20.7 & 24.1 & 5.1  & 0.0 & 13.8 \\
Show-o2-7B~\cite{showo2}                 & 4.6  & 18.2 & 16.2 & 19.0 & 19.5 & 7.5  & 0.0 & 11.9 \\
BLIP3o-NEXT-GRPO-Text-3B~\cite{blip3o}   & 3.7  & 20.9 & 15.6 & 7.9  & 13.3 & 7.8  & 0.0 & 10.5 \\
\midrule
\rowcolor{gray!12}
\textbf{\ProjectName{}-DiMOO}    & 31.5 & 58.7 & \textbf{65.8} & \underline{68.5} & \underline{48.0} & \underline{49.8} & \textbf{8.0} & \underline{50.4} \\
\rowcolor{gray!12}
\textbf{\ProjectName{}-BAGEL}    & \textbf{42.4} & \textbf{72.2} & \underline{60.0} & \textbf{76.8} & 43.8 & 49.6 & \underline{5.3} & \textbf{54.8} \\
\bottomrule
\end{tabular}
\end{table}

\subsection{\BenchName{}: structured, per-attribute evaluation}
\label{sec:res-inhouse}

GenExam provides a common external yardstick, but a single judge score cannot say \emph{which} physical attribute a diagram gets wrong. We therefore introduce \BenchName{}, an in-house benchmark of $1{,}283$ held-out, structurally annotated diagrams balanced across the six subdisciplines (\Cref{tab:data}, Methods, \Cref{sec:methods-inhouse}).

What distinguishes \BenchName{} is not the use of a judge (like GenExam, it scores generations with a vision--language model) but that the questions the judge answers are \emph{derived from each item's structured annotation} rather than authored separately. Every diagram's annotation is compiled into an item-specific bank of binary (yes/no) questions: a rule-generated \emph{local} bank that enumerates the drawn objects, forces, states and their attributes (existence, relation and direction, and value questions), and a small model-generated \emph{global} bank of whole-diagram questions. The judge (GPT-4o) is shown only the generated image and the questions, never the gold answers, and answers each ``Yes'' or ``No''. Each answer is then matched against the gold answer stored with the question, so a score decomposes into named physical facts (``normal force present: yes; the $mg\sin\theta$ component drawn: no'') rather than a single opaque number. Two properties follow.

\emph{The question list is dynamic and item-specific.} An item receives as many local questions as its physics is rich: a single block on an incline yields on the order of a dozen, whereas a dense optical figure or a multi-stage thermodynamic cycle yields dozens to over a hundred, covering every object, state, process and direction. No two items share a question list, and the length of an item's local list is a direct, annotation-derived measure of how much its diagram must get right. We use exactly this quantity to define the difficulty tertiles analysed below. A benchmark with a fixed question format cannot express this, because the number of things that can go wrong in a physics diagram is a property of the physics, not of the test designer.

\emph{The gold answers are grounded in expert-verified annotations.} Because \BenchName{} is drawn from the expert-level tier (Methods, \Cref{sec:methods-data}), its questions and their gold answers come from annotations a human expert has checked against the diagram, not from labels accepted as a model produced them.

We report three scoring modes. \emph{Local} and \emph{Global} scoring are the judge's accuracy, the fraction of questions answered correctly, on the local and global banks respectively. Because the local bank enumerates individual physical facts while the global bank asks a few whole-diagram questions, Local is the more demanding of the two and every model scores lower under it. \emph{Strict} scoring (\Cref{tab:results-strict}) instead counts an item as correct only if its fraction of wrong answers stays within a per-item tolerance $\tau \in \{0\%, 5\%, 10\%\}$ on the local bank. At $\tau = 0\%$ every question of the item must be answered correctly, so it reports the fraction of items answered perfectly. Writing $c_q=\mathbf{1}[J(\hat x,q)=y_q]$ for whether the judge $J$ answers question $q$ correctly on a generation $\hat x$ against its gold answer $y_q$ (\Cref{sec:formal-annotation}),
\begin{equation}
  \mathrm{Local}(\hat x)=\frac{1}{|Q_{\mathrm{loc}}|}\!\sum_{q\in Q_{\mathrm{loc}}}\! c_q,
  \qquad
  \mathrm{Strict}_\tau(\hat x)=\mathbf{1}\!\Bigl[\,\textstyle\sum_{q\in Q_{\mathrm{loc}}}(1-c_q)\;\le\;\tau\,\lvert Q_{\mathrm{loc}}\rvert\,\Bigr],
  \label{eq:m-scores}
\end{equation}
with $\mathrm{Global}$ defined like $\mathrm{Local}$ over the global bank $Q_{\mathrm{glo}}$.

\paragraph{Main results.}
\Cref{tab:results-inhouse-subjects} reports per-subject and overall scores in both modes. \ProjectName{} built on BAGEL attains the best overall score under both modes, $75.69$ Local and $82.54$ Global, and is the best model on every subject under Global scoring and on five of the six subjects under Local scoring. The exception is optics, where the closed reference systems Gemini~2.5 Flash Image ($82.33$) and GPT-Image-1 ($79.54$) remain ahead of our $78.92$. \ProjectName{} built on DiMOO is second almost everywhere under Global scoring and on four of six subjects under Local scoring. The most informative comparison is against the unmodified backbone: BAGEL alone scores $46.38$ Local and $62.15$ Global, so training on the structured annotations adds $29.31$ points Local and $20.39$ points Global without changing the architecture. Among open-weight baselines the strongest are Qwen-Image at $51.79$ Local and BAGEL itself at $62.15$ Global, which \ProjectName{} exceeds by $23.90$ and $20.39$ points. The closed reference systems are considerably stronger: the best of them reach $69.98$ Local (Gemini~2.5 Flash Image) and $68.56$ Global (GPT-Image-1), and \ProjectName{} still leads them by $5.71$ and $13.98$ points despite a $14$B open backbone. The gains are not confined to one subdiscipline. Thermodynamics is the hardest subject for every model, where the best baseline reaches only $58.90$ Local and no open-weight baseline exceeds $32.91$, yet \ProjectName{} reaches $66.95$.

\paragraph{Per-step faithfulness.}
Decomposing the same evaluation by the five \SPCoT{} steps (\Cref{fig:step-scores}) shows where the gain is concentrated. \ProjectName{} is the strongest system on four of the five steps ($79.5$ Scenario, $74.8$ Parameters, $69.2$ Structure, and $74.8$ Laws) and within half a point of the best on the fifth, scoring $85.8$ on Synthesis against GPT-Image-1's $86.3$. Its separation from the strong closed reference systems falls almost entirely on the three intermediate reasoning steps, where it leads the best closed model by $12.2$ points on Parameters, $6.7$ on Structure and $3.5$ on Laws, while on the two appearance-anchored endpoints (Scenario and Synthesis) those systems are level. Against open-weight models the margin is far larger and widens toward the reasoning steps. On Laws, the step that most requires inferring a coordinate system and governing equations rather than reading them off the page, \ProjectName{} reaches $74.8$ where the best open-weight baseline manages only $47.9$ and several fall to $23$--$37$. The comparison against the unmodified backbone is sharpest of all: structured supervision lifts BAGEL on every step, by roughly $26$ points on Scenario, Parameters and Structure, and by $38.8$ and $42.1$ on Laws ($36.0\!\rightarrow\!74.8$) and Synthesis ($43.7\!\rightarrow\!85.8$). The advantage is thus concentrated, as intended, on the steps that demand physical reasoning rather than surface appearance.

\paragraph{Effect of prompt difficulty.}
\Cref{tab:results-difflevel} splits each subject by prompt difficulty, where difficulty is the number of QA checks an item carries and therefore the number of physical constraints the diagram must satisfy. Every model degrades from Easy to Hard, but by very different amounts, and the ordering of the three groups is consistent: the open-weight baselines collapse, the closed reference systems hold up considerably better, and \ProjectName{} degrades least. Thermodynamics is the clearest case. The best open-weight baseline falls from $61.33$ on Easy to $27.40$ on Hard, the best closed system from $67.66$ to $53.13$, while \ProjectName{} falls only from $72.13$ to $64.99$. Its margin over the best baseline at each level therefore widens from $4.47$ points on Easy to $11.86$ points on Hard. In acoustics the same widening appears more sharply still, from $1.27$ points on Easy to $14.06$ on Hard. In optics and acoustics \ProjectName{} is essentially flat across the three levels, and its Hard score even exceeds its Easy score, $80.44$ against $79.01$ in optics and $77.39$ against $71.13$ in acoustics. Structured supervision therefore helps most where the physics is most constrained, which is precisely what the framework is designed to do. Optics is the one subject that resists this pattern: Gemini~2.5 Flash Image leads at every difficulty level and is itself nearly flat ($83.21$ to $82.32$), suggesting that the optical-path constraints our schema encodes are the ones a strong general system is already most likely to satisfy.

\paragraph{Strict scoring.}
The scores above award partial credit. \Cref{tab:results-strict} instead scores an item as correct only if the fraction of its local questions the judge answers wrongly stays within a per-item tolerance $\tau$, reported at $\tau \in \{0\%, 5\%, 10\%\}$. Under a zero tolerance ($\tau=0\%$, every question of the item correct) every model scores $0.00$ on every subject except electromagnetism, where Gemini~2.5 Flash Image, GPT-Image-1, and \ProjectName{} alike reach only $0.31$, so no current system reliably produces a diagram whose every checked attribute is right. 
Tolerating $5\%$ and $10\%$ of the questions wrong separates the models, and \ProjectName{} leads every column but one, reaching $25.62$ on electromagnetism at $\tau=10\%$ against $14.47$ for the best baseline. The exception is again optics at $\tau=10\%$, where Gemini~2.5 Flash Image reaches $18.45$ against our $9.52$. 
Even so, the absolute numbers stay low across the board, which shows that fully faithful scientific-diagram generation is far from solved and that our gains, though large in relative terms, leave substantial headroom.

\begin{table}[t]
\centering
\caption{
\textbf{\BenchName{}} main results.
Per-subject columns report the score (\%) for each of the six physics topics:
E\&M (electromagnetism),
Mech.\ (mechanics),
Optics,
QM (quantum mechanics),
Thermo.\ (thermodynamics), and
Acou.\ (acoustics).
Local and Global are the judge's yes/no accuracy on \BenchName{}'s two
question banks: the rule-generated \emph{local} bank of per-attribute
(object/force/state) questions, and the model-generated \emph{global} bank of
whole-diagram questions. Overall is the
aggregate over the six subjects under the corresponding mode. All scores are
percentages, higher is better.
Within each mode, the best value per column is in \textbf{bold} and the second
best is \underline{underlined}.}
\label{tab:results-inhouse-subjects}
\small
\setlength{\tabcolsep}{4pt}
\resizebox{0.95\linewidth}{!}{%
\begin{tabular}{l cccccc c}
\toprule
& \multicolumn{6}{c}{\textbf{Per-subject}} & \\
\cmidrule(lr){2-7}
\textbf{Model} & E\&M & Mech. & Optics & QM & Thermo. & Acou. & \textbf{Overall} \\
\midrule
\multicolumn{8}{l}{\emph{Local scoring}} \\
\midrule
Seedream 4.0~\cite{seedream}             & 65.55 & 56.19 & 77.81 & 69.67 & 42.91 & 62.42 & 64.80 \\
Gemini 2.5 Flash Image~\cite{google2025geminiflashimage}      & 67.75 & 61.63 & \textbf{82.33} & 73.27 & 58.90 & \underline{68.44} & 69.98 \\
GPT-Image-1~\cite{openai2025gptimage1} & 69.58 & 60.25 & \underline{79.54} & 71.35 & 49.24 & 66.88 & 68.32 \\
Qwen-Image~\cite{qwenimage}              & 47.05 & 43.20 & 73.06 & 47.18 & 32.91 & 52.31 & 51.79 \\
HiDream-I1-Full~\cite{hidream}           & 45.26 & 41.44 & 55.51 & 49.75 & 29.34 & 50.20 & 46.48 \\
FLUX.1 dev~\cite{flux}                   & 48.89 & 40.25 & 54.60 & 40.05 & 23.72 & 46.04 & 45.18 \\
SD 3.5 Large~\cite{sd3}                  & 51.49 & 48.96 & 64.81 & 45.34 & 30.83 & 51.89 & 51.65 \\
BAGEL~\cite{bagel}                       & 43.51 & 44.85 & 51.66 & 51.03 & 31.57 & 54.18 & 46.38 \\
DiMOO~\cite{luminadimoo}                 & 31.95 & 35.09 & 45.06 & 41.39 & 22.41 & 45.37 & 37.01 \\
Show-o2-7B~\cite{showo2}                 & 36.79 & 41.22 & 53.97 & 44.43 & 31.38 & 46.48 & 42.88 \\
BLIP3o-NEXT-GRPO-Text-3B~\cite{blip3o}   & 36.36 & 36.02 & 41.68 & 45.36 & 22.44 & 41.46 & 37.65 \\
\midrule
\rowcolor{gray!12}
\textbf{\ProjectName{}-DiMOO}    & \underline{76.00} & \underline{68.24} & 75.64 & \underline{73.41} & \underline{60.40} & 68.39 & \underline{72.05} \\
\rowcolor{gray!12}
\textbf{\ProjectName{}-BAGEL}    & \textbf{78.59} & \textbf{71.59} & 78.92 & \textbf{77.59} & \textbf{66.95} & \textbf{73.01} & \textbf{75.69} \\
\midrule
\multicolumn{8}{l}{\emph{Global scoring}} \\
\midrule
Seedream 4.0~\cite{seedream}             & 59.94 & 66.00 & 75.83 & 58.99 & 66.82 & 66.38 & 65.07 \\
Gemini 2.5 Flash Image~\cite{google2025geminiflashimage}     & 61.13 & 65.48 & 77.14 & 60.75 & 67.63 & 69.45 & 66.14 \\
GPT-Image-1~\cite{openai2025gptimage1}  & 64.69 & 68.83 & 77.38 & 61.76 & 72.60 & 68.96 & 68.56 \\
Qwen-Image~\cite{qwenimage}              & 55.94 & 60.53 & 72.62 & 51.82 & 63.01 & 60.37 & 60.20 \\
HiDream-I1-Full~\cite{hidream}           & 54.19 & 58.67 & 66.31 & 52.58 & 59.65 & 58.90 & 57.96 \\
FLUX.1 dev~\cite{flux}                   & 52.06 & 57.73 & 64.17 & 51.32 & 55.72 & 59.51 & 56.32 \\
SD 3.5 Large~\cite{sd3}                  & 56.06 & 60.53 & 68.21 & 52.96 & 61.62 & 60.12 & 59.58 \\
BAGEL~\cite{bagel}                       & 56.38 & 61.07 & 68.81 & 60.38 & 64.05 & 68.34 & 62.15 \\
DiMOO~\cite{luminadimoo}                 & 50.62 & 58.20 & 59.88 & 54.47 & 56.53 & 59.02 & 55.95 \\
Show-o2-7B~\cite{showo2}                 & 53.52 & 59.33 & 66.59 & 55.67 & 60.35 & 58.62 & 58.44 \\
BLIP3o-NEXT-GRPO-Text-3B~\cite{blip3o}   & 52.00 & 57.87 & 60.12 & 54.97 & 57.23 & 58.65 & 56.35 \\
\midrule
\rowcolor{gray!12}
\textbf{\ProjectName{}-DiMOO}    & \underline{76.18} & \underline{80.47} & \underline{79.64} & \underline{71.07} & \underline{77.46} & \underline{76.81} & \underline{77.25} \\
\rowcolor{gray!12}
\textbf{\ProjectName{}-BAGEL}    & \textbf{80.56} & \textbf{83.40} & \textbf{83.93} & \textbf{83.77} & \textbf{83.47} & \textbf{81.23} & \textbf{82.54} \\
\bottomrule
\end{tabular}%
}
\end{table}

\begin{table}[t]
\centering
\caption{
\textbf{\BenchName{}} results by prompt difficulty level.
Per-subject scores (\%) are split into Easy / Medium / Hard prompts for each
of the six physics subjects (E\&M, Mechanics, Optics, Quantum Mechanics,
Thermodynamics, and Acoustics).
The difficulty of each item is measured by the length of its QA list, the
number of question and answer checks in its structured annotation. Within each
subdiscipline, items are partitioned into Easy, Medium, and Hard by the
tertiles of this QA count: the bottom, middle, and top third. A longer QA list
implies more physical constraints to satisfy, and hence a more complex diagram
that is harder to generate faithfully, so Hard collects the items with the most
QA checks. Higher is better.
Per column, the best value is in \textbf{bold} and the second best is \underline{underlined}.}
\label{tab:results-difflevel}
\small
\setlength{\tabcolsep}{4pt}
\resizebox{\linewidth}{!}{%
\begin{tabular}{l ccc ccc ccc}
\toprule
& \multicolumn{3}{c}{\textbf{E\&M}} & \multicolumn{3}{c}{\textbf{Mechanics}} & \multicolumn{3}{c}{\textbf{Optics}} \\
\cmidrule(lr){2-4}\cmidrule(lr){5-7}\cmidrule(lr){8-10}
\textbf{Model} & Easy & Med. & Hard & Easy & Med. & Hard & Easy & Med. & Hard \\
\midrule
Seedream 4.0~\cite{seedream}             & 75.32 & 68.70 & 53.89 & 66.08 & 58.97 & 50.67 & 80.81 & 76.11 & 76.36 \\
Gemini 2.5 Flash Image~\cite{google2025geminiflashimage} & 80.56 & 71.33 & 54.34 & 67.84 & 61.79 & 58.73 & \textbf{83.21} & \textbf{80.51} & \textbf{82.32} \\
GPT-Image-1~\cite{openai2025gptimage1}   & 80.41 & 72.64 & 57.29 & 68.19 & 62.82 & 55.70 & \underline{81.79} & \underline{79.34} & 77.48 \\
Qwen-Image~\cite{qwenimage}              & 63.36 & 49.30 & 34.06 & 61.04 & 45.78 & 35.12 & 75.05 & 71.93 & 72.22 \\
HiDream-I1-Full~\cite{hidream}           & 60.27 & 45.88 & 34.75 & 59.13 & 42.33 & 34.80 & 60.05 & 59.93 & 49.59 \\
FLUX.1 dev~\cite{flux}                   & 58.95 & 51.78 & 39.05 & 58.17 & 41.89 & 32.94 & 61.71 & 56.96 & 47.97 \\
SD 3.5 Large~\cite{sd3}                  & 67.20 & 53.32 & 38.45 & 62.25 & 51.59 & 41.95 & 66.97 & 68.73 & 60.29 \\
BAGEL~\cite{bagel}                       & 61.89 & 45.55 & 29.90 & 60.53 & 48.16 & 37.20 & 63.47 & 56.10 & 40.41 \\
DiMOO~\cite{luminadimoo}                 & 47.29 & 33.36 & 22.51 & 55.68 & 36.50 & 27.98 & 52.60 & 48.77 & 37.71 \\
Show-o2-7B~\cite{showo2}                 & 56.14 & 39.16 & 23.99 & 59.67 & 42.77 & 33.63 & 65.83 & 52.50 & 46.96 \\
BLIP3o-NEXT-GRPO-Text-3B~\cite{blip3o}   & 53.58 & 36.84 & 25.64 & 54.56 & 36.69 & 30.40 & 48.63 & 44.73 & 35.85 \\
\midrule
\rowcolor{gray!12}
\textbf{\ProjectName{}-DiMOO}    & \underline{82.11} & \underline{79.35} & \underline{69.09} & \underline{74.41} & \underline{72.28} & \underline{64.74} & 78.05 & 75.60 & 74.17 \\
\rowcolor{gray!12}
\textbf{\ProjectName{}-BAGEL}    & \textbf{84.74} & \textbf{83.28} & \textbf{69.97} & \textbf{76.77} & \textbf{75.93} & \textbf{68.34} & 79.01 & 77.80 & \underline{80.44} \\
\bottomrule
\end{tabular}%
}

\vspace{2ex}

\resizebox{\linewidth}{!}{%
\begin{tabular}{l ccc ccc ccc}
\toprule
& \multicolumn{3}{c}{\textbf{Quantum Mechanics}} & \multicolumn{3}{c}{\textbf{Thermodynamics}} & \multicolumn{3}{c}{\textbf{Acoustics}} \\
\cmidrule(lr){2-4}\cmidrule(lr){5-7}\cmidrule(lr){8-10}
\textbf{Model} & Easy & Med. & Hard & Easy & Med. & Hard & Easy & Med. & Hard \\
\midrule
Seedream 4.0~\cite{seedream}             & 70.97 & 70.16 & 64.42 & 64.75 & 53.32 & 34.29 & 64.88 & 65.36 & 58.43 \\
Gemini 2.5 Flash Image~\cite{google2025geminiflashimage} & 74.13 & \underline{74.05} & 67.52 & 66.77 & \underline{67.12} & 53.13 & \underline{69.86} & \underline{72.93} & 63.33 \\
GPT-Image-1~\cite{openai2025gptimage1}   & 75.41 & 70.04 & 65.79 & 67.66 & 58.44 & 42.48 & 69.19 & 69.32 & 62.90 \\
Qwen-Image~\cite{qwenimage}              & 54.33 & 49.30 & 39.74 & 61.33 & 40.70 & 24.90 & 57.42 & 51.08 & 50.31 \\
HiDream-I1-Full~\cite{hidream}           & 56.83 & 49.11 & 44.70 & 55.96 & 35.57 & 22.98 & 56.04 & 51.80 & 45.14 \\
FLUX.1 dev~\cite{flux}                   & 50.94 & 38.99 & 34.63 & 52.31 & 29.58 & 17.27 & 51.48 & 46.67 & 43.78 \\
SD 3.5 Large~\cite{sd3}                  & 54.81 & 43.42 & 40.98 & 59.99 & 36.12 & 24.22 & 55.94 & 49.37 & 52.21 \\
BAGEL~\cite{bagel}                       & 64.95 & 49.81 & 42.30 & 57.60 & 35.39 & 27.40 & 62.72 & 54.86 & 48.68 \\
DiMOO~\cite{luminadimoo}                 & 52.09 & 39.96 & 37.14 & 50.82 & 26.61 & 17.52 & 50.81 & 48.11 & 41.34 \\
Show-o2-7B~\cite{showo2}                 & 52.70 & 45.56 & 38.82 & 56.71 & 34.79 & 26.88 & 53.04 & 47.27 & 43.74 \\
BLIP3o-NEXT-GRPO-Text-3B~\cite{blip3o}   & 52.17 & 47.32 & 40.03 & 48.88 & 26.25 & 18.67 & 48.45 & 42.93 & 37.96 \\
\midrule
\rowcolor{gray!12}
\textbf{\ProjectName{}-DiMOO}    & \underline{75.58} & 73.46 & \underline{70.53} & \underline{68.93} & 64.79 & \underline{58.68} & 67.84 & 70.50 & \underline{69.85} \\
\rowcolor{gray!12}
\textbf{\ProjectName{}-BAGEL}    & \textbf{81.34} & \textbf{78.37} & \textbf{75.28} & \textbf{72.13} & \textbf{73.57} & \textbf{64.99} & \textbf{71.13} & \textbf{73.33} & \textbf{77.39} \\
\bottomrule
\end{tabular}%
}
\end{table}

\begin{table}[t]
\centering
\caption{
\textbf{\BenchName{}} strict scores on the local question bank.
An item counts as correct only if the fraction of its local questions the
judge answers wrongly stays within the per-item tolerance Tol.; the score is
then the fraction (\%) of items that pass. Tol.\ $=0\%$ requires every question
of the item to be answered correctly, while $5\%$ and $10\%$ tolerate that fraction
of wrong answers per item. Results are reported for each of the six physics
subjects. Higher is better.
Per column, the best value is in \textbf{bold} and the second best is
\underline{underlined}; columns in which every model scores $0.00$ are left
unmarked.}
\label{tab:results-strict}
\small
\setlength{\tabcolsep}{4pt}
\resizebox{\linewidth}{!}{%
\begin{tabular}{l ccc ccc ccc}
\toprule
& \multicolumn{3}{c}{\textbf{E\&M}} & \multicolumn{3}{c}{\textbf{Mechanics}} & \multicolumn{3}{c}{\textbf{Optics}} \\
\cmidrule(lr){2-4}\cmidrule(lr){5-7}\cmidrule(lr){8-10}
\textbf{Model} & Tol.\ 0\% & Tol.\ 5\% & Tol.\ 10\% & Tol.\ 0\% & Tol.\ 5\% & Tol.\ 10\% & Tol.\ 0\% & Tol.\ 5\% & Tol.\ 10\% \\
\midrule
Seedream 4.0~\cite{seedream}             & 0.00 & 0.94 & 7.81 & 0.00 & 0.33 & 0.33 & 0.00 & 2.38 & \underline{10.12} \\
Gemini 2.5 Flash Image~\cite{google2025geminiflashimage} & \textbf{0.31} & \underline{4.40} & 14.47 & 0.00 & 0.00 & 2.01 & 0.00 & 0.60 & \textbf{18.45} \\
GPT-Image-1~\cite{openai2025gptimage1}   & \textbf{0.31} & 2.50 & 11.56 & 0.00 & 0.00 & 1.67 & 0.00 & \underline{2.98} & \underline{10.12} \\
Qwen-Image~\cite{qwenimage}              & 0.00 & 0.62 & 2.50 & 0.00 & 0.00 & 0.33 & 0.00 & 1.19 & 8.33 \\
HiDream-I1-Full~\cite{hidream}           & 0.00 & 0.62 & 2.50 & 0.00 & 0.33 & 0.67 & 0.00 & 0.00 & 1.79 \\
FLUX.1 dev~\cite{flux}                   & 0.00 & 0.62 & 2.50 & 0.00 & 0.00 & 0.00 & 0.00 & 0.60 & 1.79 \\
SD 3.5 Large~\cite{sd3}                  & 0.00 & 1.25 & 7.50 & 0.00 & 0.00 & 0.67 & 0.00 & 0.00 & 5.95 \\
BAGEL~\cite{bagel}                       & 0.00 & 1.88 & 5.31 & 0.00 & 0.00 & 1.00 & 0.00 & 0.60 & 1.79 \\
DiMOO~\cite{luminadimoo}                 & 0.00 & 0.00 & 0.31 & 0.00 & 0.00 & 0.33 & 0.00 & 1.19 & 4.76 \\
Show-o2-7B~\cite{showo2}                 & 0.00 & 0.32 & 2.86 & 0.00 & 0.00 & 1.34 & 0.00 & 0.00 & 2.99 \\
BLIP3o-NEXT-GRPO-Text-3B~\cite{blip3o}   & 0.00 & 0.00 & 1.88 & 0.00 & 0.00 & 0.33 & 0.00 & 0.00 & 0.00 \\
\midrule
\rowcolor{gray!12}
\textbf{\ProjectName{}-DiMOO}    & 0.00 & \underline{2.51} & \underline{18.81} & 0.00 & \underline{1.00} & \underline{5.00} & 0.00 & 0.00 & 2.98 \\
\rowcolor{gray!12}
\textbf{\ProjectName{}-BAGEL}    & \textbf{0.31} & \textbf{7.19} & \textbf{25.62} & 0.00 & \textbf{1.33} & \textbf{5.33} & 0.00 & \textbf{3.57} & 9.52 \\
\bottomrule
\end{tabular}%
}

\vspace{2ex}

\resizebox{\linewidth}{!}{%
\begin{tabular}{l ccc ccc ccc}
\toprule
& \multicolumn{3}{c}{\textbf{Quantum Mechanics}} & \multicolumn{3}{c}{\textbf{Thermodynamics}} & \multicolumn{3}{c}{\textbf{Acoustics}} \\
\cmidrule(lr){2-4}\cmidrule(lr){5-7}\cmidrule(lr){8-10}
\textbf{Model} & Tol.\ 0\% & Tol.\ 5\% & Tol.\ 10\% & Tol.\ 0\% & Tol.\ 5\% & Tol.\ 10\% & Tol.\ 0\% & Tol.\ 5\% & Tol.\ 10\% \\
\midrule
Seedream 4.0~\cite{seedream}             & 0.00 & 0.00 & 0.63 & 0.00 & 0.58 & 1.16 & 0.00 & 0.00 & 0.61 \\
Gemini 2.5 Flash Image~\cite{google2025geminiflashimage} & 0.00 & \textbf{1.26} & \underline{4.40} & 0.00 & \underline{1.16} & \underline{5.78} & 0.00 & 0.00 & 0.00 \\
GPT-Image-1~\cite{openai2025gptimage1}   & 0.00 & \underline{0.63} & 1.89 & 0.00 & 0.00 & 3.47 & 0.00 & 0.00 & 0.00 \\
Qwen-Image~\cite{qwenimage}              & 0.00 & 0.00 & 0.00 & 0.00 & 0.00 & 0.00 & 0.00 & 0.00 & 0.00 \\
HiDream-I1-Full~\cite{hidream}           & 0.00 & 0.00 & 0.63 & 0.00 & 0.00 & 0.58 & 0.00 & 0.00 & 0.00 \\
FLUX.1 dev~\cite{flux}                   & 0.00 & 0.00 & 0.00 & 0.00 & 0.00 & 1.16 & 0.00 & 0.00 & 0.00 \\
SD 3.5 Large~\cite{sd3}                  & 0.00 & 0.00 & 0.00 & 0.00 & 0.00 & 0.58 & 0.00 & 0.61 & 0.61 \\
BAGEL~\cite{bagel}                       & 0.00 & 0.00 & 0.63 & 0.00 & 0.00 & 2.89 & 0.00 & 0.00 & 0.61 \\
DiMOO~\cite{luminadimoo}                 & 0.00 & 0.00 & 0.00 & 0.00 & 0.00 & 0.58 & 0.00 & 0.61 & 0.61 \\
Show-o2-7B~\cite{showo2}                 & 0.00 & 0.00 & 0.00 & 0.00 & 0.00 & 2.91 & 0.00 & \underline{0.63} & 1.26 \\
BLIP3o-NEXT-GRPO-Text-3B~\cite{blip3o}   & 0.00 & 0.00 & 0.00 & 0.00 & 0.00 & 1.16 & 0.00 & 0.61 & 0.61 \\
\midrule
\rowcolor{gray!12}
\textbf{\ProjectName{}-DiMOO}    & 0.00 & 0.00 & 1.26 & 0.00 & 0.58 & 3.47 & 0.00 & 0.61 & \underline{3.07} \\
\rowcolor{gray!12}
\textbf{\ProjectName{}-BAGEL}    & 0.00 & \textbf{1.26} & \textbf{7.55} & 0.00 & \textbf{3.47} & \textbf{9.25} & 0.00 & \textbf{1.23} & \textbf{6.75} \\
\bottomrule
\end{tabular}%
}
\end{table}



\subsection{Automatic faithfulness evaluation}
\label{sec:res-autoeval}

The per-attribute scores of \Cref{sec:res-inhouse} are produced automatically, with no human in the loop at scoring time, and the evaluator reuses the annotation directly rather than asking a general judge for one holistic verdict. Each item's structured analysis is compiled by rule into the local yes/no checklist, a vision--language model adds a small global checklist, and an off-the-shelf judge model answers every question about the generated image. Faithfulness is then reported as named per-attribute outcomes (e.g.\ ``normal force present and correctly directed: yes; the $mg\sin\theta$ component drawn: no'') rather than a single opaque number (Methods, \Cref{sec:methods-autoeval}). Because the annotation is already typed and machine-parseable, the checklist is built without training a bespoke extractor: the judge is the only learned component, and it never sees the gold answers. The schema's visible/inferred split further tags each check as grounded or inferred, a distinction the current equal-weight scoring does not yet use but that a future weighted or preference-based score could.

\section{Discussion}
\label{sec:discussion}

We have shown that the principal obstacle to generating physically faithful scientific diagrams is a supervision gap, not an architectural one, and that it can be closed by making the physics behind each diagram an explicit chain of reasoning. By replacing flat captions with \SPCoT{}, a unified multi-step chain that separates what is visually grounded from what is physically inferred and types all mathematics symbolically, we obtain supervision that is dense, auditable, and aligned across six physics subdisciplines, and a generator, \ProjectName{}, that produces diagrams more consistent with the underlying physics than generic baselines.

\subsection{Why structured physical supervision helps}
\label{sec:disc-why}

A flat caption supervises a generator on appearance. A structured analysis supervises it on the reasoning that produces that appearance. Three properties of the framework explain the effect. First, the schema makes the latent variables of a diagram explicit (forces, states, processes, governing laws), so the training signal pushes the model toward diagrams whose drawn marks are consistent with stated physics rather than merely typical of the subdiscipline. Second, the visible/inferred split keeps the grounded fields verifiable against the pixels while still exposing the model to the physics the diagram is meant to teach, which prevents the annotations from drifting into hallucinated detail. Third, because the first and last steps of the schema are structurally identical across subdisciplines, representation is shared, and improvements in one domain (e.g.\ rendering a directed force arrow) transfer to analogous directed quantities in others. The same structure that improves generation also enables interpretable evaluation: the key-value form of the annotations lets an automatic evaluator report \emph{which} physical attribute is right or wrong, rather than a single opaque score.

\subsection{What a unified backbone contributes}
\label{sec:disc-unified-models}

We train \ProjectName{} on two unified multimodal backbones that differ in both mechanism and scale: BAGEL~\cite{bagel}, a $14$B mixture-of-transformers model, and DiMOO~\cite{luminadimoo}, a $7$B discrete-diffusion model. For both, the unmodified backbone is itself among our baselines, so the effect of structured supervision can be measured directly and with no change to the architecture. On \BenchName{}, BAGEL improves from $46.38$ to $75.69$ Local and from $62.15$ to $82.54$ Global, and DiMOO from $37.01$ to $72.05$ Local and from $55.95$ to $77.25$ Global. On the GenExam physics subset, BAGEL improves from $13.8$ to $54.8$ relaxed, a fourfold gain. That two mechanistically dissimilar models improve by comparable margins argues that what we are adding is supervision, not an architectural advantage particular to one design.

It would be tempting to conclude that unified models are simply the right architecture for scientific diagrams, but our baselines say otherwise. \emph{Before} structured supervision, the unified models are the weaker group: the best open unified baseline (BAGEL) trails the best open text-to-image baseline (Qwen-Image) by $5.41$ points Local on \BenchName{} and by $12.5$ points relaxed on GenExam, and the DiMOO backbone is the single weakest open model of any kind on Local, at $37.01$. Coupling understanding and generation in one network therefore buys no head start on physical faithfulness by itself. What it buys is the capacity to be taught: DiMOO begins last among open systems and, trained on structured analyses, ends second overall at $72.05$, a $35.04$-point rise that is the largest of any model and larger even than BAGEL's $29.31$. The unified design is best read not as a source of physical knowledge but as the interface through which physical knowledge can be delivered, because the network that must hold the analysis is the network that draws.

Scale is not the explanation either. \ProjectName{}-DiMOO has $7$B parameters, roughly a third of Qwen-Image, yet exceeds it by $20.26$ points Local and $24.1$ points relaxed, and on GenExam it edges past Seedream~4.0, a closed reference system. Within our own results the ordering of the two backbones is consistent but modest, with \ProjectName{}-BAGEL ahead of \ProjectName{}-DiMOO by $3.64$ points Local, far smaller than the $29$-to-$35$-point rise each shows over its own untrained baseline. The dominant variable is what the model is trained on, not how large it is or which unified formulation it uses.

\subsection{Future Work}
\label{sec:disc-futurework}
\ProjectName{} builds on the rapid progress of unified multimodal models and high-fidelity T2I generators~\cite{bagel,sd3,flux,qwenimage,emu3,januspro,showo2}, but targets an axis those systems are not optimized for (physical correctness) and addresses it through data and supervision rather than a new backbone. It is complementary to multidisciplinary evaluation efforts such as GenExam~\cite{genexam}: where such benchmarks reveal that current models fail on scientific exams, our framework provides a route to the training signal needed to close that gap, and our structured-key-value evaluator offers a finer-grained, physics-aware alternative to single-score MLLM judging.

Several limitations remain. The current scope is physics and, within it, the six subdisciplines for which we have defined schemas. Other sciences, such as chemistry and biology, would each require their own structured templates. The two annotation tiers carry different guarantees. The expert-level subset, which supports fine-tuning and all evaluation, is verified by human experts. The corpus-level tier that supplies pre-training is machine-generated and unverified, so at that scale the quality of the inferred fields still depends on the annotating model's physical reasoning, constrained but not guaranteed by the fidelity rules. Evaluation of physical faithfulness is itself an open problem: GenExam's MLLM judge and our binary-checklist evaluator both rely on a vision-language model to answer the checks, and so remain proxies for expert judgment. Finally, the strongest closed models still set the reference, and the absolute scores on the physics subset remain low across the field, indicating substantial headroom.

\SPCoT{} is designed to extend. Quantum mechanics already demonstrates the point: its diagrams share almost no visual vocabulary with a free-body diagram, yet the same five-step, visible/inferred structure accommodates them once Step~3 is allowed to describe a state space rather than a force balance. The immediate next steps are geometric optics and the remaining quantitative sciences, to which the structure generalizes naturally. Beyond generation, physics-faithful diagram synthesis enables controllable, correct figure drafting for textbooks, problem sets, and lecture notes, and the large-scale generation of physically correct diagrams as training or augmentation data for scientific-document understanding and visual question answering. We see the structured-annotation framework, an auditable, machine-parseable encoding of the physics behind a figure, as the reusable contribution: a substrate for training, for evaluation, and potentially for preference optimization against the automatic faithfulness score.

\begingroup
\sloppy
\phantomsection 
\printbibliography[heading=bibintoc]
\endgroup

\clearpage
\begingroup
\centering
{\Large\bfseries Supplementary Information\par}
\vspace{0.5em}
\endgroup

\setcounter{section}{0}
\renewcommand{\thesection}{S\arabic{section}}
\renewcommand{\thesubsection}{S\arabic{section}.\arabic{subsection}}
\renewcommand{\theHsection}{supp.\arabic{section}}
\renewcommand{\theHsubsection}{supp.\arabic{section}.\arabic{subsection}}

\section{Methods}
\label{sec:methods}

This Methods section provides full procedural detail for the data pipeline, the annotation framework, the \ProjectName{} model, training, inference, and evaluation summarized in the main text.

\subsection{Data collection and curation}
\label{sec:methods-data}

We scope the corpus to physics, organized into six subdisciplines: mechanics, electromagnetism, physical optics, thermodynamics, acoustics, and quantum mechanics. We combine four data sources (\Cref{tab:data}): (i)~a manually curated gold seed selected by annotators against an explicit, written set of source-selection principles; (ii)~large-scale mining of physics-relevant imagery from OpenDataLab~\cite{opendatalab}; (iii)~targeted crawling of physics textbooks and reference materials from public web archives; and (iv)~a procurement pipeline for high-resolution, original English-language physics textbooks supplied as original-quality PDFs, governed by a formal requirements-and-acceptance specification (format: original-quality PDF; scope: physics subdiscipline textbooks/workbooks from high-school through graduate level, including problem sets and professional/engineering texts; language: English).

Raw images pass a multi-stage filter before they are eligible for structured annotation: (i)~de-duplication and resolution/aspect-ratio thresholds; (ii)~a subdiscipline classifier that routes each image to one of the six physics domains and discards non-physics content; and (iii)~a diagram-quality screen that removes photographs, decorative figures, and images whose physical content is too sparse to annotate.

\paragraph{Two annotation levels.}
The surviving corpus is annotated at two levels, and the distinction is one of provenance as well as of quality. Every image in the corpus receives a \emph{structured annotation} generated automatically by a vision-language model, Qwen3-VL-235B-A22B~\cite{qwen3vl}, prompted with the per-subdiscipline template of \Cref{sec:methods-templates} and required to return the schema as strict JSON. This is what makes annotation at the scale of millions of images feasible, and it supplies the breadth used for pre-training. A high-quality subset is then \emph{verified by human experts}, who check the machine-generated annotation against the diagram and correct it, with particular attention to the strictly-image-faithful steps, where a hallucinated object or a force in the wrong direction would otherwise propagate into supervision. This second tier is what supervised fine-tuning and all evaluation rest on, since a benchmark can be no more reliable than the annotations its questions are compiled from. Both tiers use the same schema and the same fidelity rules; they differ in whether a human has verified the result, not in its structure. Per-subdiscipline counts for both are reported in \Cref{tab:data}.

\subsection{Physics-structured annotation framework}
\label{sec:methods-framework}

\SPCoT{} instantiates the unified five-step schema of \Cref{tab:schema} per subdiscipline; a diagram's populated schema is its \emph{structured annotation}. Each annotation is a strict JSON object; all mathematical content is typed symbolically (\Cref{sec:formal-annotation}), and missing information must be encoded as an empty string \texttt{""} or empty list \texttt{[]} rather than guessed.

\paragraph{Fidelity rules.}
Every subdiscipline designates exactly two steps as \emph{strictly faithful to the image} (\Cref{tab:schema}): mechanics, scene/object identification (Step~1) and force analysis (Step~3); thermodynamics, state identification (Step~2) and the visualized process path (Step~3); acoustics, scene identification (Step~1) and the system components (Step~2); electromagnetism, component/source identification (Step~2) and the structural part of Step~3 (circuit topology or field distribution); physical optics, component identification (Step~2) and the geometric layout of Step~3 (the optical path); quantum mechanics, component parameterization (Step~2) and the geometric definitions of Step~3 (spatial regions and boundary conditions, or the basis and its eigenstates). For these steps the annotator may record only objects, labels, arrows, forces, components, connections, and states that are explicitly drawn, and may not invent any element that is not visible; in particular, numerical values may not be supplied where the diagram shows only variables or abstract symbols. The remaining steps may be inferred by physical reasoning when they can be logically deduced from the drawn elements. The full per-subdiscipline templates are given in \Cref{sec:methods-templates}.

\subsection{Physics-structured annotation: a formal framework}
\label{sec:formal-annotation}

We give a single formal object (the \emph{structured annotation}, one instance of \SPCoT{}) and show
that pre-training (\Cref{sec:methods-training}), supervised fine-tuning (\Cref{sec:methods-sft}) and
the \BenchName{} verification protocol (\Cref{sec:methods-inhouse}) are three uses of
that same object. This makes precise the claim of the main text that the
gains follow the supervision rather than the architecture: the annotation,
its serialization, and the binary checklist compiled from it are
backbone-agnostic, and only the generative loss changes with the backbone.

\paragraph{Structured annotations as typed, grounded objects.}
Let $\mathcal{S}$ be the set of six physics subdisciplines
($|\mathcal{S}|=6$: mechanics, electromagnetism, physical optics,
thermodynamics, acoustics, quantum mechanics), and let a diagram be an image
$x$ assigned to a subdiscipline $s\in\mathcal{S}$ by the routing classifier of
\Cref{sec:methods-data}. Each subdiscipline fixes a template $\mathcal{T}_s$ that
instantiates the unified five-step schema of \Cref{tab:schema}. A structured annotation
is the populated schema
\begin{equation}
  a \;=\; \bigl(a^{(1)},\dots,a^{(5)}\bigr),
  \qquad
  a^{(k)} \;=\; \bigl\{\,f=(\kappa_f,\nu_f)\,\bigr\},
  \label{eq:annotation}
\end{equation}
where step $k$ is a set of fields $f$ with key $\kappa_f$ (e.g.
\texttt{force.direction}) and value $\nu_f$. Every field carries a
\emph{type}
\begin{equation}
  \operatorname{type}(f)\in\{\textsc{ent},\textsc{rel},\textsc{val}\},
  \label{eq:type}
\end{equation}
distinguishing entities (an object, force, component, or state that is
present), relations (a direction, a process type, a topological connection),
and values (a numeric or symbolic quantity, always typed in \LaTeX). Missing
information is recorded as the empty value $\nu_f=\varnothing$ (an empty string
or list), never guessed.

The defining property of the framework is a \emph{grounding} map that labels
every field as read from the pixels or inferred by physical reasoning,
\begin{equation}
  g(f)\in\{\mathsf{vis},\mathsf{inf}\}.
  \label{eq:grounding}
\end{equation}
Each subdiscipline designates exactly two of the five steps as strictly
image-faithful, a set $\mathcal{G}_s\subset\{1,\dots,5\}$ with
$|\mathcal{G}_s|=2$ (\Cref{tab:schema}). Grounding is determined by the step,
\begin{equation}
  g(f)=\mathsf{vis}\ \text{iff}\ f\in a^{(k)}\ \text{with}\ k\in\mathcal{G}_s .
  \label{eq:grounded-steps}
\end{equation}
The fidelity rule constrains the visible fields to elements actually drawn: a
valid annotation of $x$ must satisfy
\begin{equation}
  \forall f: g(f)=\mathsf{vis}\ \Longrightarrow\ \nu_f\in \mathrm{drawn}(x),
  \label{eq:fidelity}
\end{equation}
where $\mathrm{drawn}(x)$ is the set of objects, labels, arrows, forces,
components, connections, and states explicitly present in the image. In
particular, a numeric value may not be supplied for a visible field where the
diagram shows only a symbol. Inferred fields ($g(f)=\mathsf{inf}$) may be
deduced from the drawn elements when logically entailed, and are $\varnothing$
otherwise.

\paragraph{Annotation and verification operators.}
Corpus-scale annotation is produced by a vision--language model
$\mathrm{A}$ (Qwen3-VL-235B-A22B~\cite{qwen3vl}) prompted with $\mathcal{T}_s$ and
required to return strict JSON obeying \eqref{eq:fidelity}:
\begin{equation}
  a \;=\; \mathrm{A}\!\left(x,\,\mathcal{T}_s\right),
  \qquad
  \mathcal{D}_{\mathrm{full}}
  =\bigl\{(x,s,a)\bigr\}.
  \label{eq:annotate}
\end{equation}
The expert tier applies a human verification operator $\mathrm{H}$ that
corrects $a$ against $x$, with priority on the visible steps
$\mathcal{G}_s$, yielding
$\mathcal{D}_{\mathrm{exp}}=\{(x,s,\mathrm{H}(x,a)):(x,s,a)\in
\mathcal{D}_{\mathrm{full}}^{\,0}\}$ on a high-quality subset
$\mathcal{D}_{\mathrm{full}}^{\,0}$. Both tiers share the schema, the types
\eqref{eq:type} and the grounding \eqref{eq:grounding}. They differ only in
whether $\mathrm{H}$ has been applied.

\paragraph{Serialization and conditioning.}
A deterministic serializer $\sigma$ renders an annotation as a conditioning
token sequence,
\begin{equation}
  c \;=\; \sigma(a)\in\mathcal{V}^{\ast},
  \label{eq:serialize}
\end{equation}
over the model vocabulary $\mathcal{V}$. Because $\sigma$ preserves the keys,
types and \LaTeX{} values verbatim, the same string that conditions
generation is machine-parseable back into $a$, which is what lets the
evaluator of \Cref{sec:methods-autoeval} reuse it without a separate parser.

\paragraph{Unified generative training objective.}
Let $G_\theta$ be the unified backbone and $\mathcal{L}_{\mathrm{gen}}(\theta;
x,c)$ its conditional generation loss for target image $x$ given conditioning
$c$. Pre-training and supervised fine-tuning minimize the same
subdiscipline-balanced risk,
\begin{equation}
  \mathcal{L}(\theta;\mathcal{D})
  \;=\;
  \mathbb{E}_{s\sim w}\;
  \mathbb{E}_{(x,a)\sim \mathcal{D}_s}\;
  \mathcal{L}_{\mathrm{gen}}\!\bigl(\theta;\,x,\,\sigma(a)\bigr),
  \label{eq:train}
\end{equation}
differing only in the data tier and mixture: pre-training uses
$\mathcal{D}=\mathcal{D}_{\mathrm{full}}$ for breadth, SFT continues on
$\mathcal{D}=\mathcal{D}_{\mathrm{exp}}$ for expert-verified correctness, and
the weights $w=(w_s)_{s\in\mathcal{S}}$, $\sum_s w_s=1$, re-balance the
severe corpus imbalance of \Cref{tab:data} (a fraction of generic image--text data is
retained during pre-training to preserve general rendering ability).

The loss $\mathcal{L}_{\mathrm{gen}}$ is the only architecture-dependent term.
For the rectified-flow backbone (BAGEL~\cite{bagel}) generation is in the latent
$z=\mathcal{E}(x)$ of a frozen VAE $\mathcal{E}$: with $\epsilon\sim
\mathcal{N}(0,\mathbf{I})$, a flow time $t\sim p_t$ on $[0,1]$ and the
interpolant $z_t=(1-t)\,z+t\,\epsilon$, the model regresses the velocity
$\dot z_t=\epsilon-z$,
\begin{equation}
  \mathcal{L}_{\mathrm{gen}}^{\mathrm{RF}}(\theta;x,c)
  =\mathbb{E}_{\epsilon,\,t}
  \bigl\|\,v_\theta(z_t,t,c)-(\epsilon-z)\,\bigr\|_2^{2}.
  \label{eq:rf}
\end{equation}
For the discrete masked-diffusion backbone (DiMOO~\cite{luminadimoo}) the image is a token
grid $y=Q(x)\in\{1,\dots,K\}^{N}$ from a VQ tokenizer $Q$. With a masked
subset $M\subset\{1,\dots,N\}$ of ratio $\gamma(t)$ the model predicts the
masked tokens,
\begin{equation}
  \mathcal{L}_{\mathrm{gen}}^{\mathrm{MDM}}(\theta;x,c)
  =\mathbb{E}_{t,\,M}
  \Bigl[-\tfrac{1}{|M|}\!\sum_{i\in M}\log
  p_\theta\!\bigl(y_i\mid y_{\bar M},\,c\bigr)\Bigr].
  \label{eq:mdm}
\end{equation}
Substituting either \eqref{eq:rf} or \eqref{eq:mdm} into \eqref{eq:train}
leaves the pipeline unchanged, which is exactly why the structured-annotation
supervision transfers across backbones.

\paragraph{Structured-prompt inference.}
\label{sec:methods-extension}
At inference time, the user supplies a free-form prompt $p$, ranging from a brief colloquial request to a detailed specification. A prompt-expansion model $\Phi$ identifies the physics subdiscipline $\hat{s}$, selects the corresponding schema $T_{\hat{s}}$ from Section~S2, and directly populates a complete SP-CoT annotation $\hat{a}$:
\begin{equation}
(\hat{s},\hat{a})
=
\Phi\left(p,\{T_s\}_{s\in\mathcal{S}}\right),
\qquad
\hat{x}\sim G_{\theta}\left(\cdot\mid\sigma(\hat{a})\right).
\label{eq:infer}
\end{equation}
Here, $\mathcal{S}$ denotes the six physics subdisciplines and $\sigma$ is the deterministic serializer. At inference, we retain the template's JSON structure and step-level fidelity partition while replacing the image-oriented instruction with prompt-grounded schema population. 
In addition to the five-step physical core defined in Equation (3), the complete JSON contains the \texttt{summary\_description} and \texttt{detailed\_description} fields, all generated in a single conversion.
Fields corresponding to image-faithful steps contain only information supported by the prompt, either explicitly stated or unambiguously implied; the remaining fields may include information reliably inferred from the input and governing physics, while indeterminate fields remain empty.
The output is checked for JSON validity and schema consistency; failed outputs are regenerated, with up to three retries. The validated JSON is serialized by $\sigma$ and used as the generator's conditioning input, normalizing prompts of varying styles and detail into the structured representation used during training.

\paragraph{Evaluation as binary-checklist verification.}
The annotation of a held-out item is read as its specification and compiled
into a checklist of binary (yes/no) checks, rather than scored by a holistic
judge. A rule-based compiler turns the non-empty fields of the grounded steps
into a \emph{local} question bank $Q_{\mathrm{loc}}(a)$, drawing on the field
types of \eqref{eq:type}: presence questions from entity fields
(``does the image show $X$?''), value-grounded questions from value fields
(``is the angle labelled $30^\circ$?''), and direction or distractor questions
from relational fields. A vision--language model produces a \emph{global} bank
$Q_{\mathrm{glo}}(a)$ of holistic faithfulness questions. Every question $q$
carries a gold answer $y_q\in\{0,1\}$ fixed by the annotation. Empty fields,
required by \eqref{eq:fidelity} wherever the diagram is silent, generate no
question, so an item is never penalized for an omission its reference does not
specify.

Verification is answer-matching, not numeric comparison. A vision judge
$J$ answers each question from the generated image alone, with the gold answer
withheld, and a check passes iff its answer matches the gold:
\begin{equation}
  \hat y_q = J(\hat x, q)\in\{0,1\},
  \qquad
  c_q = \mathbf{1}\!\left[\hat y_q = y_q\right].
  \label{eq:verify}
\end{equation}
The two banks give two readings, the mean check-pass rate over each,
\begin{equation}
  \mathrm{Local}(\hat x,a)=\!\frac{1}{|Q_{\mathrm{loc}}(a)|}\!\!\sum_{q\in Q_{\mathrm{loc}}(a)}\!\!\! c_q,
  \qquad
  \mathrm{Global}(\hat x,a)=\!\frac{1}{|Q_{\mathrm{glo}}(a)|}\!\!\sum_{q\in Q_{\mathrm{glo}}(a)}\!\!\! c_q,
  \label{eq:localglobal}
\end{equation}
so Local aggregates per-attribute correctness while Global rates whole-diagram
faithfulness. A question the judge fails to answer is counted wrong. A third,
strict reading admits an item only when its number of wrong local checks stays
within a tolerance $\tau$ of the checklist length,
\begin{equation}
  \mathrm{Strict}_\tau(\hat x,a)=\mathbf{1}\!\Bigl[\,\textstyle\sum_{q\in Q_{\mathrm{loc}}(a)}(1-c_q)\;\le\;\tau\,\lvert Q_{\mathrm{loc}}(a)\rvert\,\Bigr],
  \label{eq:strict}
\end{equation}
and reports the fraction of items that satisfy it, at
$\tau\in\{0\%,5\%,10\%\}$ (\Cref{tab:results-strict}); at $\tau=0$ every local check must pass.
Finally, the size of the local bank is an annotation-derived difficulty,
\begin{equation}
  d(a)=\lvert Q_{\mathrm{loc}}(a)\rvert,
  \label{eq:difficulty}
\end{equation}
whose within-subdiscipline tertiles define the Easy/Medium/Hard strata of
\Cref{tab:results-difflevel}: a diagram is hard exactly when its physics imposes many simultaneous
checks.

\begin{algorithm}[t]
\caption{Physics-structured annotation for faithful generation and evaluation}
\label{alg:pipeline}
\begin{algorithmic}[1]
\Require raw images $\mathcal{X}$; subdisciplines $\mathcal{S}$ and templates
  $\{\mathcal{T}_s\}$; annotator $\mathrm{A}$, expert verifier $\mathrm{H}$;
  backbone $G_\theta$ with loss $\mathcal{L}_{\mathrm{gen}}$; prompt expander
  $\mathrm{E}$; mixture $w$; tolerance $\tau$
\Statex \textit{// Corpus construction (\Cref{sec:methods-data})}
\State $\mathcal{D}_{\mathrm{full}}\gets\varnothing$
\For{$x\in\mathcal{X}$ surviving de-dup / resolution / quality filters}
  \State $s\gets\textsc{Classify}(x)$;\quad
         $a\gets\mathrm{A}(x,\mathcal{T}_s)$
         \Comment{strict JSON obeying \eqref{eq:fidelity}}
  \State $\mathcal{D}_{\mathrm{full}}\gets\mathcal{D}_{\mathrm{full}}\cup\{(x,s,a)\}$
\EndFor
\State $\mathcal{D}_{\mathrm{exp}}\gets\{(x,s,\mathrm{H}(x,a)):(x,s,a)\in
       \mathcal{D}_{\mathrm{full}}^{\,0}\}$
       \Comment{expert-verified subset}
\Statex \textit{// Two-stage training (\Cref{sec:methods-training,sec:methods-sft})}
\State $\theta\gets\arg\min_\theta \mathcal{L}(\theta;\mathcal{D}_{\mathrm{full}})$
       \Comment{pre-train, Eq.~\eqref{eq:train}}
\State $\theta\gets\arg\min_\theta \mathcal{L}(\theta;\mathcal{D}_{\mathrm{exp}})$
       \Comment{fine-tune, Eq.~\eqref{eq:train}}
\Statex \textit{// Structured-prompt inference (\Cref{sec:methods-extension})}
\Function{Generate}{$p$}
  \State $(\hat{s},\hat{a})\gets
    \Phi\left(
      p,\{T_s\}_{s\in\mathcal{S}}
    \right)$
    \Comment{Eq.~\eqref{eq:infer}}
  \State \Return
    $\hat{x}\sim
      G_\theta\left(
        \cdot\mid\sigma(\hat{a})
      \right)$
\EndFunction
\Statex \textit{// Binary-checklist evaluation (\Cref{sec:methods-inhouse,sec:methods-autoeval})}
\Function{Score}{$\hat x,a$}
  \State compile local/global checklists $Q_{\mathrm{loc}},Q_{\mathrm{glo}}$ from $a$
         \Comment{binary yes/no checks}
  \State $c_q\gets\mathbf{1}[J(\hat x,q)=y_q]$ for every check $q$
         \Comment{VLM judge, Eq.~\eqref{eq:verify}}
  \State \Return
    $\bigl(\mathrm{Local},\mathrm{Global},\mathrm{Strict}_\tau\bigr)$
    \Comment{Eqs.~\eqref{eq:localglobal},~\eqref{eq:strict}}
\EndFunction
\end{algorithmic}
\end{algorithm}

\subsection{Model architecture}
\label{sec:methods-model}

\ProjectName{} is built on a unified multimodal backbone~\cite{bagel} that couples a language-model transformer with a visual generation pathway, processing text and image tokens jointly so that the model can consume structured annotations as conditioning text and emit images in the same forward process. We adopt this unified design so that the model that reasons over the physics also renders the diagram.

\subsection{Pre-training}
\label{sec:methods-training}

Pre-training conditions the model to associate physics-structured text with faithful diagram pixels. We train on (structured-annotation, image) pairs drawn from the corpus-level annotation tier (\Cref{tab:data}), using the structured JSON, or a serialized rendering of it, as the conditioning signal. We balance the data mixture across subdisciplines to avoid collapsing onto the most frequent domain, and retain a portion of generic image-text data to preserve the backbone's general rendering ability (text legibility, layout, line art).

\subsection{Supervised fine-tuning}
\label{sec:methods-sft}

We fine-tune the pre-trained backbone on the expert-level subset (\Cref{tab:data}), using the per-subdiscipline schema as the conditioning signal and the paired diagram as the generation target. The dense, physics-grounded conditioning pushes the model toward diagrams whose drawn elements are consistent with the stated physics; the visible/inferred separation lets the model learn the mapping from physical reasoning to the corresponding visual marks; and the shared structure of Steps~1 and~5 across subdisciplines yields cross-domain transfer. We balance the fine-tuning mixture across subdisciplines to counteract the corpus imbalance in \Cref{tab:data}.


\subsection{GenExam evaluation protocol}
\label{sec:methods-eval}

We evaluate on the physics subset of GenExam~\cite{genexam}, following its own subject taxonomy (circuits, electromagnetism, mechanics, optics, quantum mechanics, thermodynamics) and its MLLM-judge scoring and reporting per-subject and aggregate strict/relaxed scores. For \ProjectName{} and for any baseline run in structured mode, the original benchmark prompt is first expanded into the subdiscipline schema before generation: the raw prompt is appended to the structured template (``\emph{Given the following description, fill in the above template structure: \{PROMPT\}}'') and populated by a language model. This isolates the contribution of structured conditioning from the raw-prompt baseline.

\subsection{\BenchName{}}
\label{sec:methods-inhouse}

We construct \BenchName{}, an in-house physics benchmark of $1{,}283$ diagrams drawn as a balanced subset of the held-out test split (\Cref{tab:data}), so that no benchmark item is seen in training. Because each item carries a full structured analysis, we turn that analysis into an item-specific set of checkable questions, rather than scoring a generation with a single holistic judgment.

\paragraph{Two question banks.}
For each item we prepare two banks of \emph{binary} (yes/no) questions, each stored with a gold yes/no answer. The \emph{local} bank is generated by rule directly from the structured fields: entity fields yield existence questions (``Does the image show a normal force on the block?''); typed or directional fields yield relational questions (``Is the relative-motion vector labelled $\vec{v}$?''; ``Is this process segment isobaric?''); and value fields yield value questions (``Is the incline angle labelled $30^\circ$?''). Because these questions enumerate the drawn objects, forces, states and their attributes, an item receives as many local questions as its physics is rich, from roughly a dozen for a single free body to over a hundred for a dense optical figure; empty annotation fields, which the fidelity rules require wherever the diagram is silent, generate no question, so an item is never queried about something its reference does not specify. The \emph{global} bank is a small set of whole-diagram questions (about five per item) generated by a vision--language model to probe overall physical faithfulness rather than individual attributes.

\paragraph{Judging.}
A generated image is scored by a judge vision--language model (GPT-4o). For each item the judge is shown the generated image together with the item's questions, \emph{without} their gold answers, and is instructed to answer every question with exactly ``Yes'' or ``No''; its answers are then compared, question by question, against the gold answers, and an unparseable answer is counted as wrong. Judging is run at temperature $0$ with high image detail. Every question thus returns an individual pass or fail tagged with the physical attribute it concerns, so a model's score decomposes into interpretable per-attribute outcomes rather than a single opaque number.

\paragraph{Scoring modes.}
\emph{Local} and \emph{Global} scoring report the accuracy, that is the fraction of questions answered correctly, over the local and global banks respectively, aggregated per subject and overall (\Cref{tab:results-inhouse-subjects}). Because the local bank enumerates individual physical facts while the global bank asks a few whole-diagram questions, Local measures how many facts are right and is the more demanding of the two. \emph{Strict} scoring instead asks whether an entire item is essentially correct: an item counts as passed only if its fraction of wrong answers does not exceed a per-item tolerance $\tau$, evaluated on the local bank at $\tau \in \{0\%, 5\%, 10\%\}$ (\Cref{tab:results-strict}). At $\tau = 0\%$ every question of the item must be answered correctly, so the strict score is the fraction of items with no error at all, which stays near zero across the field. Raising $\tau$ tolerates a small percentage of wrong answers per item. Note that $\tau$ is a tolerance on the \emph{fraction of wrong answers within an item}, not on any numeric attribute value.

\paragraph{Difficulty strata.}
The number of local questions an item carries is a model-independent proxy for how many objects and steps its diagram contains, and hence for how hard it is to render faithfully. Within each subdiscipline we partition items into Easy, Medium and Hard by the tertiles of this local-question count, and report performance per stratum (\Cref{tab:results-difflevel}).

\subsection{Automatic evaluation reuses the annotation}
\label{sec:methods-autoeval}

The scoring of \Cref{sec:methods-inhouse} is fully automatic and reuses the structured annotation directly, with no human in the loop at scoring time and no separately trained extractor. The local checklist $Q_{\mathrm{loc}}(a)$ is compiled from the annotation by rule (each question's gold answer $y_q$ is fixed by the corresponding field), the global checklist $Q_{\mathrm{glo}}(a)$ is produced by a vision--language model, and a judge model answers both against the generated image, a check passing iff the judge's answer matches the gold (\Cref{eq:verify,eq:localglobal,eq:strict}). Because the serializer $\sigma$ preserves the schema's keys, types and typed values verbatim (\Cref{eq:serialize}), the same annotation that conditions generation is parsed back to build the checklist, so the evaluator needs no bespoke key-value extractor. The visible/inferred split of the schema (\Cref{eq:grounding}) tags each check as grounded or inferred; the current scores weight all local checks equally, but this tagging is what a future weighted or preference-based score could exploit to treat strictly-image-faithful attributes differently from inferred ones.

\section{Structured-annotation templates}
\label{sec:methods-templates}

This appendix gives the full per-subdiscipline annotation templates. All six are reproduced below: mechanics, thermodynamics, acoustics, electromagnetism, physical optics, and quantum mechanics. Each instantiates the unified five-step schema of \Cref{tab:schema} with subdiscipline-specific vocabulary, while Steps~1 and~5 (identification and synthesis-with-assumptions) keep an identical shape across all six, which is the structural sharing exploited in \Cref{sec:disc-why}.

\subsection{Mechanics}
\label{sec:tmpl-mech}

\begin{promptbox}{Template S1. Prompt for mechanics diagram annotation}
You are a physics diagram annotator specializing in mechanics. Your task is to analyse the given diagram and populate the structured schema below.

\textbf{Fidelity rules.} Step~1 (Scene \& Object Identification) and Step~3 (Force Analysis) must be strictly faithful to the image: include only objects, labels, arrows, and forces that are explicitly drawn; do not invent any object, parameter, or force that is not visible. Step~2 (Motion State Analysis), Step~4 (Coordinate System \& Laws), and Step~5 (Synthesis \& Conclusion) may be reasoned from the diagram when they can be logically deduced. Always prefer \texttt{""} / \texttt{[]} over assumptions. All mathematics must be valid \LaTeX{}.

Please output your result in JSON format ONLY, as follows:

\begin{promptjson}
{
  "output": {
    "summary_description": "[Concise summary of the physical scenario, strictly
                            based on the image.]",
    "detailed_description": "[Detailed caption of visible elements and layout.]",
    "detailed_analysis": {
      "step1_scene_and_object_identification": {
        "core_objects": [
          { "object_name": "", "attributes": { "mass": "", "properties": "" } }
        ],
        "environment_and_constraints": ""
      },
      "step2_motion_state_analysis": {
        "system_state": "",
        "motion_description": "",
        "kinematic_variables": { "velocity_v": "", "acceleration_a": "" }
      },
      "step3_force_analysis": [
        { "analyzed_object_name": "",
          "forces": [
            { "force_name": "", "symbol": "", "source": "",
              "direction": "", "magnitude_expression_latex": "" }
          ]
        }
      ],
      "step4_coordinate_system_and_laws": {
        "chosen_coordinate_system": "",
        "governing_physical_law": "",
        "equations": [ { "context_description": "", "equation_latex": "" } ]
      },
      "step5_synthesis_and_conclusion": {
        "key_physical_relationship": "",
        "simplifying_assumptions": []
      }
    }
  }
}
\end{promptjson}

Your response must be ONLY the JSON object, with no additional text, explanations, or commentary.
\end{promptbox}

\subsection{Thermodynamics}
\label{sec:tmpl-thermo}

\begin{promptbox}{Template S2. Prompt for thermodynamics diagram annotation}
You are a physics diagram annotator specializing in thermodynamics. Your task is to analyse the given diagram and populate the structured schema below.

\textbf{Fidelity rules.} Step~2 (State Identification) and the visualized process path in Step~3 must be strictly faithful: include only components (pistons, cylinders), states (points on a $PV$ diagram), state-variable labels ($P_1,V_1,T_1$), process paths (arrows between states), and heat/work indicators ($Q$, $W$) that are explicitly drawn. Step~1 (Scene Identification), the process-type classification in Step~3, Step~4 (Principles \& Equations), and Step~5 (Synthesis) may be inferred (e.g.\ classify a horizontal $PV$ segment as isobaric, apply the corresponding law). All mathematics must be valid \LaTeX{}.

Please output your result in JSON format ONLY, as follows:

\begin{promptjson}
{
  "caption": {
    "category": "Thermodynamics",
    "summary_description": "[Concise summary, strictly based on the image.]",
    "detailed_description": "[Detailed description: diagram type (e.g. P-V),
      labeled axes, state points, process paths, schematics.]",
    "detailed_analysis": {
      "step1_scene_identification_and_system_definition": {
        "physical_domain": "Thermodynamics",
        "system_definition": "[e.g. n moles of ideal monatomic gas in a cylinder.]",
        "problem_type": "[Single Process / Cycle / Heat Engine Schematic]"
      },
      "step2_state_identification_and_parameterization": {
        "thermodynamic_states": [
          { "state_name": "[A, 1, ...]",
            "state_variables": [
              { "symbol_latex": "[P_A, V_A, T_A]", "description": "",
                "value_latex": "" } ] }
        ]
      },
      "step3_process_and_interaction_analysis": {
        "processes": [
          { "process_name": "[A -> B]",
            "process_type": "[Isobaric/Isothermal/Adiabatic/Isochoric ...]",
            "energy_transfers": [
              { "type": "[Heat/Work]", "direction": "[Into/Out of system ...]",
                "symbol_latex": "[Q_{AB}, W_{AB}]" } ] }
        ]
      },
      "step4_principles_and_equation_formulation": {
        "governing_physical_laws": [ "[First Law, Ideal Gas Law, ...]" ],
        "equations": [ { "context_description": "",
                         "equation_latex": "\\Delta U_{AB} = Q_{AB} - W_{AB}" } ]
      },
      "step5_synthesis_and_assumption_declaration": {
        "key_findings_or_relationships": "[net work per cycle, efficiency, ...]",
        "simplifying_assumptions": [ "[ideal gas, quasi-static, frictionless ...]" ]
      }
    }
  }
}
\end{promptjson}

Your response must be ONLY the JSON object, with no additional text, explanations, or commentary.
\end{promptbox}

\subsection{Acoustics}
\label{sec:tmpl-acou}

\begin{promptbox}{Template S3. Prompt for acoustics diagram annotation}
You are a physics diagram annotator specializing in acoustics. Your task is to analyse the given diagram and populate the structured schema below.

\textbf{Fidelity rules.} Step~1 (Scene Identification) and the system components in Step~2 must be strictly faithful to the image: include only sources, observers, media, boundaries, wavefronts, and motion arrows that are explicitly drawn. Step~3 (Geometry \& Wave Analysis), Step~4 (Principles \& Equations), and Step~5 (Synthesis) may be inferred when they can be logically deduced from the drawn elements. Always prefer \texttt{""} / \texttt{[]} over assumptions. All mathematics must be valid \LaTeX{}.

Please output your result in JSON format ONLY, as follows:

\begin{promptjson}
{
  "caption": {
    "category": "Acoustics",
    "summary_description": "[Concise summary, strictly based on the image.]",
    "detailed_description": "[source, observer, medium, wavefronts, motion arrows.]",
    "detailed_analysis": {
      "step1_scene_identification_and_phenomenon_judgment": {
        "physical_domain": "Acoustics",
        "sub_domain": "[Wave Propagation / Interference / ...]",
        "phenomenon_type": "[Doppler / Standing Wave / Beats / ...]"
      },
      "step2_system_component_identification_and_parameterization": {
        "system_components": [
          { "name": "[S, O]", "type": "[Source/Medium/Observer/Boundary]",
            "parameters": [ { "symbol_latex": "[f_s, v_o, c, L]",
                              "description": "", "value_latex": "" } ] }
        ]
      },
      "step3_geometry_and_wave_analysis": {
        "spatial_arrangement": { "description": "",
          "relative_motion_vector_latex": "" },
        "wave_properties": { "wave_type": "[Transverse/Longitudinal]",
          "wave_form": "[Plane/Spherical/Standing]",
          "wavelength_expression_latex": "\\lambda = c/f_s" }
      },
      "step4_principles_and_equation_formulation": {
        "coordinate_system_and_conventions": { "description": "" },
        "governing_physical_principles": [ "[Doppler Effect / Superposition ...]" ],
        "equations": [ { "context_description": "",
          "equation_latex": "f_o = f_s \\left( \\frac{c + v_o}{c} \\right)" } ]
      },
      "step5_synthesis_and_assumption_declaration": {
        "key_findings_or_relationships": "",
        "simplifying_assumptions": [ "[non-dispersive medium, point source ...]" ]
      }
    }
  }
}
\end{promptjson}

Your response must be ONLY the JSON object, with no additional text, explanations, or commentary.
\end{promptbox}

\subsection{Electromagnetism}
\label{sec:tmpl-em}

The electromagnetism schema must cover two structurally different families of diagram: lumped-element circuits and continuous field configurations. Rather than splitting the subdiscipline, Step~3 offers both a \texttt{circuit\_topology} and a \texttt{field\_distribution} block, and the annotator populates whichever the diagram supports, leaving the other empty. Step~1 records the family explicitly (\texttt{sub\_domain}, \texttt{problem\_type}) so that the choice is auditable.

\begin{promptbox}{Template S4. Prompt for electromagnetism diagram annotation}
You are a physics diagram annotator specializing in electromagnetism. Your task is to analyse the given diagram and populate the structured schema below.

\textbf{Fidelity rules.} Step~2 (Component/Source Identification) and the structural parts of Step~3 (Circuit Topology / Field Distribution) must be strictly faithful to the image: include only components, sources, labels, value annotations, arrows (e.g.\ for current), and connections that are explicitly drawn, and do not invent any component, parameter, or structural relationship that is not visible. Step~1 (Scene Identification), Step~4 (Principles \& Equations), and Step~5 (Synthesis) require expert physical reasoning derived from the visible elements. In Step~3, fill in either the circuit topology or the field distribution according to the problem type, and leave the other empty. All mathematics must be valid \LaTeX{}, and missing information must be left as \texttt{""} / \texttt{[]} rather than guessed.

Please output your result in JSON format ONLY, as follows:

\begin{promptjson}
{
  "caption": {
    "category": "Electromagnetism",
    "summary_description": "[Concise summary, strictly based on the image.]",
    "detailed_description": "[Component layout, labels, symbols, current-direction
      arrows, field-line distributions.]",
    "detailed_analysis": {
      "step1_scene_identification_and_type_judgment": {
        "physical_domain": "Electromagnetism",
        "sub_domain": "[Circuit Theory / Electrostatics / Magnetostatics /
                        Electrodynamics]",
        "problem_type": "[e.g. DC RC transient, AC RLC series, point charge in a
                          uniform E-field, Ampere's law for a solenoid]"
      },
      "step2_component_source_identification_and_parameterization": {
        "components_and_sources": [
          { "name": "[R1, C1, q1 ...]",
            "type": "[Resistor / Capacitor / Voltage Source / Point Charge /
                      Current-Carrying Wire]",
            "parameters": [
              { "symbol_latex": "[R_1, C_1, V_s, q_1]", "description": "",
                "value_latex": "[e.g. 100\\Omega, 10\\mu F, U_0\\cos(\\omega t)]" } ] }
        ]
      },
      "step3_structural_topology_and_state_analysis": {
        "//": "Fill in either the circuit topology or the field distribution.",
        "circuit_topology": {
          "connection_type": "[Series / Parallel / Bridge]",
          "nodes": "[e.g. ['A', 'B', 'Ground']]",
          "loops": "[e.g. ['Loop 1: V_s-R1-C1', ...]]"
        },
        "field_distribution": {
          "source_geometry": "[Point / Infinite Line / Infinite Plane]",
          "field_geometry": "[Radial / Uniform / Cylindrically Symmetric]"
        },
        "system_state": "[DC Steady-State / AC Steady-State / Transient /
                          Electrostatic Equilibrium / Magnetostatic]"
      },
      "step4_inference_of_principles_and_equation_formulation": {
        "coordinate_system_and_conventions": {
          "spatial_coordinate_system": "[Cartesian / Cylindrical / Spherical]",
          "circuit_conventions": "[Passive sign convention, assumed current
                                   directions for KCL]"
        },
        "governing_physical_laws": [
          "[Ohm's Law, KVL, KCL, Gauss's Law, Ampere's Law, Lorentz Force Law ...]"
        ],
        "equations": [
          { "context_description": "[e.g. Applying KVL to Loop 1]",
            "equation_latex": "\\oint \\vec{E} \\cdot d\\vec{A} =
                               \\frac{Q_{enc}}{\\epsilon_0}" }
        ]
      },
      "step5_synthesized_output_and_assumption_declaration": {
        "key_findings_or_relationships": "[e.g. resonance frequency of the circuit,
          electric field expression at a specific point]",
        "simplifying_assumptions": [
          "[ideal wire (zero resistance), ideal voltage source, uniform magnetic
            field, quasi-static approximation, no fringing fields ...]" ]
      }
    }
  }
}
\end{promptjson}

Your response must be ONLY the JSON object, with no additional text, explanations, or commentary.
\end{promptbox}

\subsection{Physical optics}
\label{sec:tmpl-optics}

The optics schema targets \emph{physical} optics: interference, diffraction, and polarization. Its distinguishing element is Step~3, which records the optical path explicitly, since the physics of these diagrams is carried by path and phase differences rather than by forces or states. Step~4 accordingly replaces a flat list of equations with typed \texttt{phenomenon\_conditions} (maxima, minima, transmission), each carrying its own condition equation and the range of its order index.

\begin{promptbox}{Template S5. Prompt for physical-optics diagram annotation}
You are a physics diagram annotator specializing in physical optics. Your task is to analyse the given diagram and populate the structured schema below.

\textbf{Fidelity rules.} Step~2 (System Component Identification) and the geometric layout in Step~3 (Optical Path Analysis) must be strictly faithful to the image: include only components (light sources, slits, gratings, screens), labels ($\lambda, d, a, L, \theta$), and drawn ray paths that are explicitly shown, and do not invent any component or parameter that is not visible. Step~1 (Phenomenon Judgment), Step~4 (Principles \& Equations), and Step~5 (Synthesis) require expert physical reasoning derived from the visible elements. All mathematics must be valid \LaTeX{}, and missing information must be left as \texttt{""} / \texttt{[]} rather than guessed.

Please output your result in JSON format ONLY, as follows:

\begin{promptjson}
{
  "caption": {
    "category": "Physical Optics",
    "summary_description": "[Concise summary, strictly based on the image.]",
    "detailed_description": "[Light source, aperture/element type, observation
      screen, labeled distances and angles.]",
    "detailed_analysis": {
      "step1_scene_identification_and_phenomenon_judgment": {
        "physical_domain": "Optics",
        "sub_domain": "Physical Optics",
        "phenomenon_type": "[Interference / Diffraction / Polarization /
                             Combined Interference and Diffraction]"
      },
      "step2_system_component_identification_and_parameterization": {
        "system_components": [
          { "name": "[S1, G, P1 ...]",
            "type": "[Light Source / Single Slit / Double Slit / Diffraction
                      Grating / Polarizer / Analyzer / Observation Screen]",
            "parameters": [
              { "symbol_latex": "[\\lambda, d, a, L, N]",
                "description": "[wavelength, slit separation, slit width,
                                 distance to screen, lines per unit length]",
                "value_latex": "" } ] }
        ]
      },
      "step3_optical_path_and_phase_analysis": {
        "geometric_layout": {
          "description": "[Path of light through the system, e.g. a plane wave is
            incident normally on a grating and observed at an angle theta.]"
        },
        "path_and_phase_difference": {
          "context": "[What causes the difference, e.g. path difference between
                       waves from two adjacent slits.]",
          "path_difference_latex": "\\Delta x = d\\sin(\\theta)",
          "phase_difference_latex": "\\Delta\\phi =
                                     \\frac{2\\pi}{\\lambda}d\\sin(\\theta)"
        }
      },
      "step4_principles_and_equation_formulation": {
        "governing_physical_principles": [
          "[Huygens' Principle, Principle of Superposition, Malus's Law ...]"
        ],
        "phenomenon_conditions": [
          { "condition_type": "[Constructive Interference (Maxima) / Destructive
              Interference (Minima) / Diffraction Minima / Polarization
              Transmission]",
            "equation_latex": "[e.g. d\\sin(\\theta) = m\\lambda,
                                a\\sin(\\theta) = m\\lambda,
                                I = I_0\\cos^2(\\theta)]",
            "variable_definitions": "[e.g. m = 0, \\pm 1, \\pm 2, ...]" }
        ]
      },
      "step5_synthesis_and_assumption_declaration": {
        "key_findings_or_relationships": "[e.g. angular separation of bright
          fringes, fringe spacing on the screen (y_m).]",
        "simplifying_assumptions": [
          "[Fraunhofer (far-field) approximation (L >> a), paraxial (small-angle)
            approximation, monochromatic and coherent source, infinitesimally
            narrow slits ...]" ]
      }
    }
  }
}
\end{promptjson}

Your response must be ONLY the JSON object, with no additional text, explanations, or commentary.
\end{promptbox}

\subsection{Quantum mechanics}
\label{sec:tmpl-qm}

Quantum mechanics is the subdiscipline whose diagrams are least uniform: a potential well, an energy-level diagram, a photoelectric-effect apparatus, and a Stern--Gerlach setup share almost no visual vocabulary. The schema absorbs this by declaring the formalism up front (Step~1: \texttt{formalism\_type}, \texttt{scenario\_type}, \texttt{degrees\_of\_freedom}) and by giving Step~3 two alternative state descriptions, spatial boundary conditions for wave-mechanics problems and a basis with eigenstates for spin and atomic problems. Step~4 separates evolution equations from measurement equations, so that dynamics and the Born rule are recorded as distinct fields.

\begin{promptbox}{Template S6. Prompt for quantum-mechanics diagram annotation}
You are a physics diagram annotator specializing in quantum mechanics. Your task is to analyse the given diagram and populate the structured schema below.

\textbf{Fidelity rules.} Step~2 (Component Parameterization) and the geometric definitions in Step~3 (State Analysis) must be strictly faithful to the image: record potential profiles $V(x)$ (steps, wells), explicit labels ($V_0, a, L, \Phi, n$), specific ket vectors ($|{+}z\rangle$), and experimental apparatus (Stern--Gerlach magnets, circuits) exactly as drawn. Do not invent numerical values or components if only variables or abstract symbols are shown. Step~1 (Scene Identification), Step~4 (Equation Formulation), and Step~5 (Synthesized Output) require expert physical reasoning derived from the visible elements. All mathematics must be valid \LaTeX{}, and missing information must be left as \texttt{""} / \texttt{[]} rather than guessed.

Please output your result in JSON format ONLY, as follows:

\begin{promptjson}
{
  "caption": {
    "category": "Quantum Mechanics",
    "summary_description": "[Concise summary, strictly based on the image, e.g.
      energy-level transitions in a hydrogen atom generating an emission
      spectrum, or light incident on a metal plate demonstrating the
      photoelectric effect.]",
    "detailed_description": "[Potential shapes V(x), orbitals, incoming photons,
      energy-level diagrams, experimental circuits, Stern-Gerlach magnets.]",
    "detailed_analysis": {
      "step1_scene_identification": {
        "physical_domain": "Quantum Mechanics",
        "formalism_type": "[Wave Mechanics / Matrix Mechanics /
                            Phenomenological (Energy Balance)]",
        "scenario_type": "[Potential Well / Hydrogen Atom / Photoelectric Effect /
                           EM Wave-Photon / Spin System]",
        "degrees_of_freedom": "[1D Spatial / 3D Spherical / Spin /
                                Photon Frequency]"
      },
      "step2_component_parameterization": {
        "interaction_landscape_and_hamiltonian": {
          "description": "[Environment/components, e.g. Coulomb potential
            V(r)=-ke^2/r, metal surface with work function Phi, external field B]",
          "core_operator_or_energy_term_latex": "[e.g. \\hat{H} =
            -\\frac{\\hbar^2}{2m}\\nabla^2 + V(r), E_k = hf - \\Phi]"
        },
        "system_parameters": [
          { "symbol_latex": "[V_0, a, \\lambda, \\Phi, n, m]",
            "description": "[barrier height, well width, wavelength, work
                             function, principal quantum number, mass]",
            "value_from_diagram_latex": "[Value if explicit, else empty]" }
        ]
      },
      "step3_state_analysis": {
        "//": "Define the state space: spatial regions (wells) OR basis states
               (spins/atoms).",
        "spatial_boundary_conditions": {
          "regions": "[e.g. Region I (x<0) V=0, Region II (0<x<L) V=0]",
          "boundary_equations_latex": "[e.g. \\psi(0) = 0, \\psi(L) = 0]"
        },
        "basis_and_eigenstates": {
          "basis_description": "[e.g. hydrogen orbitals |n,l,m>, spin z-basis]",
          "initial_state_vector_latex": "[e.g. |\\Psi(0)\\rangle =
            \\frac{1}{\\sqrt{2}}(|+\\rangle + |-\\rangle)]"
        }
      },
      "step4_equation_formulation": {
        "governing_laws": "[Schrodinger Equation, Born Rule, Energy Conservation]",
        "dynamic_equations_latex": [
          { "name": "Evolution/State Equation",
            "equation": "[e.g. i\\hbar \\frac{\\partial}{\\partial t}\\Psi =
                          \\hat{H}\\Psi, E_n = \\hbar\\omega(n+1/2)]" }
        ],
        "measurement_equations_latex": [
          { "name": "Observable/Probability",
            "equation": "[e.g. \\hat{S}_z |\\pm\\rangle =
              \\pm \\frac{\\hbar}{2} |\\pm\\rangle,
              P(E_n) = |\\langle n|\\Psi \\rangle|^2]" }
        ]
      },
      "step5_synthesized_output": {
        "key_phenomena_and_results": "[e.g. quantized energy spectrum, tunneling
          probability T>0, photoelectric emission occurs if hf > Phi]",
        "simplifying_assumptions": [
          "[non-relativistic approximation, time-independent potential, ideal
            photon gas, infinite potential walls ...]" ]
      }
    }
  }
}
\end{promptjson}

Your response must be ONLY the JSON object, with no additional text, explanations, or commentary.
\end{promptbox}

\subsection{A gallery of structured annotations}
\label{sec:methods-samples}

Samples~S1--S18 instantiate the annotation framework on eighteen diagrams drawn from the corpus, three per subdiscipline, selected to span distinct diagram genres within each domain (for example, an inclined-plane free-body construction, a rolling-with-slipping analysis and a two-body pulley system in mechanics, or a Carnot and an Otto cycle in thermodynamics) rather than to flatter the model. Each card juxtaposes the reference diagram with a condensed rendering of its expert-verified structured analysis, organised along the five schema steps of \Cref{tab:schema}; all mathematical content is transcribed verbatim from the annotation, in the symbolic form the schema mandates, so that governing relations such as $Ma_1 = Mg\sin\theta - T - F$ or $d\sin\theta = m\lambda$ appear exactly as they are stored.

The cards make three properties of the framework visible at once. First, the two steps that each subdiscipline designates as \emph{strictly faithful to the image} (\Cref{tab:schema}) are marked with a dagger ($^\dagger$); these fields report only what is drawn, and reading them against the adjacent diagram is precisely the verification the fidelity rules are designed to support, whereas the unmarked steps carry the physical reasoning inferred from those grounded observations. Second, the identification and synthesis steps retain an identical role across all six domains, so that heterogeneous diagrams (a pulley on an incline, a crossed-field velocity selector, a spherical-mirror ray construction, a $P$--$V$ cycle, a pipe resonance, a hydrogenic radial distribution) are reduced to a single, uniform representation. Third, because every field is a typed key--value entry rather than free-form prose, each card is machine-parseable in the same form consumed during training and reused, per item, to derive the questions \BenchName{} scores a generation against (\Cref{sec:res-inhouse}). Together the eighteen samples illustrate how the schema converts a flat image--caption pair into an auditable chain of physical reasoning, uniformly across the six subdisciplines.

\newcommand{\samplebody}[8]{%
  \begin{minipage}[t]{0.40\linewidth}%
    \vspace{0pt}\centering
    \includegraphics[width=\linewidth]{#1}\\[2pt]
    {\scriptsize\itshape #2}%
  \end{minipage}\hfill
  \begin{minipage}[t]{0.575\linewidth}%
    \vspace{0pt}%
    \sampleeyebrow{#3}#4#5#6#7#8%
  \end{minipage}%
}

\begin{samplebox}{Sample S1\quad Mechanics}
\samplebody{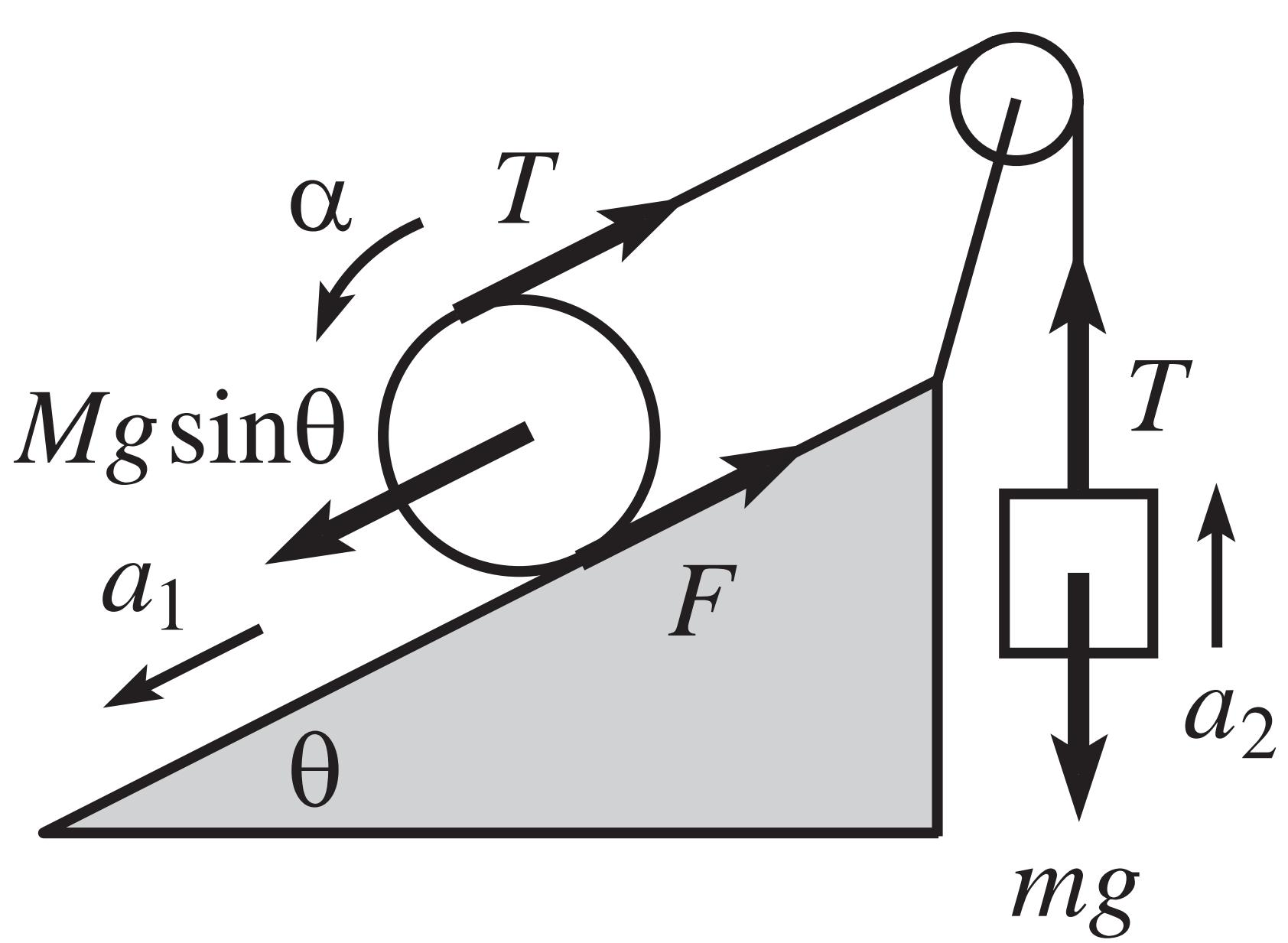}{Reference diagram from the corpus.}%
{Fundamentals of mechanics}
{\stepline{1~Scenario$^\dagger$}{A block of mass $M$ on an incline of angle $\theta$, roped over a pulley to a hanging mass $m$.}}
{\stepline{2~Parameters}{Dynamic: the block accelerates down-slope ($a_1$), the hanging mass upward ($a_2$); the pulley has angular acceleration $\alpha$.}}
{\stepline{3~Structure$^\dagger$}{On the block, $Mg\sin\theta$ down-slope, tension $T$ up-slope, friction $F$ up-slope; on the mass, weight $mg$ down and tension $T$ up.}}
{\stepline{4~Laws}{Newton's second law: $Ma_1 = Mg\sin\theta - T - F$ and $ma_2 = T - mg$.}}
{\stepline{5~Synthesis}{The motion is fixed by the force balance and the rope tension. \emph{Assumes} a frictionless pulley and negligible air resistance.}}
\end{samplebox}
\medskip

\begin{samplebox}{Sample S2\quad Mechanics}
\samplebody{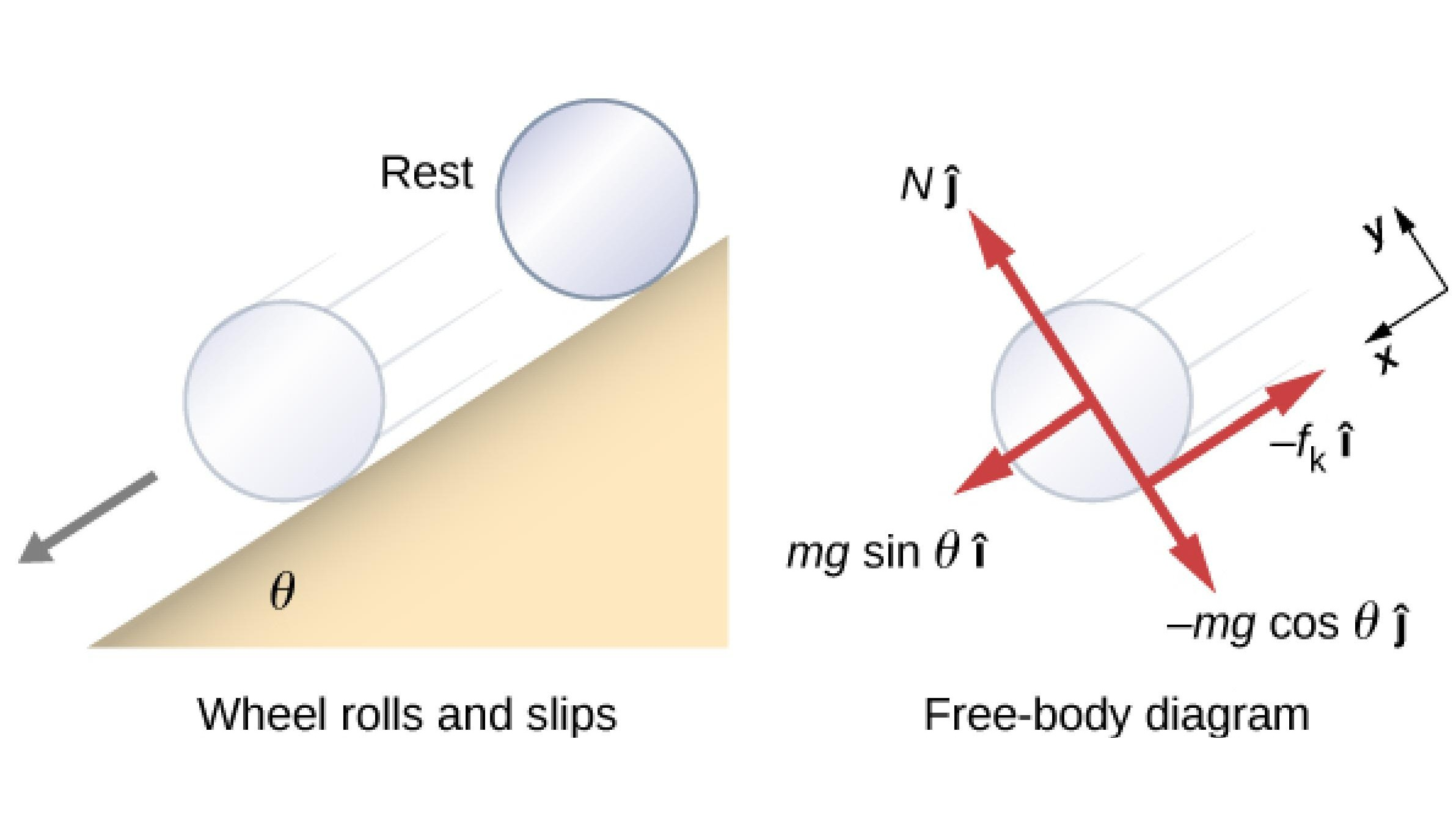}{Reference diagram from the corpus.}%
{Fundamentals of mechanics}
{\stepline{1~Scenario$^\dagger$}{A wheel of mass $m$ rolling and slipping down an incline of angle $\theta$, with its free-body diagram.}}
{\stepline{2~Parameters}{The wheel translates down the incline while slipping (kinetic-friction regime).}}
{\stepline{3~Structure$^\dagger$}{Normal force $N\,\hat{\jmath}$, gravity components $mg\sin\theta\,\hat{\imath}$ (down-slope) and $-mg\cos\theta\,\hat{\jmath}$, kinetic friction $-f_k\,\hat{\imath}$.}}
{\stepline{4~Laws}{Newton's second law along the incline: $ma = mg\sin\theta - f_k$.}}
{\stepline{5~Synthesis}{The motion follows from gravity, the normal force and friction. \emph{Assumes} kinetic friction due to slipping and a rigid incline.}}
\end{samplebox}
\medskip

\begin{samplebox}{Sample S3\quad Mechanics}
\samplebody{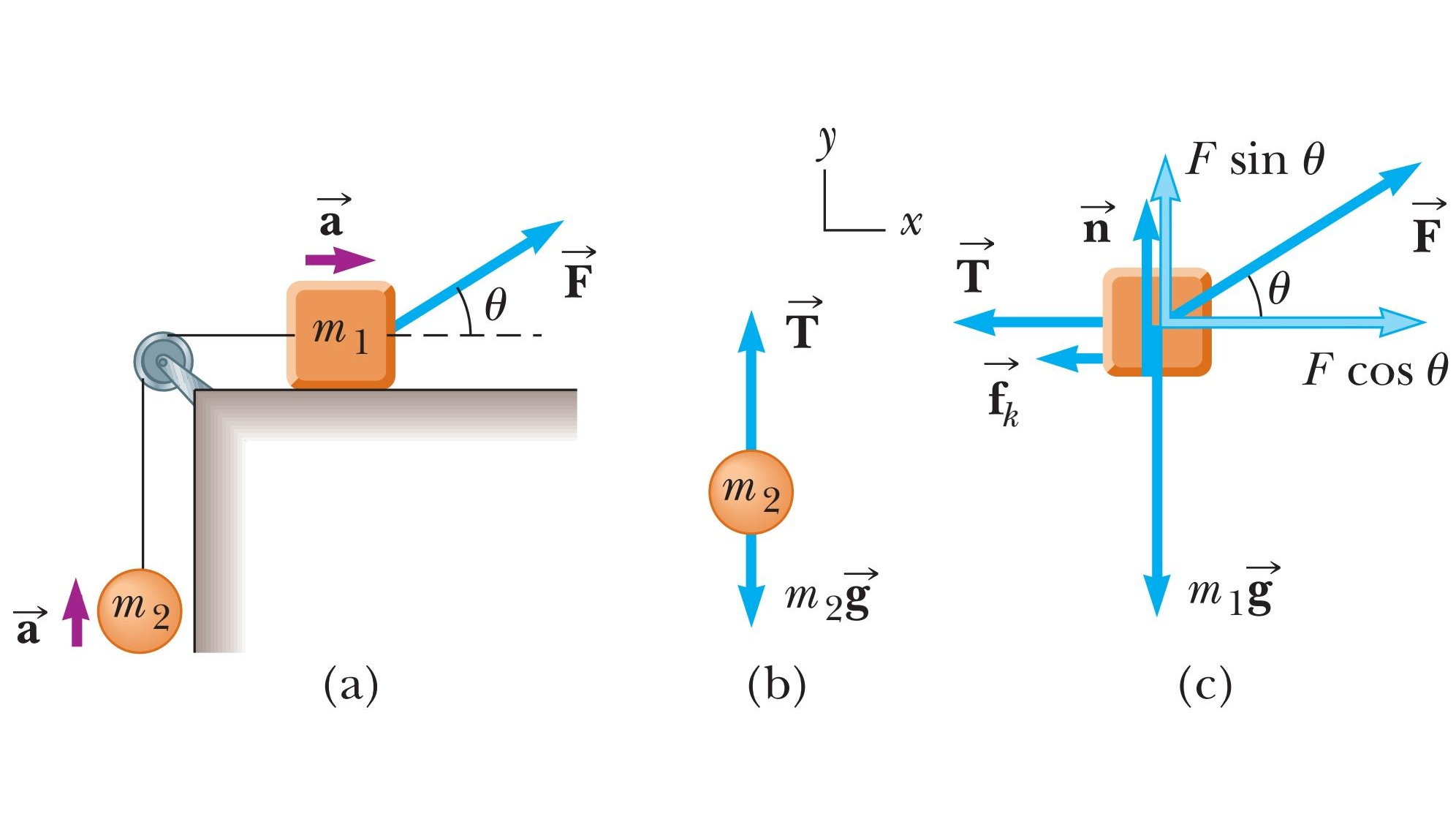}{Reference diagram from the corpus.}%
{Fundamentals of mechanics}
{\stepline{1~Scenario$^\dagger$}{Two masses: $m_1$ on a horizontal surface pulled by a force $\vec{F}$ at angle $\theta$, roped over a pulley to a hanging mass $m_2$.}}
{\stepline{2~Parameters}{Dynamic: $m_1$ moves horizontally and $m_2$ vertically, with acceleration $\vec{a}$.}}
{\stepline{3~Structure$^\dagger$}{On $m_1$: applied $\vec{F}$, normal $\vec{n}$ (up), kinetic friction $\vec{f_k}$ (opposing motion), weight $m_1\vec{g}$ (down); on $m_2$: tension $\vec{T}$ (up), weight $m_2\vec{g}$ (down).}}
{\stepline{4~Laws}{Newton's second law: $m_1a = F\cos\theta - f_k$, $n = m_1g - F\sin\theta$, and $m_2a = T - m_2g$.}}
{\stepline{5~Synthesis}{The acceleration follows from the net force on both masses. \emph{Assumes} a frictionless massless pulley and an inextensible string.}}
\end{samplebox}
\medskip

\begin{samplebox}{Sample S4\quad Electromagnetism}
\samplebody{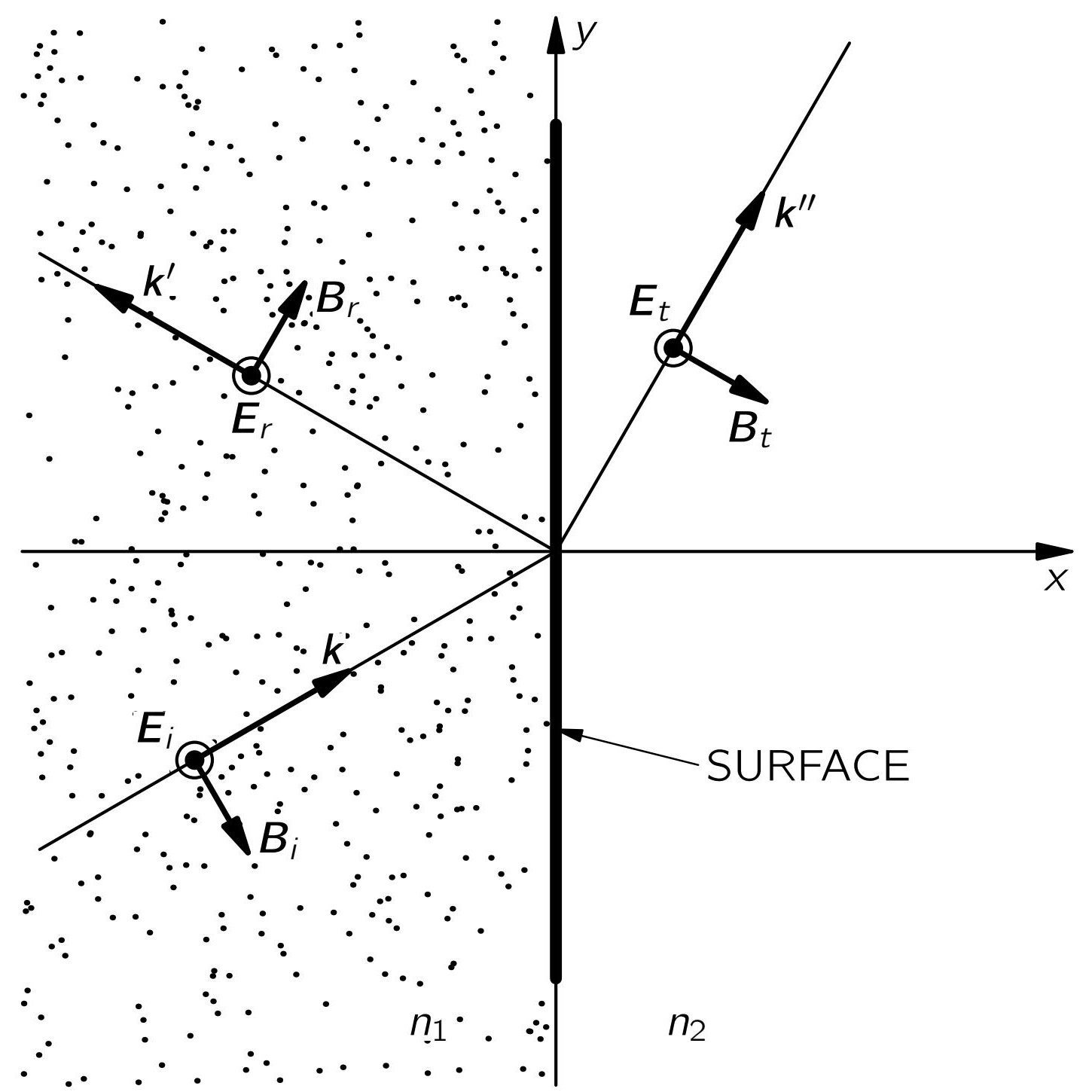}{Reference diagram from the corpus.}%
{Electrodynamics}
{\stepline{1~Scenario}{A plane electromagnetic wave reflected and transmitted at a boundary between media of index $n_1$ and $n_2$.}}
{\stepline{2~Parameters$^\dagger$}{Incident $(E_i,B_i,\vec{k})$, reflected $(E_r,B_r,\vec{k}')$ and transmitted $(E_t,B_t,\vec{k}'')$ fields drawn at the interface.}}
{\stepline{3~Structure$^\dagger$}{Plane-wave field geometry: reflection and transmission across the surface, in Cartesian $x$--$y$ coordinates.}}
{\stepline{4~Laws}{Snell's law $n_1\sin\theta_i = n_2\sin\theta_t$, with the electromagnetic boundary conditions at the interface.}}
{\stepline{5~Synthesis}{Fixes the angles and amplitudes of the reflected and transmitted waves. \emph{Assumes} an ideal plane wave and no absorption.}}
\end{samplebox}
\medskip

\begin{samplebox}{Sample S5\quad Electromagnetism}
\samplebody{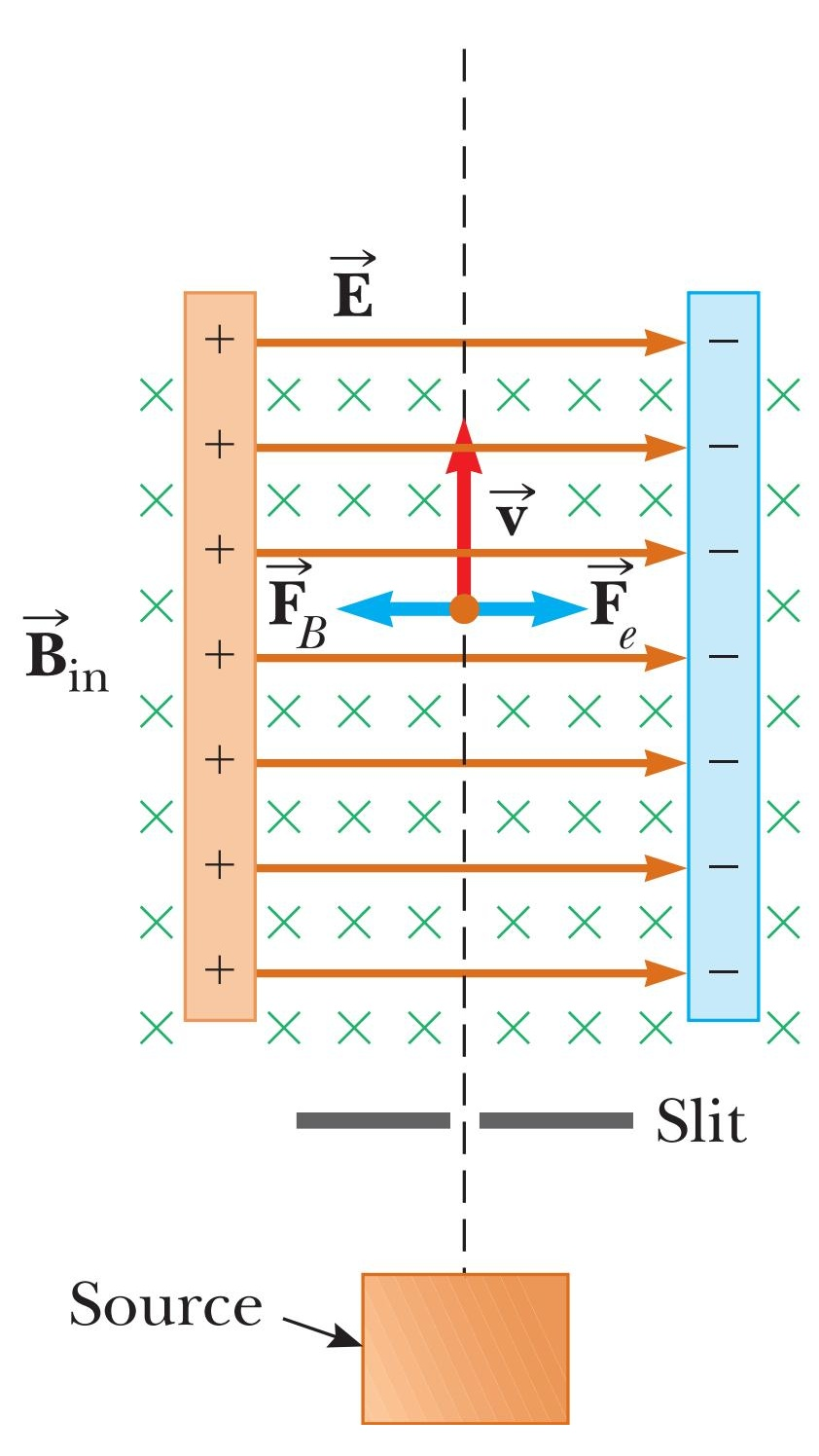}{Reference diagram from the corpus.}%
{Electrodynamics}
{\stepline{1~Scenario}{A charged particle moving through crossed (perpendicular) electric and magnetic fields --- a velocity selector.}}
{\stepline{2~Parameters$^\dagger$}{A uniform field $\vec{E}$ between charged plates and a uniform $\vec{B}_{\mathrm{in}}$ into the page; particle velocity $\vec{v}$.}}
{\stepline{3~Structure$^\dagger$}{Uniform fields in the region; the electric force $\vec{F}_e$ and magnetic force $\vec{F}_B$ act oppositely on the moving charge.}}
{\stepline{4~Laws}{Lorentz force with the balance $\vec{F}_e+\vec{F}_B=0$, i.e.\ $qE = qvB$.}}
{\stepline{5~Synthesis}{Selects the speed $v = E/B$ at which the forces balance. \emph{Assumes} uniform fields and negligible gravity.}}
\end{samplebox}
\medskip

\begin{samplebox}{Sample S6\quad Electromagnetism}
\samplebody{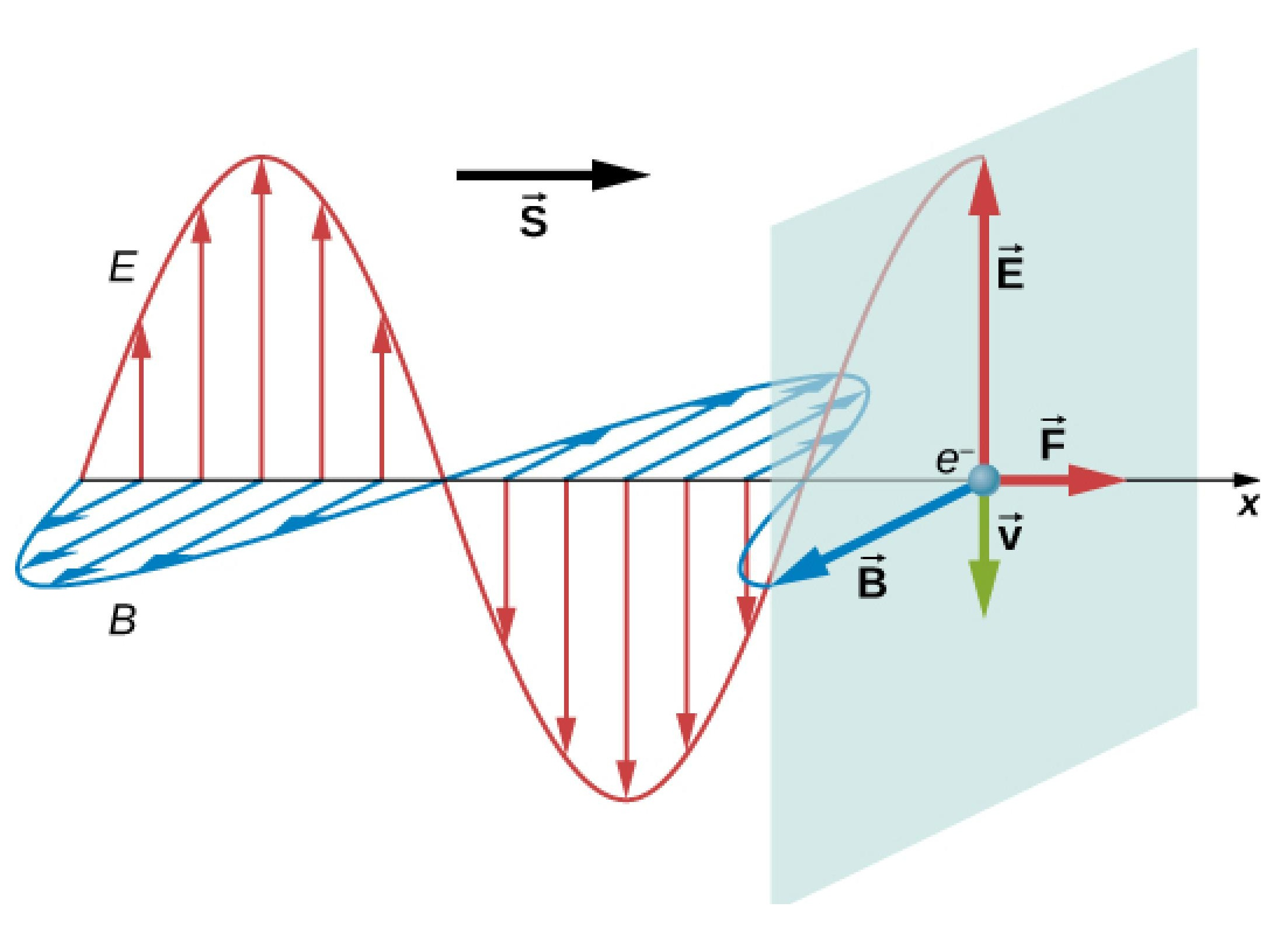}{Reference diagram from the corpus.}%
{Electrodynamics}
{\stepline{1~Scenario}{A plane electromagnetic wave propagating along $x$ (Poynting vector $\vec{S}$), with an electron placed in its fields.}}
{\stepline{2~Parameters$^\dagger$}{Transverse electric field $\vec{E}$ and magnetic field $\vec{B}$; electron of charge $q$ moving with velocity $\vec{v}$.}}
{\stepline{3~Structure$^\dagger$}{A transverse plane wave with $\vec{E}$ and $\vec{B}$ mutually perpendicular; the electron experiences a force $\vec{F}$.}}
{\stepline{4~Laws}{Maxwell's equations and the Lorentz force $\vec{F} = q(\vec{E} + \vec{v}\times\vec{B})$.}}
{\stepline{5~Synthesis}{The electron is driven by the wave's fields. \emph{Assumes} a plane wave and non-relativistic motion.}}
\end{samplebox}
\medskip

\begin{samplebox}{Sample S7\quad Physical optics}
\samplebody{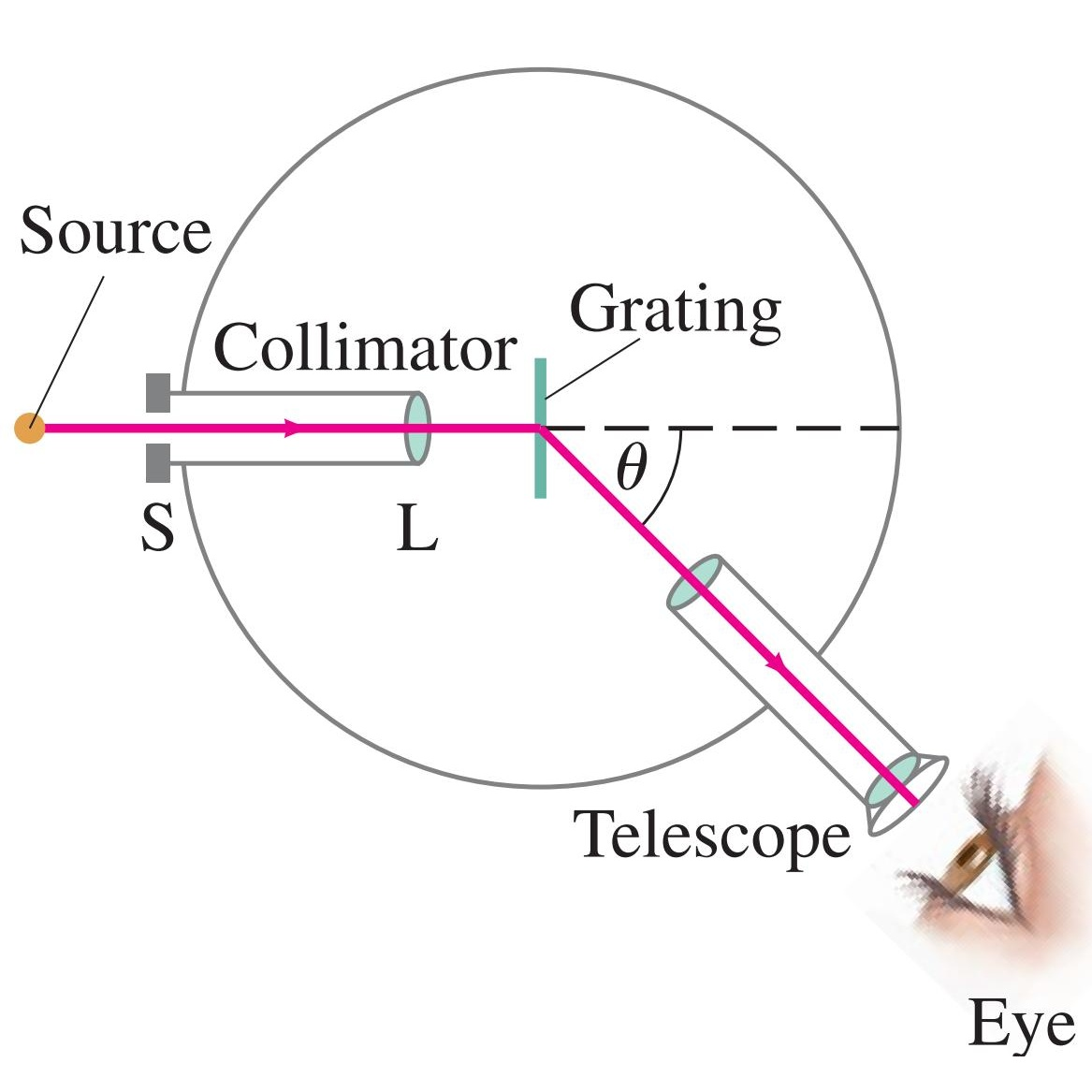}{Reference diagram from the corpus.}%
{Diffraction-grating spectrometer}
{\stepline{1~Scenario}{A slit source $S$, a collimator lens $L$, a transmission grating, a telescope and the eye, along an optical axis (dashed).}}
{\stepline{2~Parameters$^\dagger$}{Components as labelled; one diffracted order leaves the grating at angle $\theta$ to the axis.}}
{\stepline{3~Structure$^\dagger$}{Source $\to$ collimator $\to$ collimated plane wave $\to$ grating (diffracts) $\to$ order at $\theta$ $\to$ telescope $\to$ eye.}}
{\stepline{4~Laws}{Grating equation $d\sin\theta = m\lambda$; thin-lens collimation $\tfrac{1}{f}=\tfrac{1}{s}+\tfrac{1}{s'}$ with $s\approx f$.}}
{\stepline{5~Synthesis}{The collimator forms a plane wave that the grating angularly disperses by wavelength. \emph{Assumes} the source at the focal plane and near-normal incidence.}}
\end{samplebox}
\medskip

\begin{samplebox}{Sample S8\quad Physical optics}
\samplebody{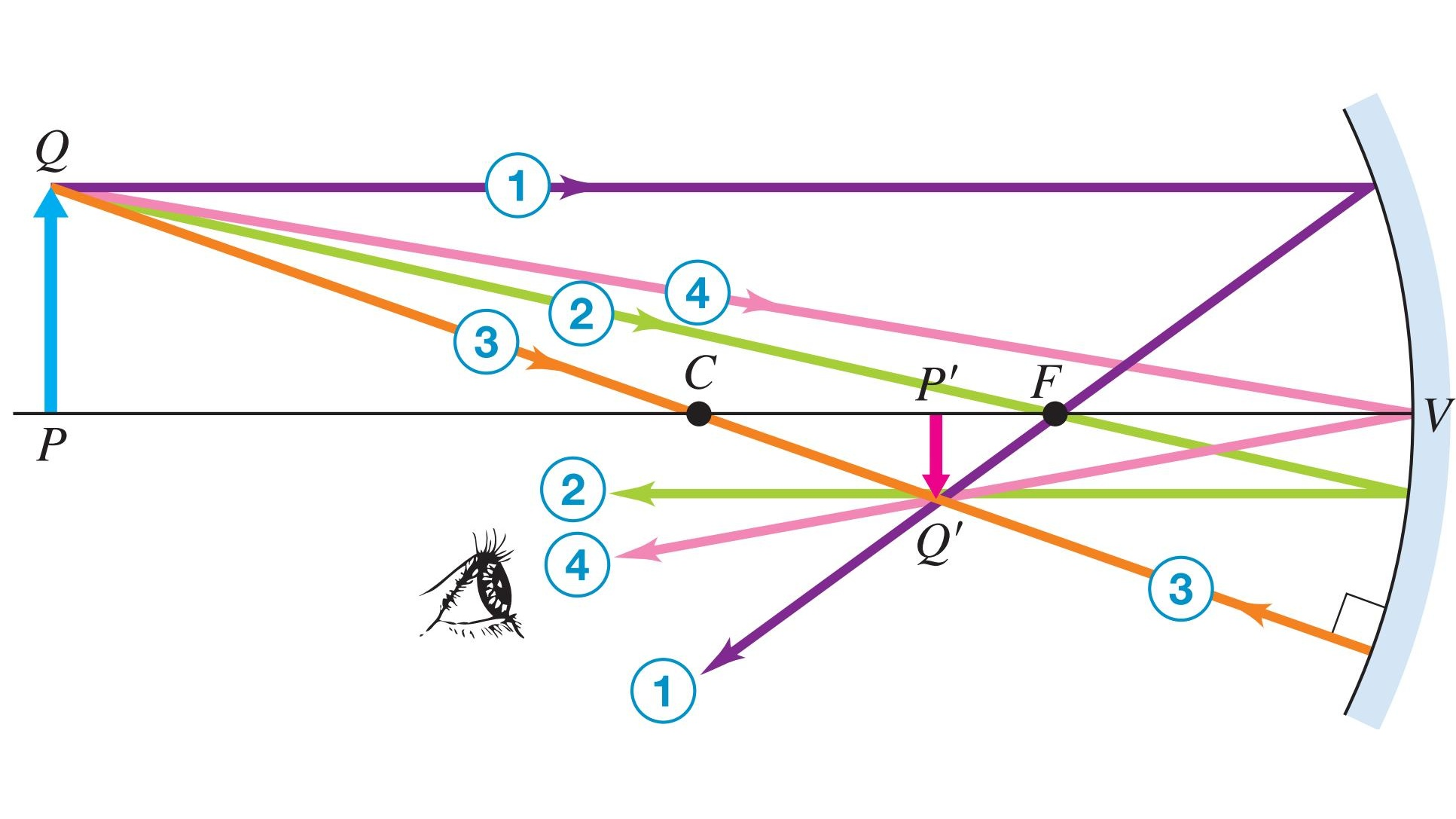}{Reference diagram from the corpus.}%
{Geometric optics: spherical mirror}
{\stepline{1~Scenario}{Image formation by a concave spherical mirror: an upright object $PQ$, with vertex $V$, focus $F$ and centre of curvature $C$.}}
{\stepline{2~Parameters$^\dagger$}{Object point $Q$ (base $P$) and image point $Q'$ (top $P'$); principal points $V,F,C$ and construction rays 1--4.}}
{\stepline{3~Structure$^\dagger$}{Rays from $Q$ reflect at the mirror and converge to $Q'$, forming an inverted real image below the axis.}}
{\stepline{4~Laws}{Law of reflection; the parallel-, focal- and centre-rays fix the image point.}}
{\stepline{5~Synthesis}{All rays from one object point meet at one image point. \emph{Assumes} paraxial rays and an ideal mirror.}}
\end{samplebox}
\medskip

\begin{samplebox}{Sample S9\quad Physical optics}
\samplebody{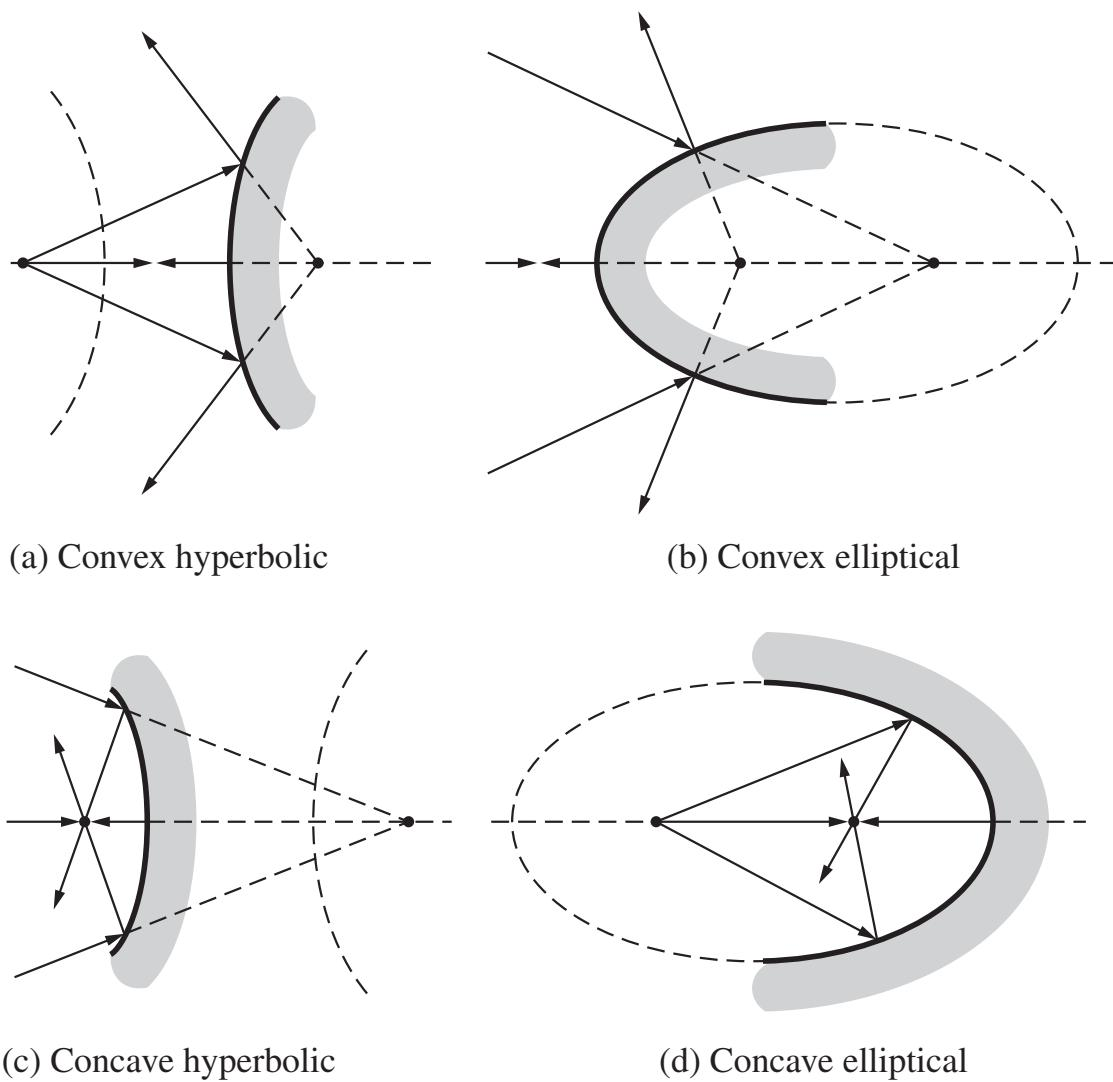}{Reference diagram from the corpus.}%
{Geometric optics: conic mirrors}
{\stepline{1~Scenario}{Four ray diagrams of conic-section mirror segments (convex/concave, hyperbolic/elliptical) with foci $F_1,F_2$ on the optical axis.}}
{\stepline{2~Parameters$^\dagger$}{Reflective segments cut from a hyperbola or an ellipse; the foci marked on the axis.}}
{\stepline{3~Structure$^\dagger$}{Rays reflect at each local surface element by the law of reflection; the conic foci set where rays converge or appear to diverge.}}
{\stepline{4~Laws}{Law of reflection, with the focus-sum property of the ellipse and the focus-difference property of the hyperbola.}}
{\stepline{5~Synthesis}{Conic mirrors redirect rays between their foci. \emph{Assumes} ideal specular reflection.}}
\end{samplebox}
\medskip

\begin{samplebox}{Sample S10\quad Thermodynamics}
\samplebody{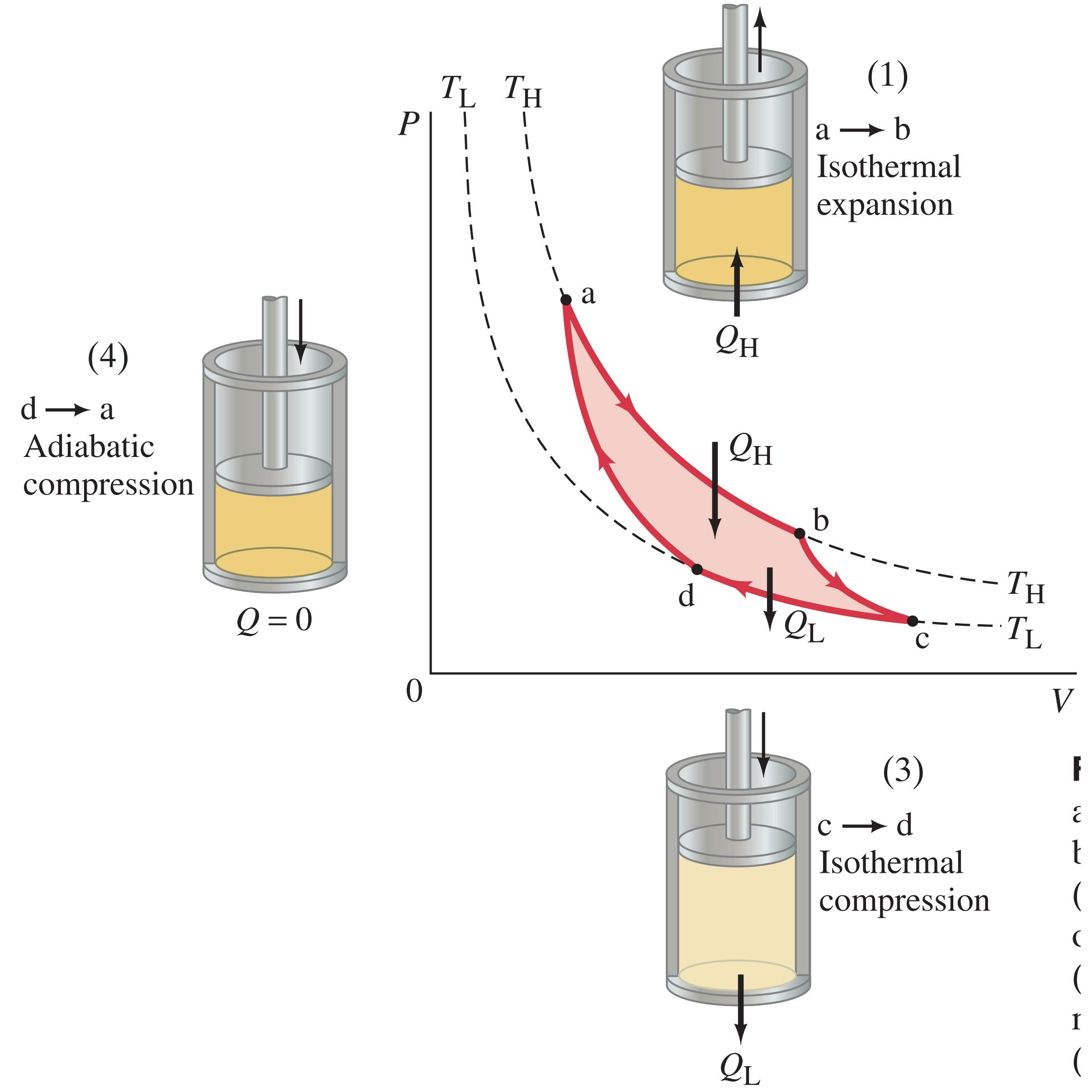}{Reference diagram from the corpus.}%
{Thermodynamic cycle}
{\stepline{1~Scenario}{An ideal gas in a piston--cylinder assembly executing a Carnot cycle, shown on a $P$--$V$ diagram.}}
{\stepline{2~Parameters$^\dagger$}{States $a,b$ at $T_H$ and $c,d$ at $T_L$, each with its $(P,V,T)$.}}
{\stepline{3~Structure$^\dagger$}{$a\!\to\!b$ isothermal expansion ($Q_H$ in), $b\!\to\!c$ adiabatic expansion, $c\!\to\!d$ isothermal compression ($Q_L$ out), $d\!\to\!a$ adiabatic compression.}}
{\stepline{4~Laws}{First law and the ideal-gas law: $\Delta U_{ab}=Q_H-W_{ab}$, $W_{ab}=nRT_H\ln(V_b/V_a)$.}}
{\stepline{5~Synthesis}{The efficiency is set by $T_H$ and $T_L$. \emph{Assumes} an ideal gas and quasi-static, reversible processes.}}
\end{samplebox}
\medskip

\begin{samplebox}{Sample S11\quad Thermodynamics}
\samplebody{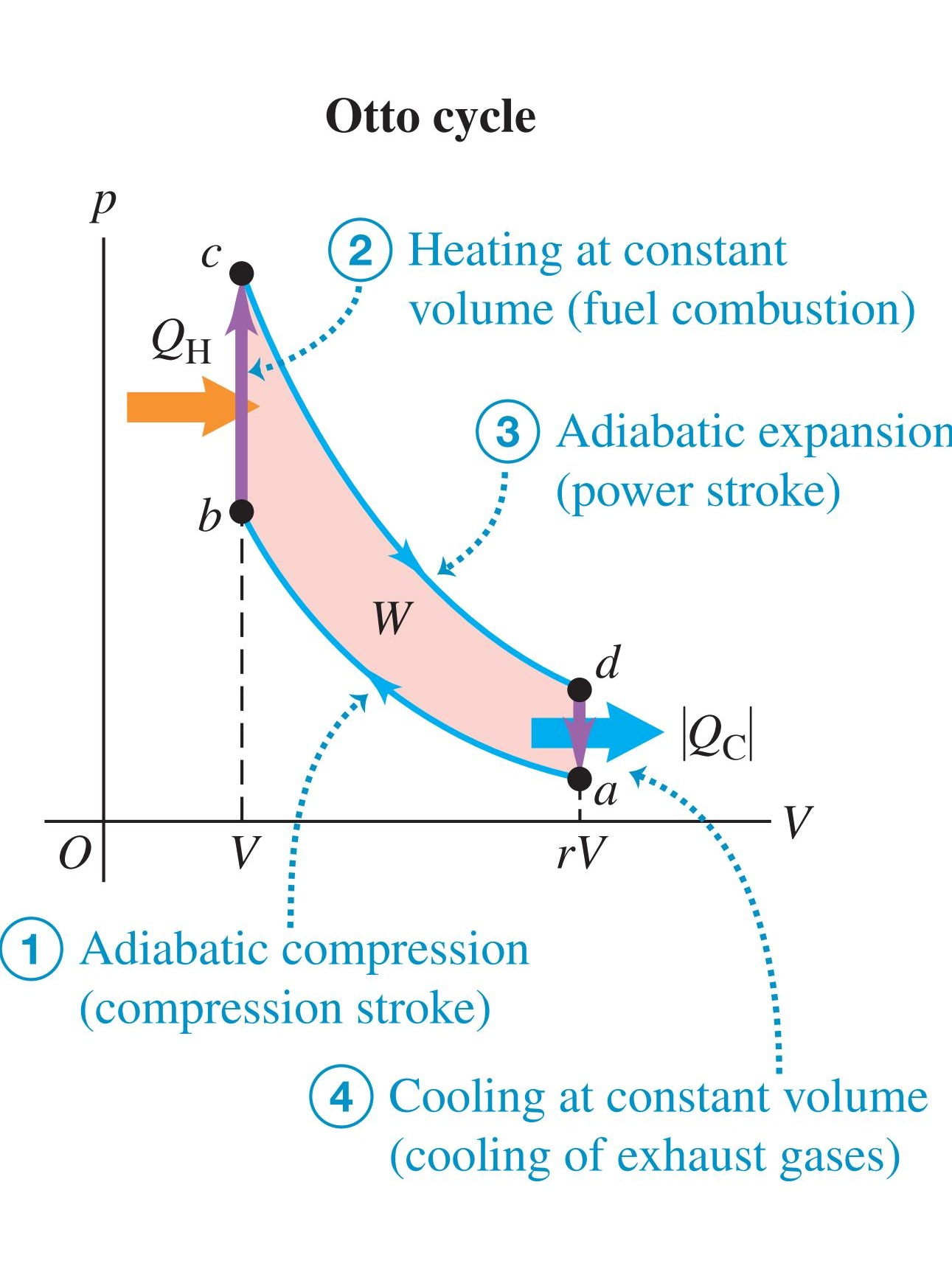}{Reference diagram from the corpus.}%
{Thermodynamic cycle}
{\stepline{1~Scenario}{An ideal gas in a piston--cylinder assembly executing an Otto cycle, shown on a $P$--$V$ diagram.}}
{\stepline{2~Parameters$^\dagger$}{Four states $a,b,c,d$ marking the corners of the cycle.}}
{\stepline{3~Structure$^\dagger$}{$a\!\to\!b$ adiabatic compression, $b\!\to\!c$ isochoric heating ($Q_H$ in), $c\!\to\!d$ adiabatic expansion, $d\!\to\!a$ isochoric cooling ($Q_C$ out).}}
{\stepline{4~Laws}{First law and the ideal-gas law; over the cycle $\Delta U = Q_H - Q_C - W$.}}
{\stepline{5~Synthesis}{The efficiency follows from the heat input and net work. \emph{Assumes} an ideal gas, quasi-static reversible processes and a frictionless piston.}}
\end{samplebox}
\medskip

\begin{samplebox}{Sample S12\quad Thermodynamics}
\samplebody{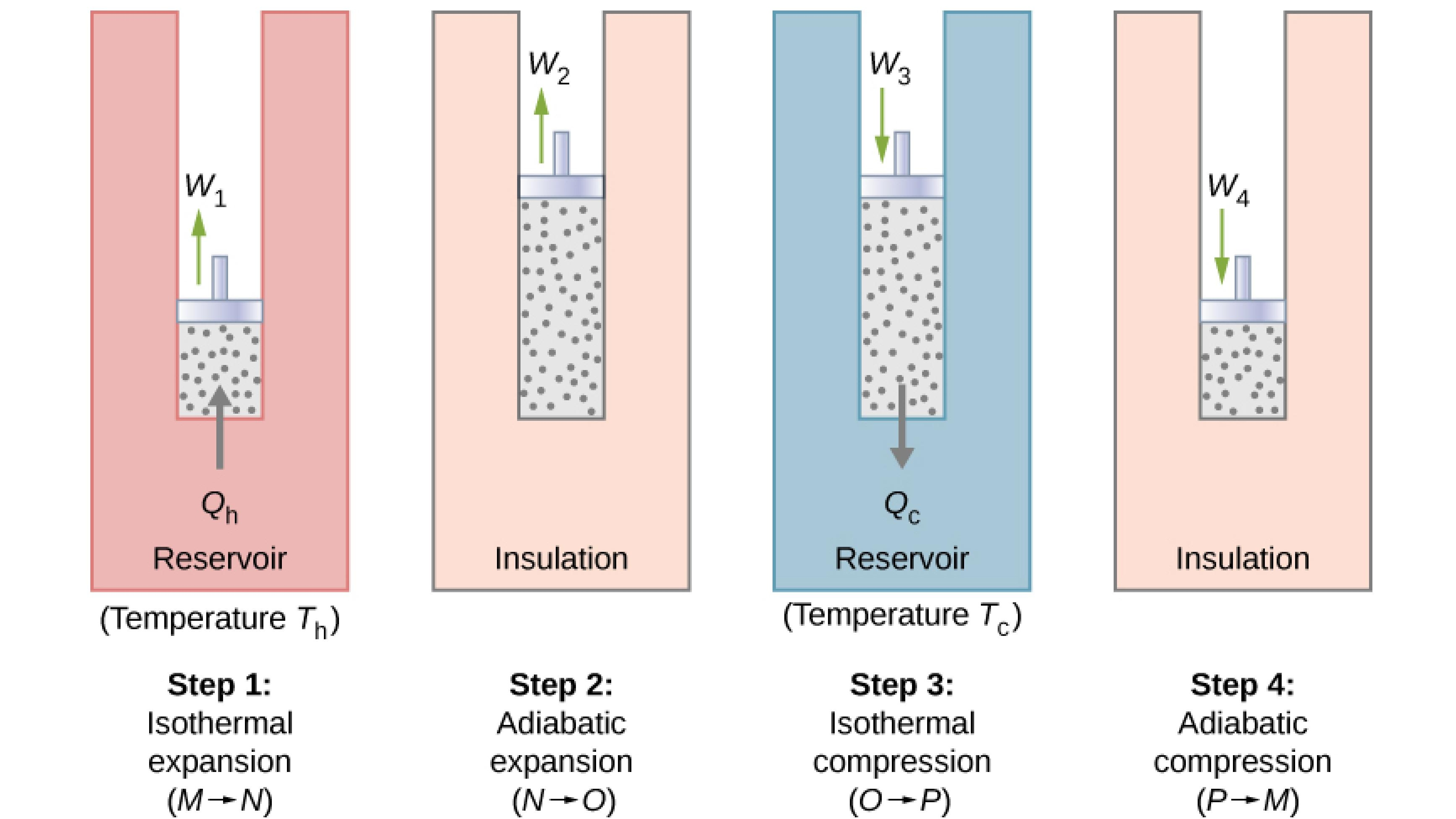}{Reference diagram from the corpus.}%
{Thermodynamic cycle}
{\stepline{1~Scenario}{An ideal gas in a piston--cylinder executing a Carnot cycle, drawn as four piston stages between a hot ($T_h$) and a cold ($T_c$) reservoir.}}
{\stepline{2~Parameters$^\dagger$}{Four states $M,N,O,P$ at the corners of the cycle.}}
{\stepline{3~Structure$^\dagger$}{$M\!\to\!N$ isothermal expansion ($Q_h$ in, $W_1$), $N\!\to\!O$ adiabatic expansion ($W_2$), $O\!\to\!P$ isothermal compression ($Q_c$ out, $W_3$), $P\!\to\!M$ adiabatic compression ($W_4$).}}
{\stepline{4~Laws}{First law and the ideal-gas law: $\Delta U_{MN}=Q_h-W_1$, $W_1=nRT_h\ln(V_N/V_M)$.}}
{\stepline{5~Synthesis}{The efficiency is set by the reservoir temperatures $T_h,T_c$. \emph{Assumes} an ideal gas and quasi-static, reversible processes.}}
\end{samplebox}
\medskip

\begin{samplebox}{Sample S13\quad Acoustics}
\samplebody{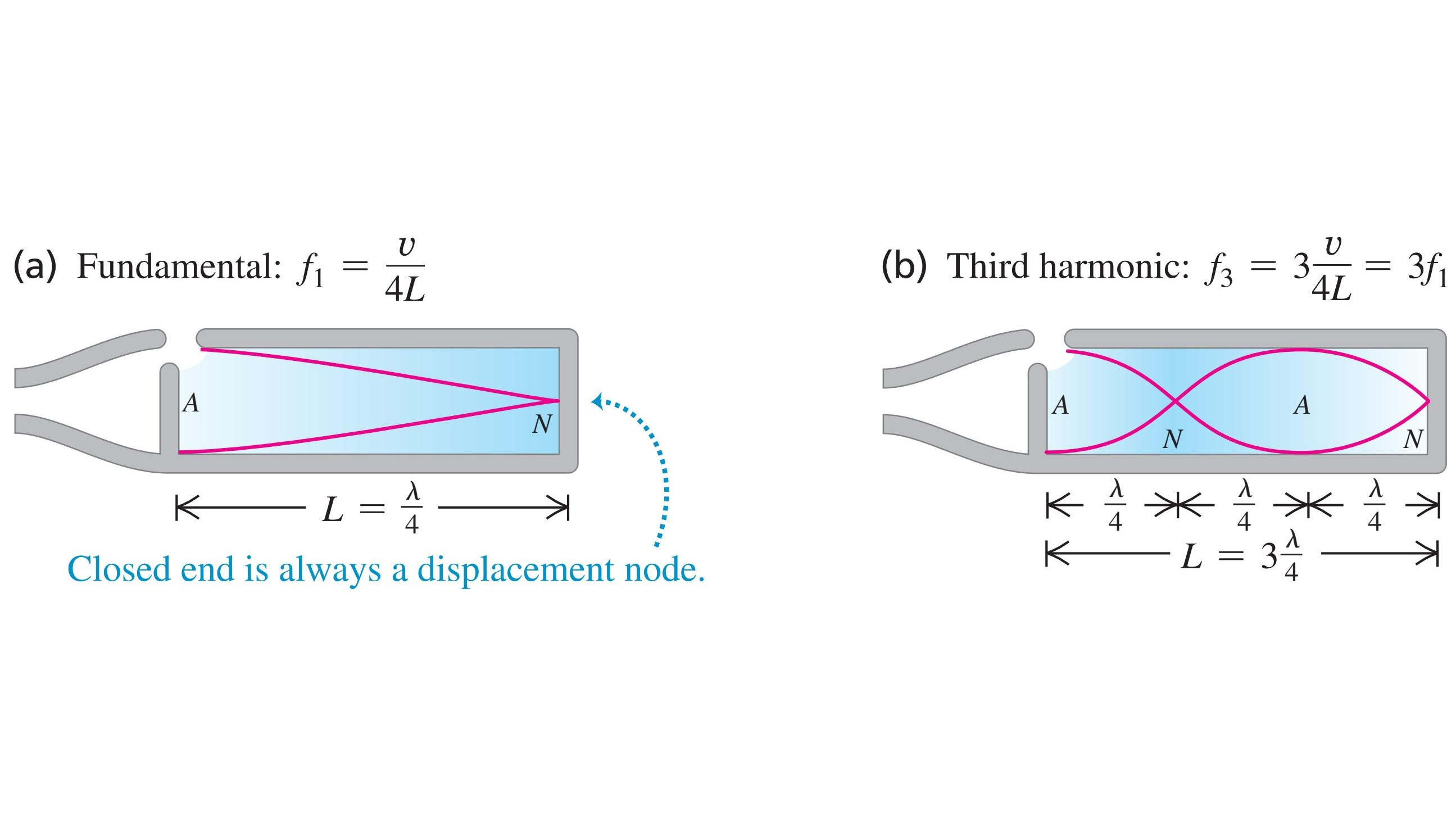}{Reference diagram from the corpus.}%
{Standing waves in pipes}
{\stepline{1~Scenario$^\dagger$}{A standing wave in a closed pipe: (a) the fundamental and (b) the third harmonic.}}
{\stepline{2~Parameters$^\dagger$}{A displacement node $N$ at the closed end and antinode $A$ at the open end; pipe length $L=\lambda/4$ (fundamental), $3\lambda/4$ (third).}}
{\stepline{3~Structure}{A longitudinal standing wave, with $\lambda=4L$ for the fundamental and $\lambda=4L/3$ for the third harmonic.}}
{\stepline{4~Laws}{Closed-pipe standing-wave conditions: $f_1=v/4L$ and $f_3=3v/4L=3f_1$.}}
{\stepline{5~Synthesis}{The harmonics are set by the pipe length and the speed of sound. \emph{Assumes} a non-dispersive medium, a rigid pipe and no dissipation.}}
\end{samplebox}
\medskip

\begin{samplebox}{Sample S14\quad Acoustics}
\samplebody{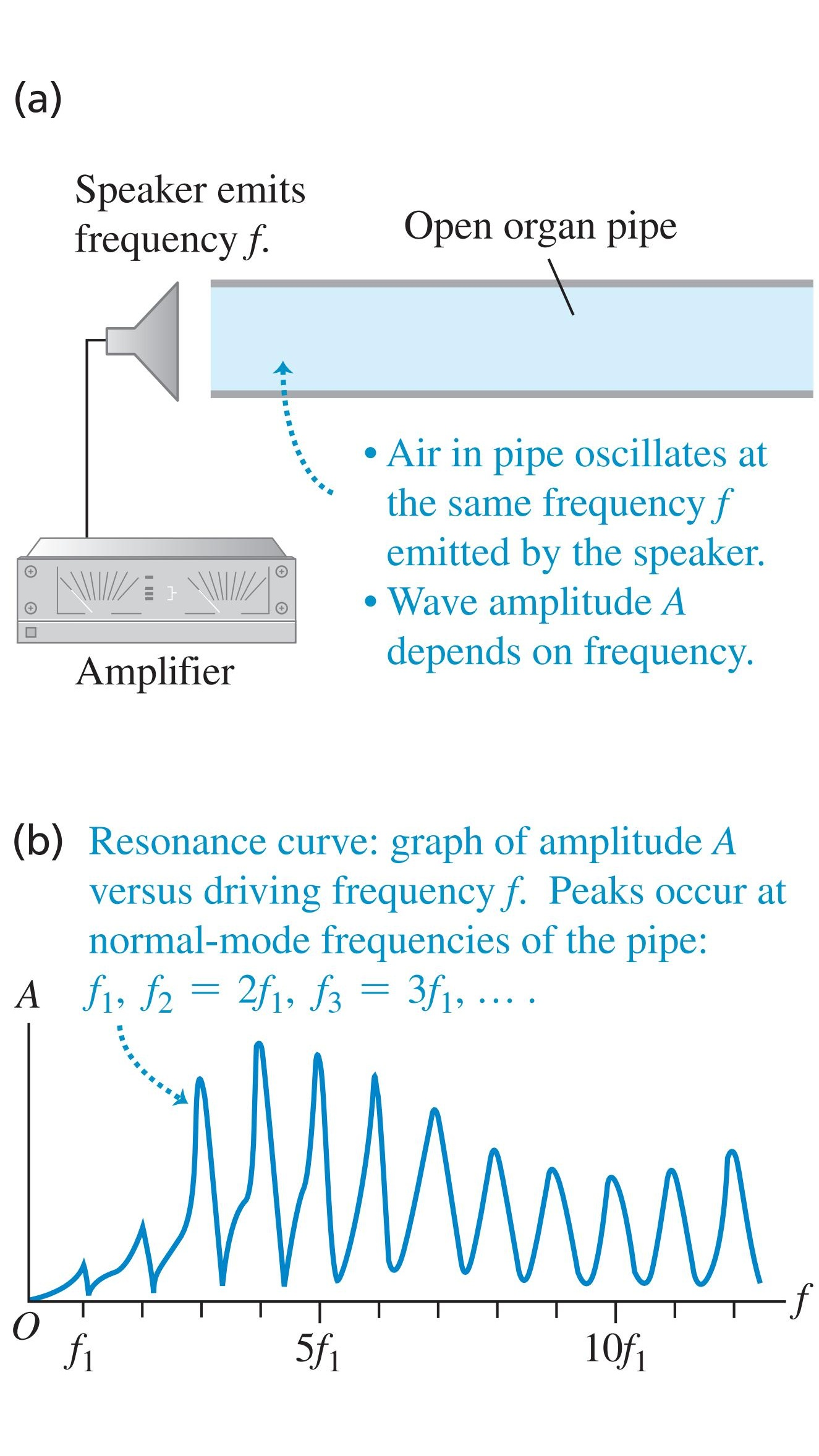}{Reference diagram from the corpus.}%
{Resonance in pipes}
{\stepline{1~Scenario$^\dagger$}{A speaker drives an open organ pipe at frequency $f$, exciting resonance.}}
{\stepline{2~Parameters$^\dagger$}{A sound source (the speaker, frequency $f$) feeding the open pipe; the air oscillates at the driving frequency $f$.}}
{\stepline{3~Structure}{A longitudinal standing wave in the pipe; the amplitude-versus-frequency curve peaks at the normal modes.}}
{\stepline{4~Laws}{Resonance condition: peaks at $f_n = n f_1$, integer multiples of the fundamental.}}
{\stepline{5~Synthesis}{The pipe resonates at integer multiples of $f_1$. \emph{Assumes} a non-dispersive medium, a point source and no dissipation.}}
\end{samplebox}
\medskip

\begin{samplebox}{Sample S15\quad Acoustics}
\samplebody{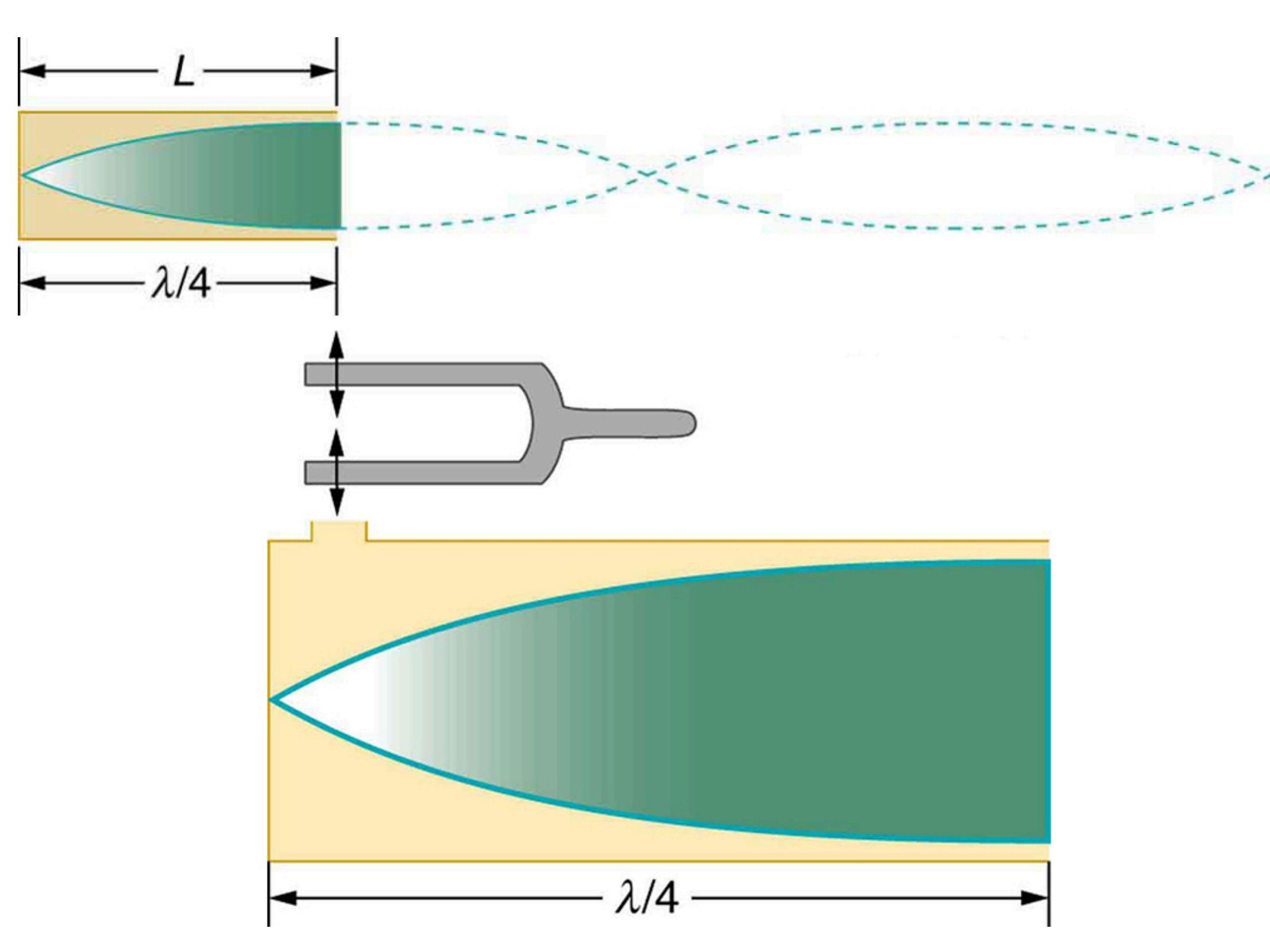}{Reference diagram from the corpus.}%
{Resonance in pipes}
{\stepline{1~Scenario$^\dagger$}{A tuning fork exciting a standing wave in a closed pipe of length $L$, at resonance.}}
{\stepline{2~Parameters$^\dagger$}{A tuning-fork source at the open end and a closed pipe of length $L$, with a displacement node at the closed end.}}
{\stepline{3~Structure}{A longitudinal standing wave whose pipe length is a quarter wavelength.}}
{\stepline{4~Laws}{Closed-pipe resonance condition $L = \lambda/4$ (fundamental).}}
{\stepline{5~Synthesis}{The pipe resonates in its fundamental quarter-wavelength mode. \emph{Assumes} a non-dispersive homogeneous medium and a rigid pipe.}}
\end{samplebox}
\medskip

\begin{samplebox}{Sample S16\quad Quantum mechanics}
\samplebody{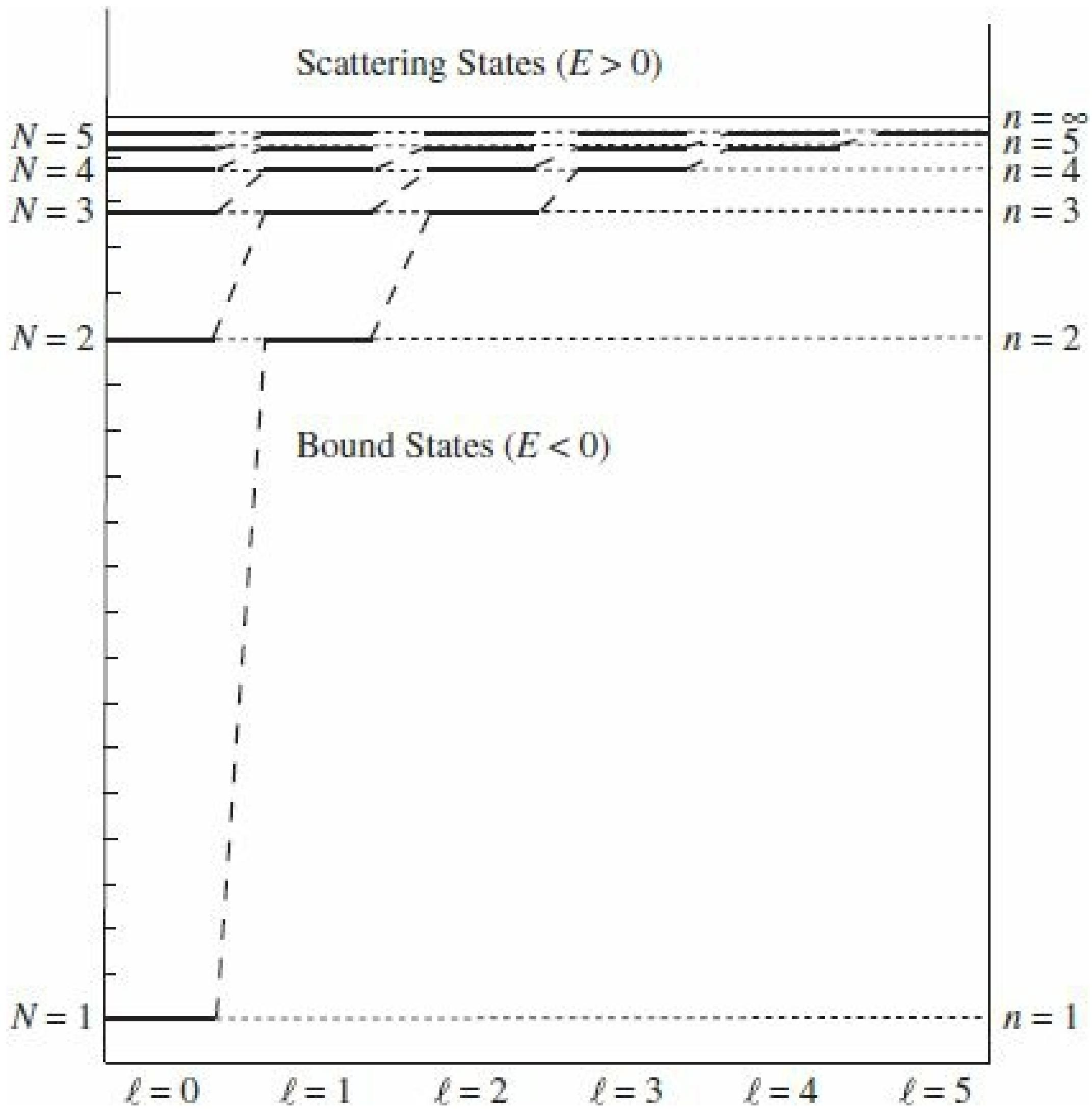}{Reference diagram from the corpus.}%
{Wave mechanics, 1D}
{\stepline{1~Scenario}{An energy-level diagram of a one-dimensional potential well, showing bound ($E<0$) and scattering ($E>0$) states.}}
{\stepline{2~Parameters$^\dagger$}{Hamiltonian $\hat{H}=-\tfrac{\hbar^2}{2m}\nabla^2+V(x)$; states labelled by quantum numbers $N$, $n$ and $\ell$.}}
{\stepline{3~Structure$^\dagger$}{Two regimes, $E<0$ (bound) and $E>0$ (scattering), populated by the energy eigenstates.}}
{\stepline{4~Laws}{The Schr\"odinger equation $i\hbar\,\partial_t\Psi=\hat{H}\Psi$.}}
{\stepline{5~Synthesis}{A discrete set of bound levels below a continuous scattering spectrum. \emph{Assumes} a time-independent potential.}}
\end{samplebox}
\medskip

\begin{samplebox}{Sample S17\quad Quantum mechanics}
\samplebody{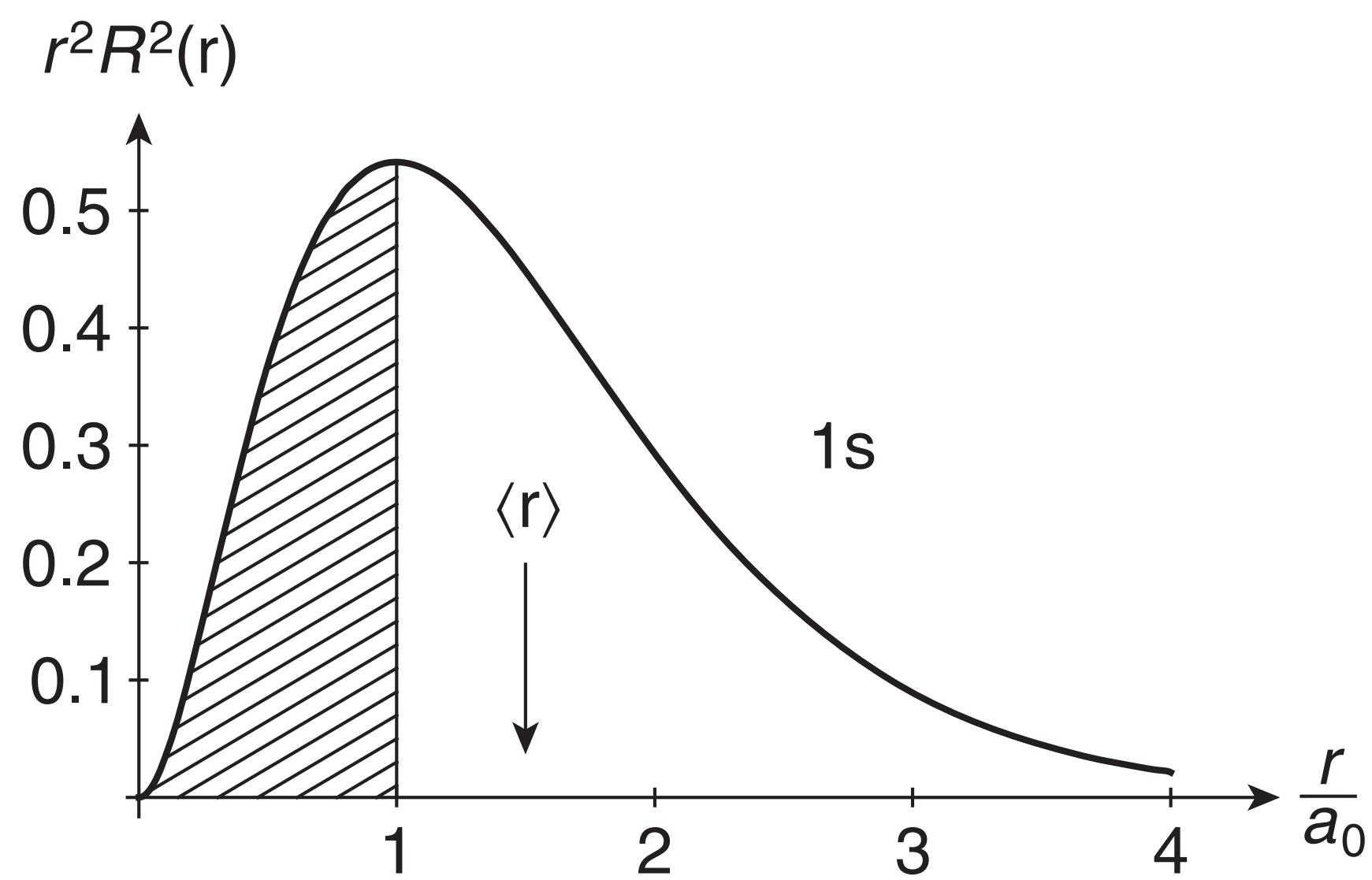}{Reference diagram from the corpus.}%
{Atomic structure: hydrogen}
{\stepline{1~Scenario}{The radial probability distribution $r^2R^2(r)$ of the hydrogen 1s orbital, versus $r/a_0$.}}
{\stepline{2~Parameters$^\dagger$}{Coulomb potential $V(r)=-ke^2/r$; Hamiltonian $\hat{H}=-\tfrac{\hbar^2}{2m}\nabla^2+V(r)$; Bohr radius $a_0$.}}
{\stepline{3~Structure$^\dagger$}{Radial domain $r\ge 0$ with $\psi(0)=0$; the state is the hydrogen orbital $|1s\rangle$.}}
{\stepline{4~Laws}{Radial Schr\"odinger equation $-\tfrac{\hbar^2}{2m}R''(r)+V(r)R(r)=ER(r)$; probability $P(r)=r^2R^2(r)$.}}
{\stepline{5~Synthesis}{The radial distribution peaks near the Bohr radius $a_0$. \emph{Assumes} a non-relativistic, central potential.}}
\end{samplebox}
\medskip

\begin{samplebox}{Sample S18\quad Quantum mechanics}
\samplebody{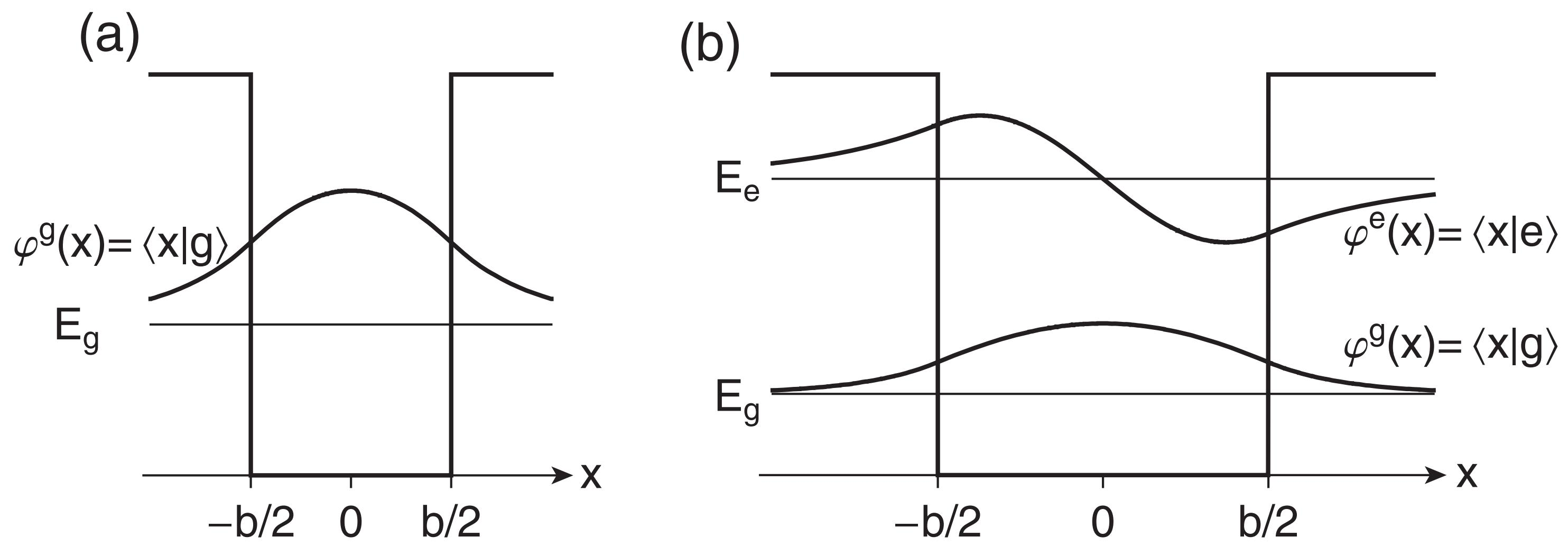}{Reference diagram from the corpus.}%
{Wave mechanics, 1D}
{\stepline{1~Scenario}{A one-dimensional potential well of width $b$, showing the ground- and excited-state wavefunctions $\varphi_g,\varphi_e$ at energies $E_g,E_e$.}}
{\stepline{2~Parameters$^\dagger$}{Hamiltonian $\hat{H}=-\tfrac{\hbar^2}{2m}\tfrac{d^2}{dx^2}+V(x)$; the well spans $-b/2 < x < b/2$.}}
{\stepline{3~Structure$^\dagger$}{Boundary conditions $\psi(-b/2)=\psi(b/2)=0$; eigenstates $\varphi_g=\langle x|g\rangle$ and $\varphi_e=\langle x|e\rangle$.}}
{\stepline{4~Laws}{The Schr\"odinger equation $i\hbar\,\partial_t\Psi=\hat{H}\Psi$.}}
{\stepline{5~Synthesis}{Quantized levels, here the ground and first excited state. \emph{Assumes} a time-independent potential.}}
\end{samplebox}
\medskip

\subsection{Complete structured annotations}
\label{sec:methods-fullcases}

Whereas \Cref{sec:methods-samples} condenses each annotation to a single line per schema step, this appendix reproduces the \emph{complete} structured record for one diagram in each subdiscipline (Cases~S1--S6), rendered field by field in the hierarchical form used by our annotation viewer, with the reference diagram shown alongside. Each case is the machine-generated, expert-verified JSON object exactly as it is stored: numbered step bands carry both a human-readable title and the underlying machine key (e.g.\ \texttt{step1\_scene\_and\_object\_identification}); every key is shown with its value; nested objects, forces and states are set as indented sub-entries; all mathematical content is retained verbatim in its stored form; and fields left empty under the fidelity rules are marked explicitly as \annempty{}. The empty markers are themselves informative, recording where the annotator declined to assert a value the diagram does not license. Read against the condensed cards of \Cref{sec:methods-samples}, these records expose the full granularity of the supervision signal that a flat caption would collapse to a single sentence.

\begin{anncase}{Case S1\quad Mechanics: complete structured annotation}
\begin{minipage}[t]{0.36\linewidth}\vspace{0pt}\centering
\includegraphics[width=\linewidth]{imgs/gallery/mech1.png}\\[2pt]
{\scriptsize\itshape Reference diagram.}\end{minipage}\hfill
\begin{minipage}[t]{0.60\linewidth}\vspace{0pt}
\annrow{category}{Fundamentals of Mechanics}
\annrow{summary}{A pulley system with a block on an inclined plane and a hanging mass.}
\annrow{detailed}{The diagram shows a block on an inclined plane with angle $ \theta $, connected by a rope over a pulley to a hanging mass. The forces acting on the block include $ Mg\sin\theta $, tension $ T $, and friction $ F $. The block has an acceleration $ a_1 $ down the incline. The hanging mass experiences gravitational force $ mg $, tension $ T $, and has an upward acceleration $ a_2 $. The pulley has an angular acceleration $ \alpha $.}
\end{minipage}\par\smallskip
\annstep{Scene and object identification}{step1\_\allowbreak scene\_\allowbreak and\_\allowbreak object\_\allowbreak identification}
\annrow{core objects}{}
{\setlength{\annind}{1.3em}
\annitem{Block on Incline}
{\setlength{\annind}{2.6em}
\annline{\annlab{mass}\hspace{0.3em}M\annsep{}\annlab{properties}\hspace{0.3em}Inclined at angle $ \theta $}
\par}
\annitem{Hanging Mass}
{\setlength{\annind}{2.6em}
\annline{\annlab{mass}\hspace{0.3em}m\annsep{}\annlab{properties}\hspace{0.3em}\annempty}
\par}
\annitem{Pulley}
{\setlength{\annind}{2.6em}
\annline{\annlab{mass}\hspace{0.3em}\annempty\annsep{}\annlab{properties}\hspace{0.3em}Connected to block and mass}
\par}
\par}
\annrow{environment}{Inclined plane with angle $ \theta $, pulley system}
\annstep{Motion state analysis}{step2\_\allowbreak motion\_\allowbreak state\_\allowbreak analysis}
\annrow{state}{Dynamic}
\annrow{motion}{Block accelerates down the incline, mass accelerates upward.}
\annrow{kinematics}{\annlab{velocity v}\hspace{0.3em}\annempty\annsep{}\annlab{acceleration a}\hspace{0.3em}$a_1, a_2$}
\annstep{Force analysis}{step3\_\allowbreak force\_\allowbreak analysis}
\annitem{Block on Incline}
{\setlength{\annind}{1.3em}
\annline{\annlab{forces}}
\annitem{Gravitational Component}
{\setlength{\annind}{2.6em}
\annline{\annlab{symbol}\hspace{0.3em}$Mg\sin\theta$\annsep{}\annlab{source}\hspace{0.3em}Gravity\annsep{}\annlab{direction}\hspace{0.3em}Down the incline\annsep{}\annlab{magnitude}\hspace{0.3em}$Mg\sin\theta$}
\par}
\annitem{Tension}
{\setlength{\annind}{2.6em}
\annline{\annlab{symbol}\hspace{0.3em}$T$\annsep{}\annlab{source}\hspace{0.3em}Rope\annsep{}\annlab{direction}\hspace{0.3em}Up the incline\annsep{}\annlab{magnitude}\hspace{0.3em}$T$}
\par}
\annitem{Friction}
{\setlength{\annind}{2.6em}
\annline{\annlab{symbol}\hspace{0.3em}$F$\annsep{}\annlab{source}\hspace{0.3em}Surface\annsep{}\annlab{direction}\hspace{0.3em}Up the incline\annsep{}\annlab{magnitude}\hspace{0.3em}$F$}
\par}
\par}
\annitem{Hanging Mass}
{\setlength{\annind}{1.3em}
\annline{\annlab{forces}}
\annitem{Gravitational Force}
{\setlength{\annind}{2.6em}
\annline{\annlab{symbol}\hspace{0.3em}$mg$\annsep{}\annlab{source}\hspace{0.3em}Gravity\annsep{}\annlab{direction}\hspace{0.3em}Downward\annsep{}\annlab{magnitude}\hspace{0.3em}$mg$}
\par}
\annitem{Tension}
{\setlength{\annind}{2.6em}
\annline{\annlab{symbol}\hspace{0.3em}$T$\annsep{}\annlab{source}\hspace{0.3em}Rope\annsep{}\annlab{direction}\hspace{0.3em}Upward\annsep{}\annlab{magnitude}\hspace{0.3em}$T$}
\par}
\par}
\annstep{Coordinate system and laws}{step4\_\allowbreak coordinate\_\allowbreak system\_\allowbreak and\_\allowbreak laws}
\annrow{coordinates}{Inclined plane coordinates for block, vertical for mass}
\annrow{law}{Newton's Second Law}
\annrow{equations}{}
{\setlength{\annind}{1.3em}
\annitem{Block on incline}
{\setlength{\annind}{2.6em}
\annline{\annlab{equation}\hspace{0.3em}$Ma_1 = Mg\sin\theta - T - F$}
\par}
\annitem{Hanging mass}
{\setlength{\annind}{2.6em}
\annline{\annlab{equation}\hspace{0.3em}$ma_2 = T - mg$}
\par}
\par}
\annstep{Synthesis and conclusion}{step5\_\allowbreak synthesis\_\allowbreak and\_\allowbreak conclusion}
\annrow{key relationship}{The system's motion is determined by the balance of forces and the tension in the rope.}
\annrow{assumptions}{Neglect air resistance; Assume pulley is frictionless}
\end{anncase}
\medskip

\begin{anncase}{Case S2\quad Electromagnetism: complete structured annotation}
\begin{minipage}[t]{0.36\linewidth}\vspace{0pt}\centering
\includegraphics[width=\linewidth]{imgs/gallery/em1.png}\\[2pt]
{\scriptsize\itshape Reference diagram.}\end{minipage}\hfill
\begin{minipage}[t]{0.60\linewidth}\vspace{0pt}
\annrow{category}{Electromagnetism}
\annrow{summary}{Diagram showing electromagnetic wave interaction with a surface, illustrating incident, reflected, and transmitted waves.}
\annrow{detailed}{The diagram depicts a surface separating two media with refractive indices $ n_1 $ and $ n_2 $. An incident wave with electric field $ E_i $ and magnetic field $ B_i $ approaches the surface with wave vector $ \mathbf{k} $. The reflected wave has electric field $ E_r $, magnetic field $ B_r $, and wave vector $ \mathbf{k}' $. The transmitted wave has electric field $ E_t $, magnetic field $ B_t $, and wave vector $ \mathbf{k}'' $. The coordinate system is defined with $ x $ and $ y $ axes.}
\end{minipage}\par\smallskip
\annstep{Scene identification and type judgment}{step1\_\allowbreak scene\_\allowbreak identification\_\allowbreak and\_\allowbreak type\_\allowbreak judgment}
\annrow{domain}{Electromagnetism}
\annrow{sub-domain}{Electrodynamics}
\annrow{problem}{Wave Reflection and Transmission at a Boundary}
\annstep{Component source identification and parameterization}{step2\_\allowbreak component\_\allowbreak source\_\allowbreak identification\_\allowbreak and\_\allowbreak parameterization}
\annrow{components}{}
{\setlength{\annind}{1.3em}
\annitem{E\_i}
{\setlength{\annind}{2.6em}
\annline{\annlab{type}\hspace{0.3em}Electric Field}
\annline{\annlab{parameters}}
\annitem{$E_i$}
{\setlength{\annind}{3.9em}
\annline{\annlab{description}\hspace{0.3em}Incident Electric Field\annsep{}\annlab{value}\hspace{0.3em}\annempty}
\par}
\par}
\annitem{B\_i}
{\setlength{\annind}{2.6em}
\annline{\annlab{type}\hspace{0.3em}Magnetic Field}
\annline{\annlab{parameters}}
\annitem{$B_i$}
{\setlength{\annind}{3.9em}
\annline{\annlab{description}\hspace{0.3em}Incident Magnetic Field\annsep{}\annlab{value}\hspace{0.3em}\annempty}
\par}
\par}
\annitem{E\_r}
{\setlength{\annind}{2.6em}
\annline{\annlab{type}\hspace{0.3em}Electric Field}
\annline{\annlab{parameters}}
\annitem{$E_r$}
{\setlength{\annind}{3.9em}
\annline{\annlab{description}\hspace{0.3em}Reflected Electric Field\annsep{}\annlab{value}\hspace{0.3em}\annempty}
\par}
\par}
\annitem{B\_r}
{\setlength{\annind}{2.6em}
\annline{\annlab{type}\hspace{0.3em}Magnetic Field}
\annline{\annlab{parameters}}
\annitem{$B_r$}
{\setlength{\annind}{3.9em}
\annline{\annlab{description}\hspace{0.3em}Reflected Magnetic Field\annsep{}\annlab{value}\hspace{0.3em}\annempty}
\par}
\par}
\annitem{E\_t}
{\setlength{\annind}{2.6em}
\annline{\annlab{type}\hspace{0.3em}Electric Field}
\annline{\annlab{parameters}}
\annitem{$E_t$}
{\setlength{\annind}{3.9em}
\annline{\annlab{description}\hspace{0.3em}Transmitted Electric Field\annsep{}\annlab{value}\hspace{0.3em}\annempty}
\par}
\par}
\annitem{B\_t}
{\setlength{\annind}{2.6em}
\annline{\annlab{type}\hspace{0.3em}Magnetic Field}
\annline{\annlab{parameters}}
\annitem{$B_t$}
{\setlength{\annind}{3.9em}
\annline{\annlab{description}\hspace{0.3em}Transmitted Magnetic Field\annsep{}\annlab{value}\hspace{0.3em}\annempty}
\par}
\par}
\par}
\annstep{Structural topology and state analysis}{step3\_\allowbreak structural\_\allowbreak topology\_\allowbreak and\_\allowbreak state\_\allowbreak analysis}
\annrow{circuit}{\annlab{connection}\hspace{0.3em}\annempty\annsep{}\annlab{nodes}\hspace{0.3em}\annempty\annsep{}\annlab{loops}\hspace{0.3em}\annempty}
\annrow{field}{\annlab{source geom.}\hspace{0.3em}Plane Wave\annsep{}\annlab{field geom.}\hspace{0.3em}Reflection and Transmission}
\annrow{state}{Electromagnetic Wave Interaction}
\annstep{Inference of principles and equation formulation}{step4\_\allowbreak inference\_\allowbreak of\_\allowbreak principles\_\allowbreak and\_\allowbreak equation\_\allowbreak formulation}
\annrow{conventions}{\annlab{coordinates}\hspace{0.3em}Cartesian (x,y)\annsep{}\annlab{conventions}\hspace{0.3em}\annempty}
\annrow{laws}{Snell's Law; Boundary Conditions for Electromagnetic Fields}
\annrow{equations}{}
{\setlength{\annind}{1.3em}
\annitem{Snell's Law for Refraction}
{\setlength{\annind}{2.6em}
\annline{\annlab{equation}\hspace{0.3em}$n_1 \sin(\theta_i) = n_2 \sin(\theta_t)$}
\par}
\annitem{Boundary Condition for Electric Fields}
{\setlength{\annind}{2.6em}
\annline{\annlab{equation}\hspace{0.3em}$E_{i} + E_{r} = E_{t}$}
\par}
\par}
\annstep{Synthesized output and assumption declaration}{step5\_\allowbreak synthesized\_\allowbreak output\_\allowbreak and\_\allowbreak assumption\_\allowbreak declaration}
\annrow{key findings}{Determine the angles and magnitudes of reflected and transmitted waves.}
\annrow{assumptions}{Idealized plane wave; No absorption at the boundary}
\end{anncase}
\medskip

\begin{anncase}{Case S4\quad Thermodynamics: complete structured annotation}
\begin{minipage}[t]{0.36\linewidth}\vspace{0pt}\centering
\includegraphics[width=\linewidth]{imgs/gallery/thermo1.png}\\[2pt]
{\scriptsize\itshape Reference diagram.}\end{minipage}\hfill
\begin{minipage}[t]{0.60\linewidth}\vspace{0pt}
\annrow{category}{Thermodynamics}
\annrow{summary}{An ideal gas undergoes a Carnot cycle consisting of four stages, as shown on a P-V diagram.}
\annrow{detailed}{The diagram is a P-V diagram with labeled axes for pressure (P) and volume (V). It shows four states labeled a, b, c, and d. The process paths are indicated with arrows: a to b (isothermal expansion), b to c (adiabatic expansion), c to d (isothermal compression), and d to a (adiabatic compression). There are schematics of a piston-cylinder assembly for each process, with heat transfer labeled as Q\_H and Q\_L.}
\end{minipage}\par\smallskip
\annstep{Scene identification and system definition}{step1\_\allowbreak scene\_\allowbreak identification\_\allowbreak and\_\allowbreak system\_\allowbreak definition}
\annrow{domain}{Thermodynamics}
\annrow{system}{An ideal gas confined in a piston-cylinder assembly undergoing a Carnot cycle.}
\annrow{problem}{Thermodynamic Cycle}
\annstep{State identification and parameterization}{step2\_\allowbreak state\_\allowbreak identification\_\allowbreak and\_\allowbreak parameterization}
\annrow{states}{}
{\setlength{\annind}{1.3em}
\annitem{a}
{\setlength{\annind}{2.6em}
\annline{\annlab{variables}}
\annitem{$P_a, V_a, T_H$}
{\setlength{\annind}{3.9em}
\annline{\annlab{description}\hspace{0.3em}Pressure, Volume, and Temperature at state a\annsep{}\annlab{value}\hspace{0.3em}\annempty}
\par}
\par}
\annitem{b}
{\setlength{\annind}{2.6em}
\annline{\annlab{variables}}
\annitem{$P_b, V_b, T_H$}
{\setlength{\annind}{3.9em}
\annline{\annlab{description}\hspace{0.3em}Pressure, Volume, and Temperature at state b\annsep{}\annlab{value}\hspace{0.3em}\annempty}
\par}
\par}
\annitem{c}
{\setlength{\annind}{2.6em}
\annline{\annlab{variables}}
\annitem{$P_c, V_c, T_L$}
{\setlength{\annind}{3.9em}
\annline{\annlab{description}\hspace{0.3em}Pressure, Volume, and Temperature at state c\annsep{}\annlab{value}\hspace{0.3em}\annempty}
\par}
\par}
\annitem{d}
{\setlength{\annind}{2.6em}
\annline{\annlab{variables}}
\annitem{$P_d, V_d, T_L$}
{\setlength{\annind}{3.9em}
\annline{\annlab{description}\hspace{0.3em}Pressure, Volume, and Temperature at state d\annsep{}\annlab{value}\hspace{0.3em}\annempty}
\par}
\par}
\par}
\annstep{Process and interaction analysis}{step3\_\allowbreak process\_\allowbreak and\_\allowbreak interaction\_\allowbreak analysis}
\annrow{processes}{}
{\setlength{\annind}{1.3em}
\annitem{a -> b}
{\setlength{\annind}{2.6em}
\annline{\annlab{type}\hspace{0.3em}Isothermal Expansion}
\annline{\annlab{energy}}
\annitem{$Q_H$}
{\setlength{\annind}{3.9em}
\annline{\annlab{type}\hspace{0.3em}Heat\annsep{}\annlab{direction}\hspace{0.3em}Into system}
\par}
\par}
\annitem{b -> c}
{\setlength{\annind}{2.6em}
\annline{\annlab{type}\hspace{0.3em}Adiabatic Expansion\annsep{}\annlab{energy}\hspace{0.3em}\annempty}
\par}
\annitem{c -> d}
{\setlength{\annind}{2.6em}
\annline{\annlab{type}\hspace{0.3em}Isothermal Compression}
\annline{\annlab{energy}}
\annitem{$Q_L$}
{\setlength{\annind}{3.9em}
\annline{\annlab{type}\hspace{0.3em}Heat\annsep{}\annlab{direction}\hspace{0.3em}Out of system}
\par}
\par}
\annitem{d -> a}
{\setlength{\annind}{2.6em}
\annline{\annlab{type}\hspace{0.3em}Adiabatic Compression\annsep{}\annlab{energy}\hspace{0.3em}\annempty}
\par}
\par}
\annstep{Principles and equation formulation}{step4\_\allowbreak principles\_\allowbreak and\_\allowbreak equation\_\allowbreak formulation}
\annrow{laws}{First Law of Thermodynamics; Ideal Gas Law}
\annrow{equations}{}
{\setlength{\annind}{1.3em}
\annitem{First Law applied to isothermal expansion a -> b}
{\setlength{\annind}{2.6em}
\annline{\annlab{equation}\hspace{0.3em}$\Delta U_{ab} = Q_H - W_{ab}$}
\par}
\annitem{Work done during isothermal expansion a -> b}
{\setlength{\annind}{2.6em}
\annline{\annlab{equation}\hspace{0.3em}$W_{ab} = nRT_H \ln\left(\frac{V_b}{V_a}\right)$}
\par}
\par}
\annstep{Synthesis and assumption declaration}{step5\_\allowbreak synthesis\_\allowbreak and\_\allowbreak assumption\_\allowbreak declaration}
\annrow{key findings}{The efficiency of the Carnot cycle is determined by the temperatures T\_H and T\_L.}
\annrow{assumptions}{The working substance is an ideal gas; All processes are quasi-static and reversible; The piston is massless and frictionless; The cylinder walls are perfectly insulating}
\end{anncase}
\medskip

\begin{anncase}{Case S5\quad Acoustics: complete structured annotation}
\begin{minipage}[t]{0.36\linewidth}\vspace{0pt}\centering
\includegraphics[width=\linewidth]{imgs/gallery/acou1.png}\\[2pt]
{\scriptsize\itshape Reference diagram.}\end{minipage}\hfill
\begin{minipage}[t]{0.60\linewidth}\vspace{0pt}
\annrow{category}{Acoustics}
\annrow{summary}{A closed pipe demonstrates standing waves at the fundamental frequency and third harmonic.}
\annrow{detailed}{The diagram shows a closed pipe with a displacement node at the closed end and an antinode at the open end. (a) illustrates the fundamental frequency with one node and one antinode, while (b) shows the third harmonic with two nodes and two antinodes. The length of the pipe is related to the wavelength of the sound.}
\end{minipage}\par\smallskip
\annstep{Scene identification and phenomenon judgment}{step1\_\allowbreak scene\_\allowbreak identification\_\allowbreak and\_\allowbreak phenomenon\_\allowbreak judgment}
\annrow{domain}{Acoustics}
\annrow{sub-domain}{Standing Waves in Pipes}
\annrow{phenomenon}{Standing Wave in a Closed Pipe}
\annstep{System component identification and parameterization}{step2\_\allowbreak system\_\allowbreak component\_\allowbreak identification\_\allowbreak and\_\allowbreak parameterization}
\annrow{components}{}
{\setlength{\annind}{1.3em}
\annitem{A}
{\setlength{\annind}{2.6em}
\annline{\annlab{type}\hspace{0.3em}Antinode\annsep{}\annlab{parameters}\hspace{0.3em}\annempty}
\par}
\annitem{N}
{\setlength{\annind}{2.6em}
\annline{\annlab{type}\hspace{0.3em}Node\annsep{}\annlab{parameters}\hspace{0.3em}\annempty}
\par}
\annitem{L}
{\setlength{\annind}{2.6em}
\annline{\annlab{type}\hspace{0.3em}Pipe Length}
\annline{\annlab{parameters}}
\annitem{$L$}
{\setlength{\annind}{3.9em}
\annline{\annlab{description}\hspace{0.3em}Length of the pipe\annsep{}\annlab{value}\hspace{0.3em}$\frac{\lambda}{4} \text{ for fundamental, } \frac{3\lambda}{4} \text{ for third harmonic}$}
\par}
\par}
\par}
\annstep{Geometry and wave analysis}{step3\_\allowbreak geometry\_\allowbreak and\_\allowbreak wave\_\allowbreak analysis}
\annrow{arrangement}{\annlab{description}\hspace{0.3em}The pipe is closed at one end, creating a node, and open at the other, creating an antinode.\annsep{}\annlab{rel. motion}\hspace{0.3em}\annempty}
\annrow{wave}{\annlab{wave type}\hspace{0.3em}Longitudinal\annsep{}\annlab{wave form}\hspace{0.3em}Standing Wave\annsep{}\annlab{wavelength}\hspace{0.3em}$\lambda = 4L \text{ for fundamental, } \lambda = \frac{4L}{3} \text{ for third harmonic}$}
\annstep{Principles and equation formulation}{step4\_\allowbreak principles\_\allowbreak and\_\allowbreak equation\_\allowbreak formulation}
\annrow{conventions}{\annlab{description}\hspace{0.3em}The closed end is a displacement node, and the open end is a displacement antinode.}
\annrow{principles}{Standing Wave Conditions in a Closed Pipe}
\annrow{equations}{}
{\setlength{\annind}{1.3em}
\annitem{Fundamental frequency equation}
{\setlength{\annind}{2.6em}
\annline{\annlab{equation}\hspace{0.3em}$f_1 = \frac{v}{4L}$}
\par}
\annitem{Third harmonic frequency equation}
{\setlength{\annind}{2.6em}
\annline{\annlab{equation}\hspace{0.3em}$f_3 = 3 \frac{v}{4L} = 3f_1$}
\par}
\par}
\annstep{Synthesis and assumption declaration}{step5\_\allowbreak synthesis\_\allowbreak and\_\allowbreak assumption\_\allowbreak declaration}
\annrow{key findings}{The fundamental frequency and harmonics are determined by the length of the pipe and the speed of sound.}
\annrow{assumptions}{The medium is non-dispersive and homogeneous; The pipe is perfectly rigid; Energy dissipation is neglected}
\end{anncase}
\medskip

\begin{anncase}{Case S6\quad Quantum mechanics: complete structured annotation}
\begin{minipage}[t]{0.36\linewidth}\vspace{0pt}\centering
\includegraphics[width=\linewidth]{imgs/gallery/qm1.png}\\[2pt]
{\scriptsize\itshape Reference diagram.}\end{minipage}\hfill
\begin{minipage}[t]{0.60\linewidth}\vspace{0pt}
\annrow{category}{Quantum Mechanics}
\annrow{summary}{Energy level diagram showing bound and scattering states.}
\annrow{detailed}{The diagram displays energy levels with bound states (E < 0) and scattering states (E > 0). The vertical axis represents energy, while the horizontal axis shows different angular momentum states labeled as \textbackslash{}ell = 0, 1, 2, 3, 4, 5. Bound states are shown for N = 1 and N = 2, while scattering states are shown for N = 3, 4, 5. Each energy level is labeled with quantum numbers N and n.}
\end{minipage}\par\smallskip
\annstep{Scene identification}{step1\_\allowbreak scene\_\allowbreak identification}
\annrow{domain}{Quantum Mechanics}
\annrow{formalism}{Wave Mechanics}
\annrow{scenario}{Potential Well}
\annrow{d.o.f.}{1D Spatial}
\annstep{Component parameterization}{step2\_\allowbreak component\_\allowbreak parameterization}
\annrow{Hamiltonian}{\annlab{description}\hspace{0.3em}Potential well with discrete energy levels for bound states and continuous levels for scattering states.\annsep{}\annlab{operator}\hspace{0.3em}$\hat{H} = -\frac{\hbar^2}{2m}\nabla^2 + V(x)$}
\annrow{parameters}{}
{\setlength{\annind}{1.3em}
\annitem{$N$}
{\setlength{\annind}{2.6em}
\annline{\annlab{description}\hspace{0.3em}Principal Quantum Number\annsep{}\annlab{value}\hspace{0.3em}\annempty}
\par}
\annitem{$n$}
{\setlength{\annind}{2.6em}
\annline{\annlab{description}\hspace{0.3em}Radial Quantum Number\annsep{}\annlab{value}\hspace{0.3em}\annempty}
\par}
\annitem{$\ell$}
{\setlength{\annind}{2.6em}
\annline{\annlab{description}\hspace{0.3em}Angular Momentum Quantum Number\annsep{}\annlab{value}\hspace{0.3em}\annempty}
\par}
\par}
\annstep{State analysis}{step3\_\allowbreak state\_\allowbreak analysis}
\annrow{boundary}{\annlab{regions}\hspace{0.3em}Region I (E < 0), Region II (E > 0)\annsep{}\annlab{boundary eqs}\hspace{0.3em}\annempty}
\annrow{basis}{\annlab{basis description}\hspace{0.3em}Energy eigenstates for bound and scattering states.\annsep{}\annlab{initial state}\hspace{0.3em}\annempty}
\annstep{Equation formulation}{step4\_\allowbreak equation\_\allowbreak formulation}
\annrow{laws}{Schrödinger Equation}
\annrow{dynamics}{}
{\setlength{\annind}{1.3em}
\annitem{Evolution/State Equation}
{\setlength{\annind}{2.6em}
\annline{\annlab{equation}\hspace{0.3em}$i\hbar \frac{\partial}{\partial t}\Psi = \hat{H}\Psi$}
\par}
\par}
\annrow{measurement}{\annempty}
\annstep{Synthesized output}{step5\_\allowbreak synthesized\_\allowbreak output}
\annrow{key phenomena}{Quantized energy levels for bound states and continuous spectrum for scattering states.}
\annrow{assumptions}{Time-independent potential}
\end{anncase}
\medskip

\section{Qualitative and quantitative comparisons}
\label{sec:qual}
\subsection{Qualitative analysis}
\label{sec:qual-analysis}

To complement the aggregate scores with direct visual evidence, we show, for nine prompts spanning the six subdisciplines, the diagram that every evaluated model produces from an identical caption (\Cref{fig:qual-mechanics-92,fig:qual-mechanics-16,fig:qual-optics-11,fig:qual-optics-153,fig:qual-electromagnetism-57,fig:qual-electromagnetism-168,fig:qual-physicalacoustics-11,fig:qual-quantummechanics-5,fig:qual-thermodynamics-0}). Each panel pairs the prompt with the generated figures and lists, for reference, the per-model faithfulness on that prompt's own question bank; the output of \ProjectName{} is outlined in the accent colour. The examples make concrete the failure modes summarised by the quantitative results: general-purpose systems tend to render a photorealistic or decorative scene that disregards the requested physical structure (forces, directions, coordinate axes, and governing relations), whereas \ProjectName{} reproduces the labelled quantities and the relationships between them that the prompt calls for.

\definecolor{qcardborder}{HTML}{D6DCE5}
\definecolor{qtextcol}{HTML}{5C6470}
\definecolor{qchipbg}{HTML}{E7EEFA}
\definecolor{qchipfg}{HTML}{2F6FE0}
\definecolor{qourscol}{HTML}{2F6FE0}
\newtcbox{\qchip}{on line, colback=qchipbg, colframe=qchipbg, boxrule=0pt, arc=2.5pt, boxsep=0pt, left=4pt, right=4pt, top=1.2pt, bottom=1.2pt, nobeforeafter, fontupper=\bfseries\footnotesize\color{qchipfg}}
\newlength{\qw}\setlength{\qw}{0.158\linewidth}
\newcommand{\qh}{1.95cm}
\newcommand{\qtile}[1]{\parbox[c][\dimexpr\qw-8pt\relax][c]{\dimexpr\qw-8pt\relax}{\centering\includegraphics[width=\dimexpr\qw-11pt\relax,height=\dimexpr\qw-11pt\relax,keepaspectratio]{#1}}}
\newcommand{\qcell}[2]{\begin{minipage}[t]{\qw}\centering{\setlength{\fboxsep}{2pt}\setlength{\fboxrule}{0.4pt}\fcolorbox{qcardborder}{white}{\qtile{#1}}}\\[1.5pt]{\scriptsize\color{qtextcol}#2}\end{minipage}}
\newcommand{\qcellO}[2]{\begin{minipage}[t]{\qw}\centering{\setlength{\fboxsep}{2pt}\setlength{\fboxrule}{1.3pt}\fcolorbox{qourscol}{white}{\qtile{#1}}}\\[1.5pt]{\scriptsize\color{qourscol}\textbf{#2}\\[0.2pt]\ProjectName{}-BAGEL}\end{minipage}}
\newcommand{\qcaseheader}[4]{\begin{minipage}{\linewidth}\qchip{#1}\par\vspace{2pt}{\small #3}\end{minipage}\par\vspace{4pt}}

\begin{figure}[H]\centering
\qcaseheader{Mechanics}{92}{The diagram shows a horizontal beam supported at points~A and~B. A point load of 2500~lb acts vertically downward at the left end; a distributed load of 75~lb/ft acts between points~C and~D; and a further point load of 3000~lb acts vertically downward near the right end. The support and load spacings are marked 6~ft, 12~ft, and 2~ft.}{Ours \textbf{81.5\%} \textperiodcentered\ Qwen 24.1\% \textperiodcentered\ FLUX 18.5\% \textperiodcentered\ 54~QA}
\setlength{\tabcolsep}{2pt}
\begin{tabular}{@{}cccccc@{}}
\qcell{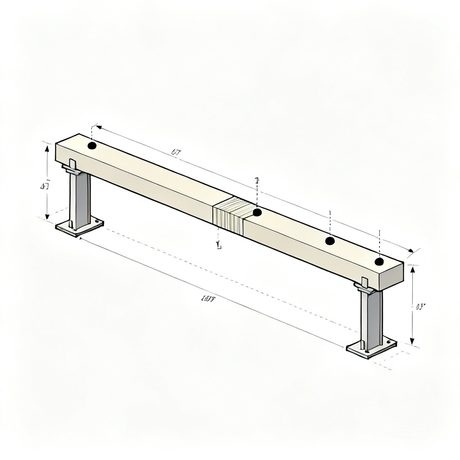}{Seedream 4.0} & \qcell{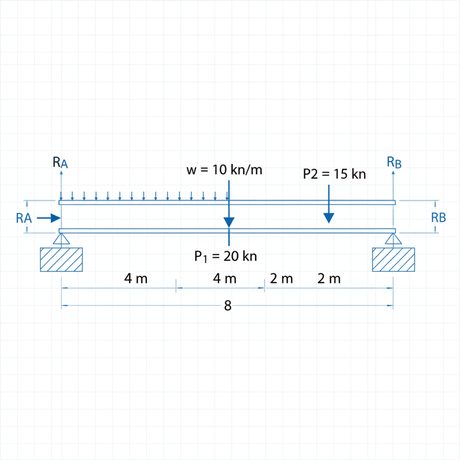}{Gemini 2.5 Flash} & \qcell{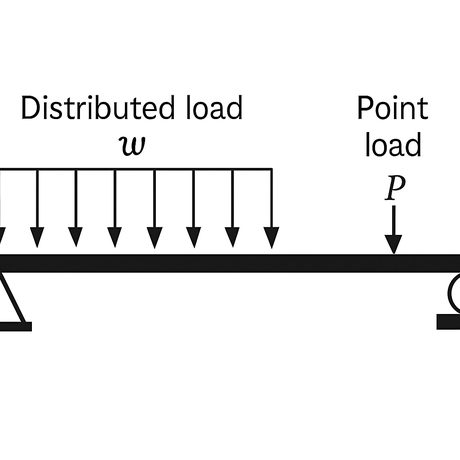}{GPT-Image-1} & \qcell{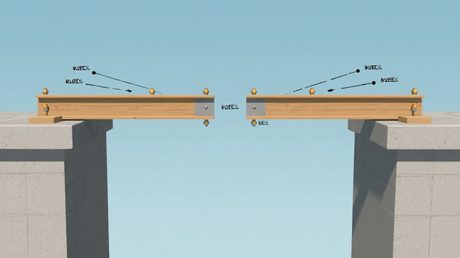}{Qwen-Image} &
\qcell{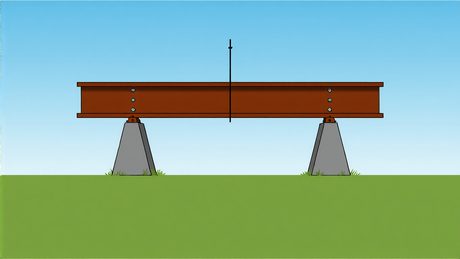}{HiDream-I1} & \qcell{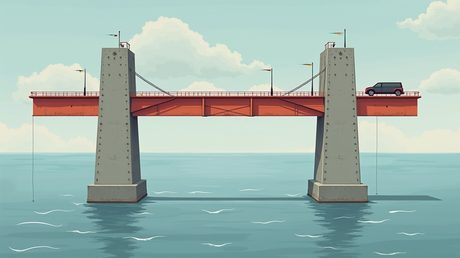}{FLUX.1 dev} \\[2.5pt]\qcell{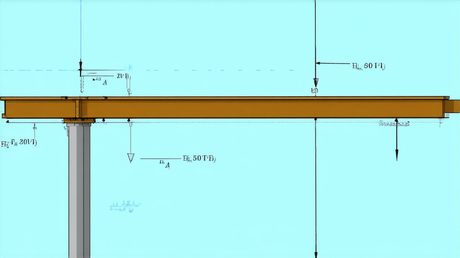}{SD 3.5 Large} & \qcell{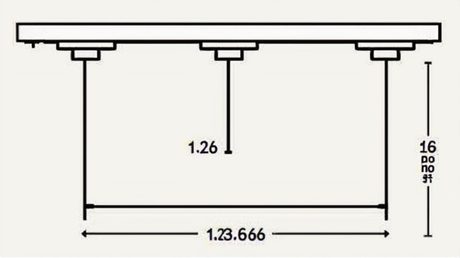}{BAGEL} &
\qcell{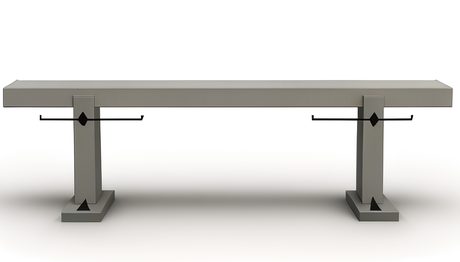}{DiMOO} & \qcell{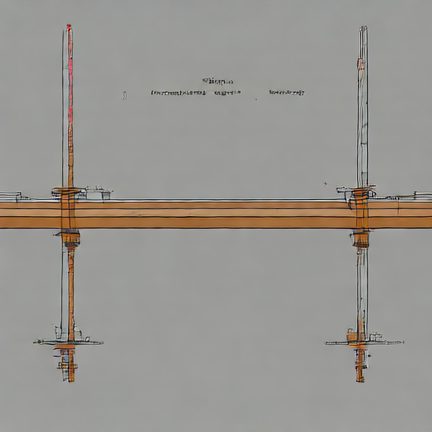}{Show-o2} & \qcell{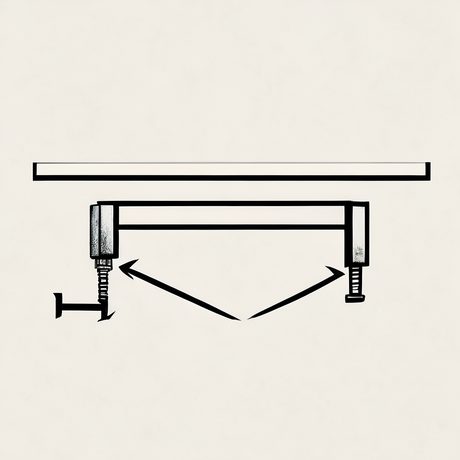}{BLIP3o} & \qcellO{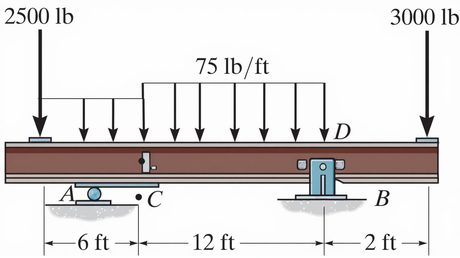}{Ours} \\
\end{tabular}
\caption{\textbf{Mechanics}: Qualitative comparison between our method and competing methods.}
\label{fig:qual-mechanics-92}
\end{figure}

\begin{figure}[H]\centering
\qcaseheader{Mechanics}{16}{Two blocks $m_1$ and $m_2$ on a horizontal surface are connected by a rope. Block $m_2$ is pulled by an external force $\vec{F}$ directed to the right, and the rope tension is labelled $T$.}{Ours \textbf{90.6\%} \textperiodcentered\ Qwen 30.2\% \textperiodcentered\ FLUX 30.2\% \textperiodcentered\ 53~QA}
\setlength{\tabcolsep}{2pt}
\begin{tabular}{@{}cccccc@{}}
\qcell{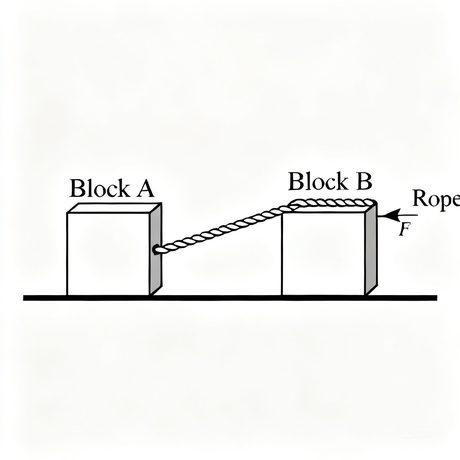}{Seedream 4.0} & \qcell{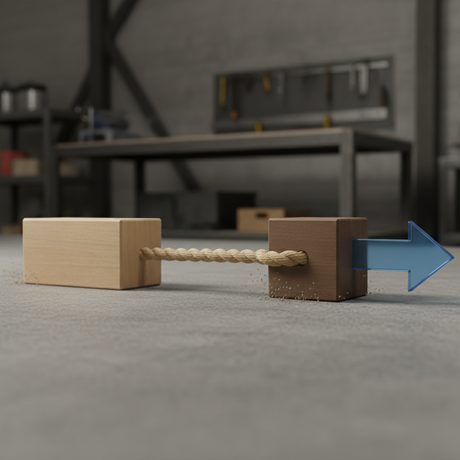}{Gemini 2.5 Flash} & \qcell{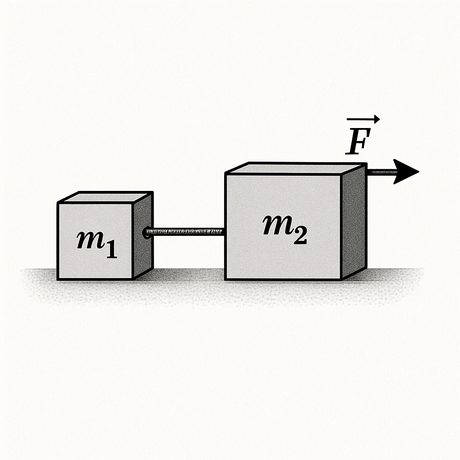}{GPT-Image-1} & \qcell{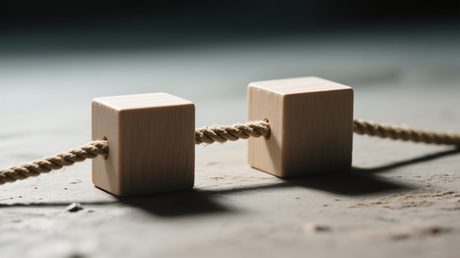}{Qwen-Image} &
\qcell{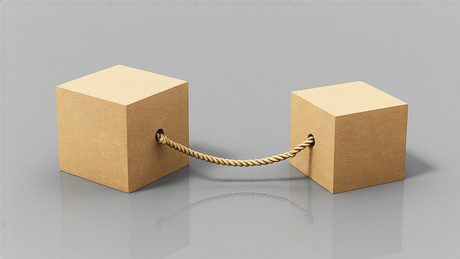}{HiDream-I1} & \qcell{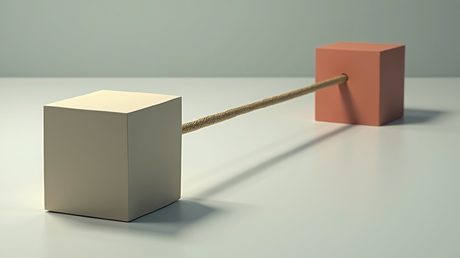}{FLUX.1 dev} \\[2.5pt]\qcell{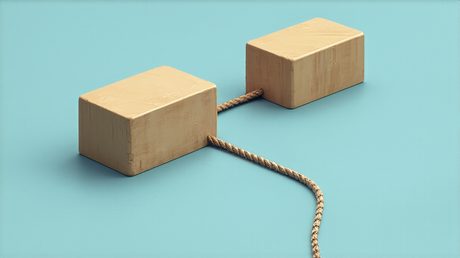}{SD 3.5 Large} & \qcell{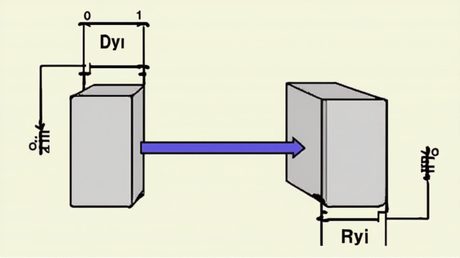}{BAGEL} &
\qcell{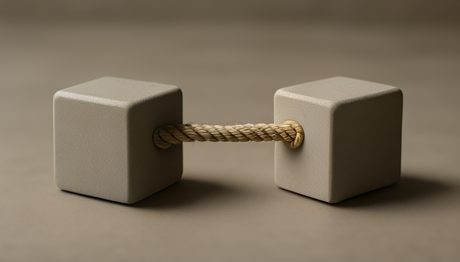}{DiMOO} & \qcell{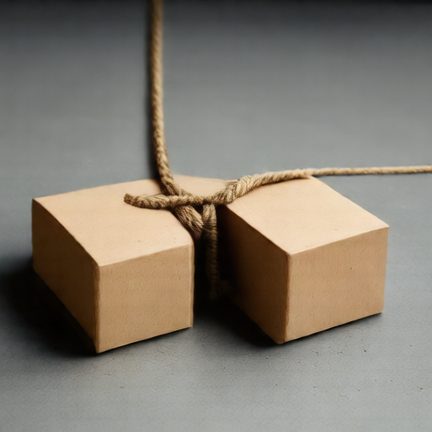}{Show-o2} & \qcell{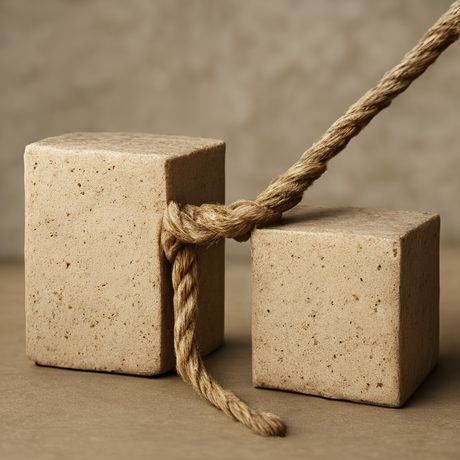}{BLIP3o} & \qcellO{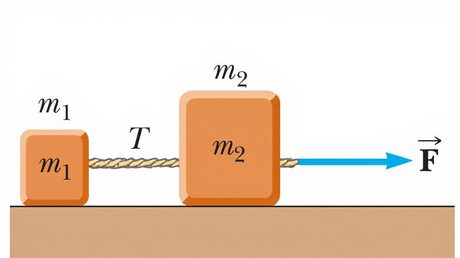}{Ours} \\
\end{tabular}
\caption{\textbf{Mechanics}: Qualitative comparison between our method and competing methods.}
\label{fig:qual-mechanics-16}
\end{figure}

\begin{figure}[H]\centering
\qcaseheader{Physical optics}{11}{A horizontal interface separates air above from a transparent medium below (light-blue shading). A dashed vertical line marks the surface normal at the point of incidence. An incident ray arrives from the upper right at angle $\theta$ to the normal and refracts into the lower medium, bending toward the normal as it enters the higher-index material; no reflected ray is drawn.}{Ours \textbf{96.3\%} \textperiodcentered\ Qwen 34.1\% \textperiodcentered\ FLUX 56.1\% \textperiodcentered\ 82~QA}
\setlength{\tabcolsep}{2pt}
\begin{tabular}{@{}cccccc@{}}
\qcell{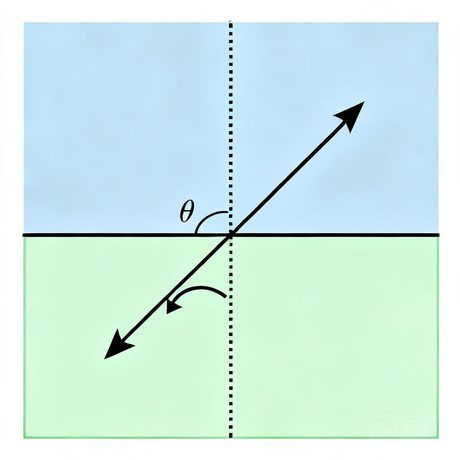}{Seedream 4.0} & \qcell{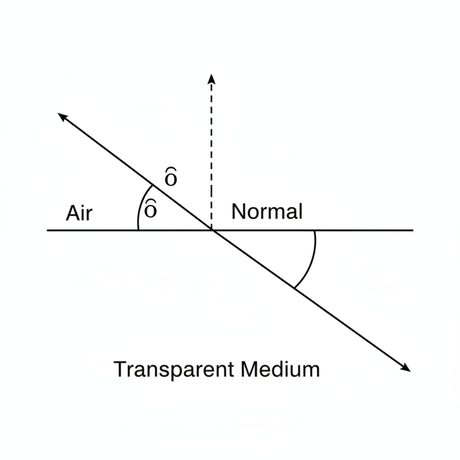}{Gemini 2.5 Flash} & \qcell{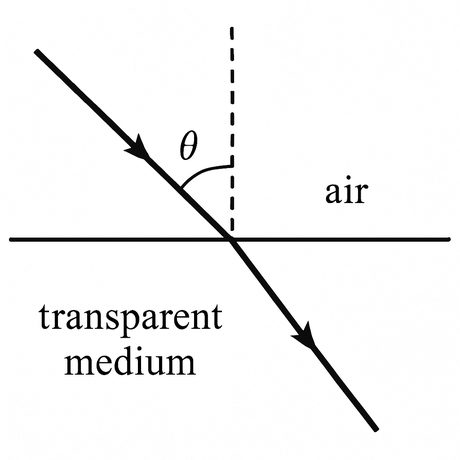}{GPT-Image-1} & \qcell{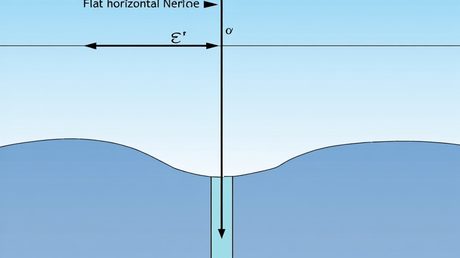}{Qwen-Image} &
\qcell{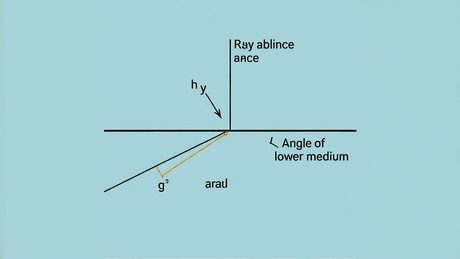}{HiDream-I1} & \qcell{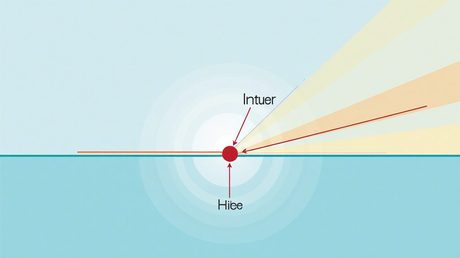}{FLUX.1 dev} \\[2.5pt]\qcell{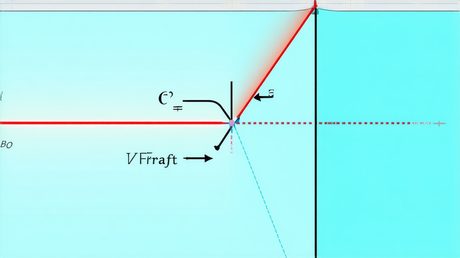}{SD 3.5 Large} & \qcell{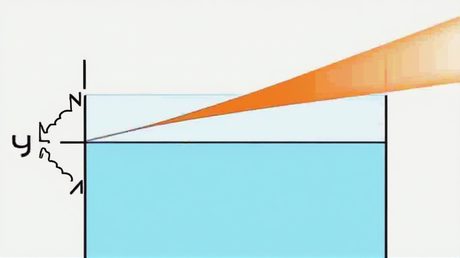}{BAGEL} &
\qcell{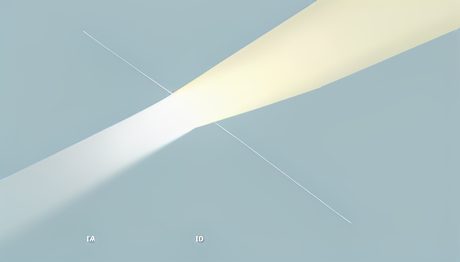}{DiMOO} & \qcell{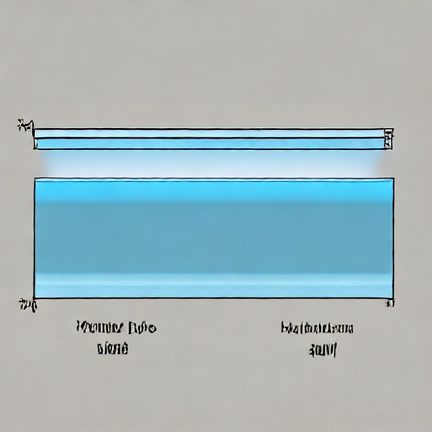}{Show-o2} & \qcell{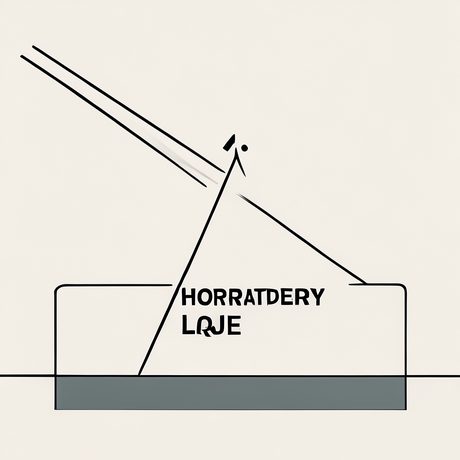}{BLIP3o} & \qcellO{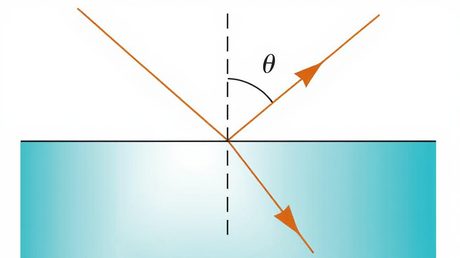}{Ours} \\
\end{tabular}
\caption{\textbf{Physical optics}: Qualitative comparison between our method and competing methods.}
\label{fig:qual-optics-11}
\end{figure}

\begin{figure}[H]\centering
\qcaseheader{Physical optics}{153}{A collimated beam of wavelength $\lambda$ strikes a narrow vertical slit of width $a$ formed by two plates. Dashed rays radiate from the slit toward a screen on the right, which shows a red intensity envelope with a central maximum and side lobes. Two angular positions are marked, $\theta_2 = 45^\circ$ and a smaller $\theta_1$, measured from the central axis.}{Ours \textbf{87.4\%} \textperiodcentered\ Qwen 55.3\% \textperiodcentered\ FLUX 30.1\% \textperiodcentered\ 103~QA}
\setlength{\tabcolsep}{2pt}
\begin{tabular}{@{}cccccc@{}}
\qcell{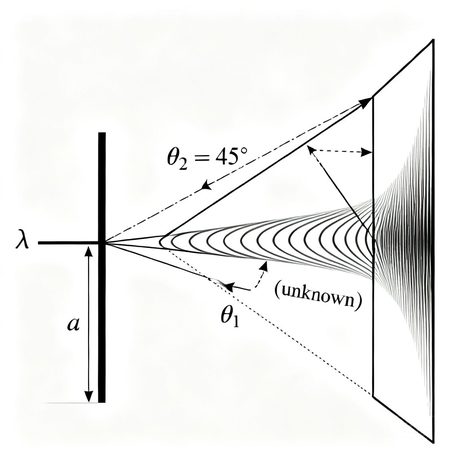}{Seedream 4.0} & \qcell{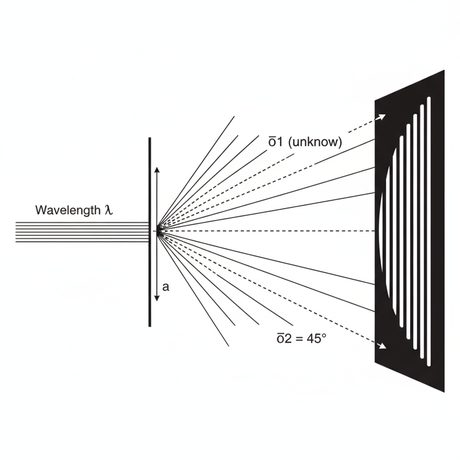}{Gemini 2.5 Flash} & \qcell{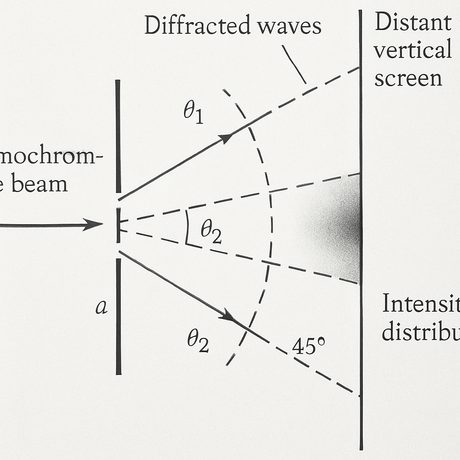}{GPT-Image-1} & \qcell{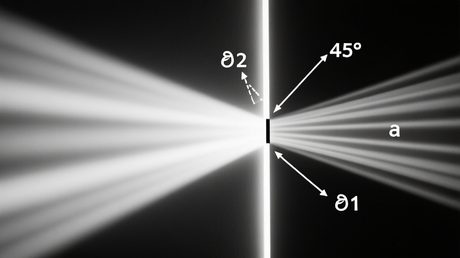}{Qwen-Image} &
\qcell{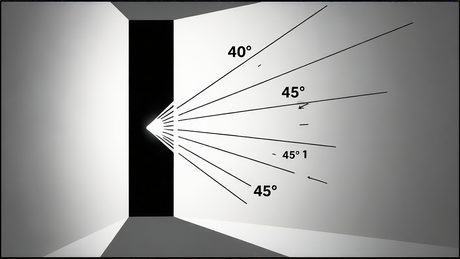}{HiDream-I1} & \qcell{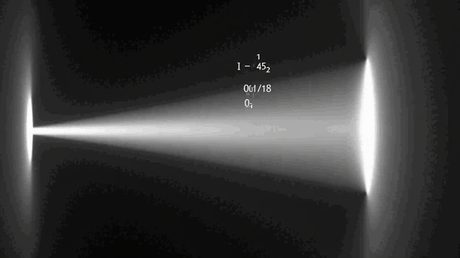}{FLUX.1 dev} \\[2.5pt]\qcell{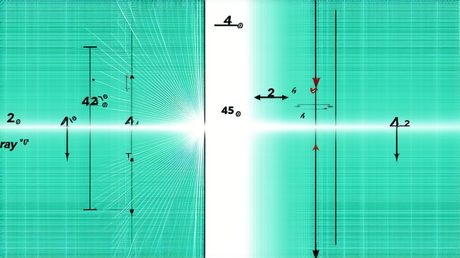}{SD 3.5 Large} & \qcell{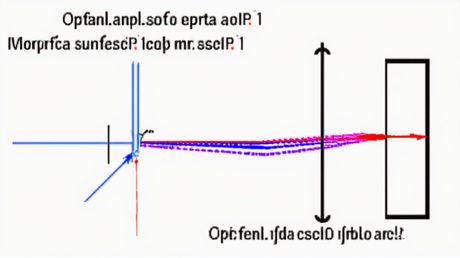}{BAGEL} &
\qcell{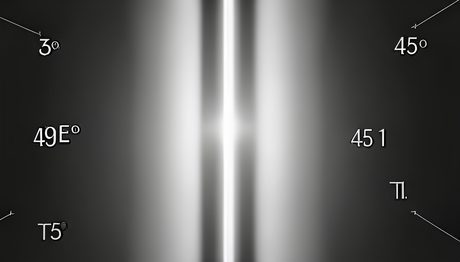}{DiMOO} & \qcell{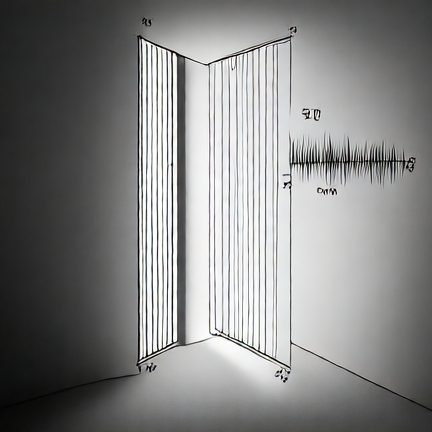}{Show-o2} & \qcell{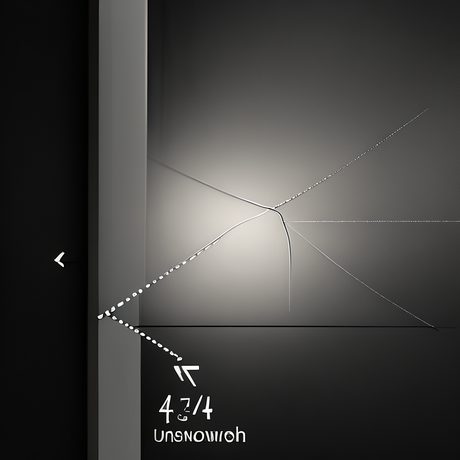}{BLIP3o} & \qcellO{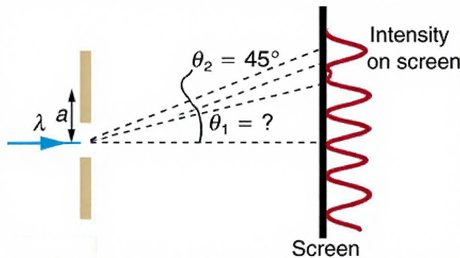}{Ours} \\
\end{tabular}
\caption{\textbf{Physical optics}: Qualitative comparison between our method and competing methods.}
\label{fig:qual-optics-153}
\end{figure}

\begin{figure}[H]\centering
\qcaseheader{Electromagnetism}{57}{An operational amplifier drives a 200~mV source through a $10\,\mathrm{k}\Omega$ resistor into the inverting input, with the non-inverting input grounded and $\pm 15$~V supplies. The feedback network uses $40\,\mathrm{k}\Omega$ and $20\,\mathrm{k}\Omega$ resistors, with a $30\,\mathrm{k}\Omega$ resistor to ground at their junction. The output voltage $v_o$ appears across a $10\,\mathrm{k}\Omega$ load, and current directions $i_a$ and $i_o$ are indicated.}{Ours \textbf{85.4\%} \textperiodcentered\ Qwen 24.4\% \textperiodcentered\ FLUX 25.6\% \textperiodcentered\ 82~QA}
\setlength{\tabcolsep}{2pt}
\begin{tabular}{@{}cccccc@{}}
\qcell{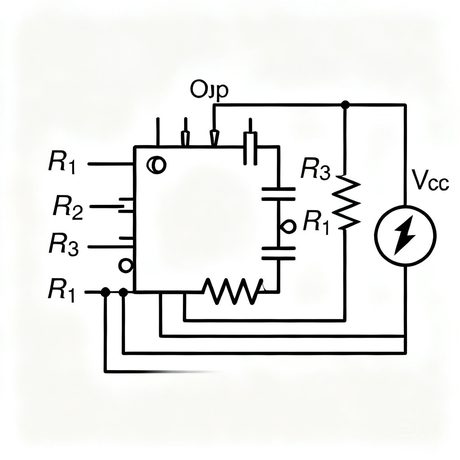}{Seedream 4.0} & \qcell{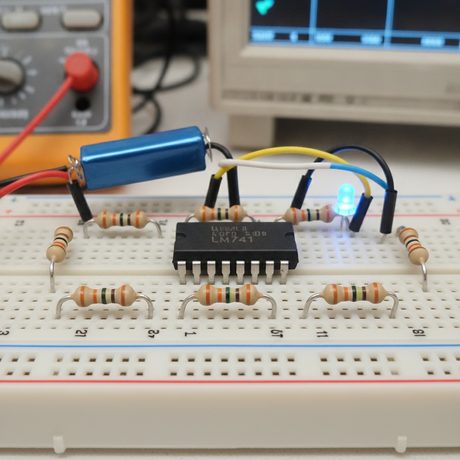}{Gemini 2.5 Flash} & \qcell{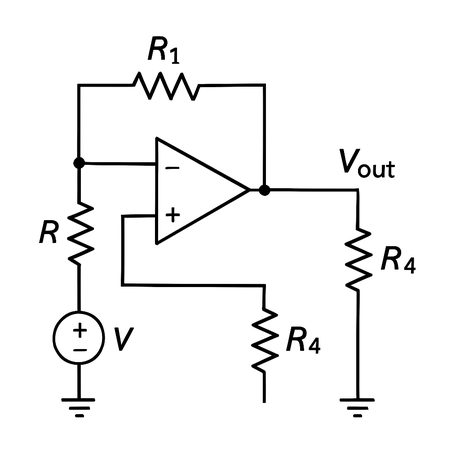}{GPT-Image-1} & \qcell{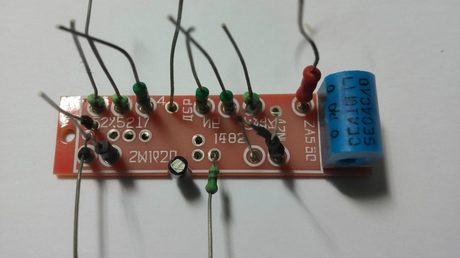}{Qwen-Image} &
\qcell{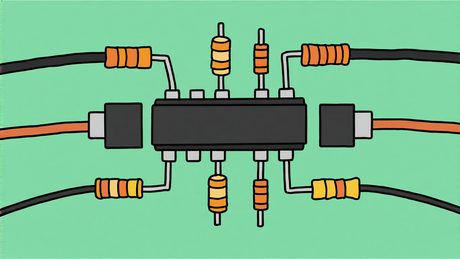}{HiDream-I1} & \qcell{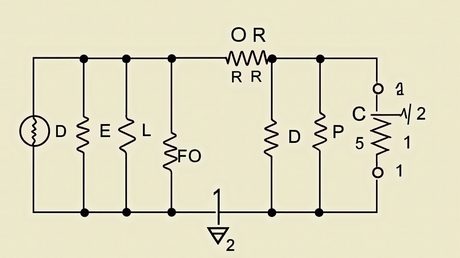}{FLUX.1 dev} \\[2.5pt]\qcell{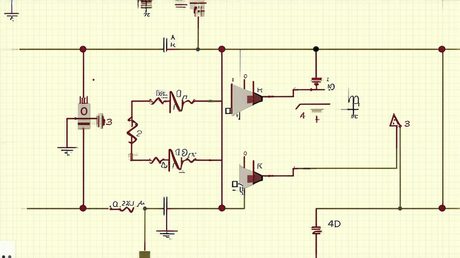}{SD 3.5 Large} & \qcell{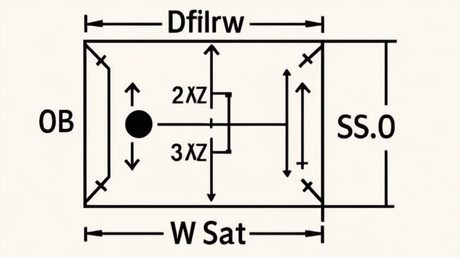}{BAGEL} &
\qcell{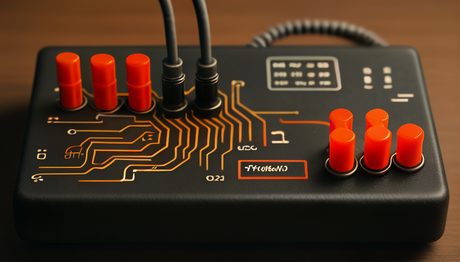}{DiMOO} & \qcell{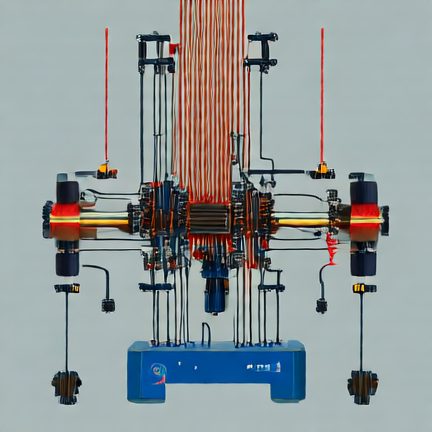}{Show-o2} & \qcell{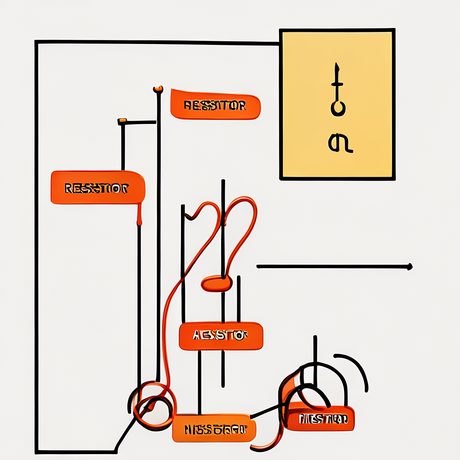}{BLIP3o} & \qcellO{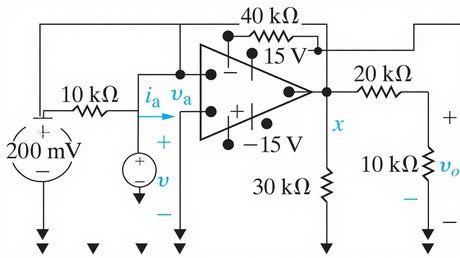}{Ours} \\
\end{tabular}
\caption{\textbf{Electromagnetism}: Qualitative comparison between our method and competing methods.}
\label{fig:qual-electromagnetism-57}
\end{figure}

\begin{figure}[H]\centering
\qcaseheader{Electromagnetism}{168}{An electron ($e^-$) enters the region between two parallel plates with initial velocity $v_0$ along the $x$-axis. The upper plate is positive and the lower negative, giving a uniform downward electric field, so the electron's path curves upward and it exits with velocity $v_1$ at an angle $\theta$.}{Ours \textbf{92.6\%} \textperiodcentered\ Qwen 37.0\% \textperiodcentered\ FLUX 37.0\% \textperiodcentered\ 54~QA}
\setlength{\tabcolsep}{2pt}
\begin{tabular}{@{}cccccc@{}}
\qcell{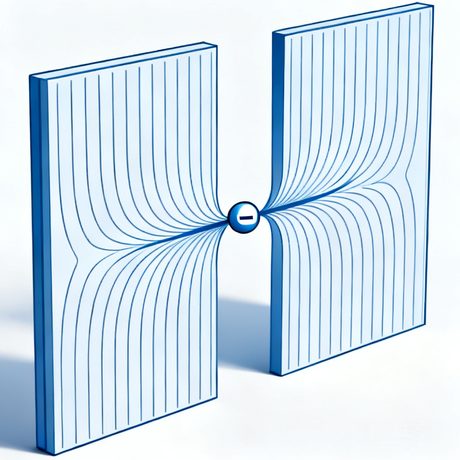}{Seedream 4.0} & \qcell{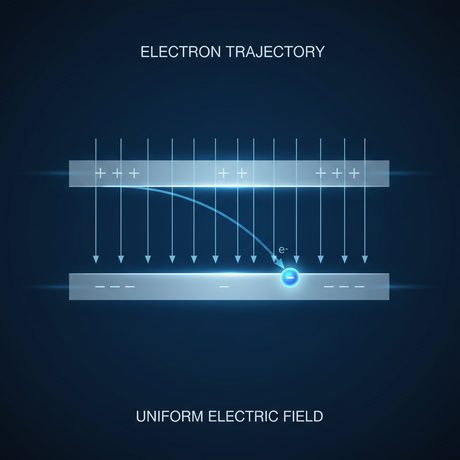}{Gemini 2.5 Flash} & \qcell{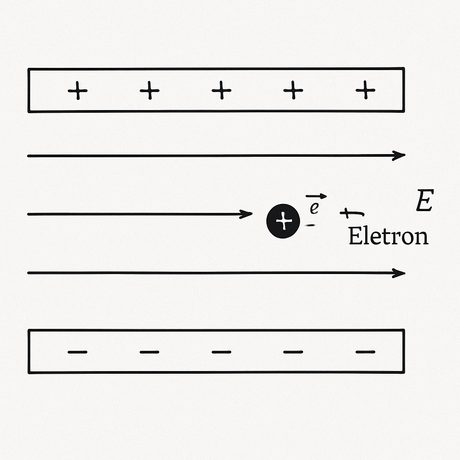}{GPT-Image-1} & \qcell{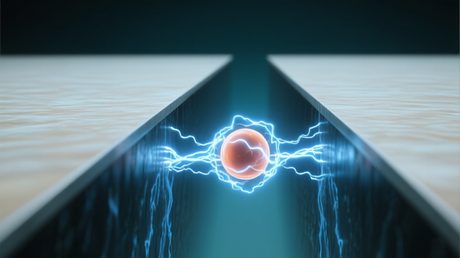}{Qwen-Image} &
\qcell{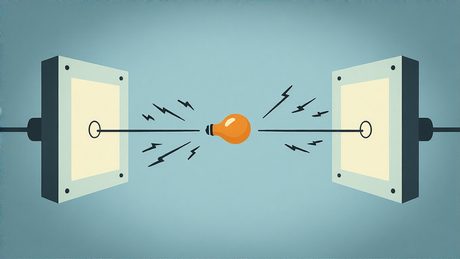}{HiDream-I1} & \qcell{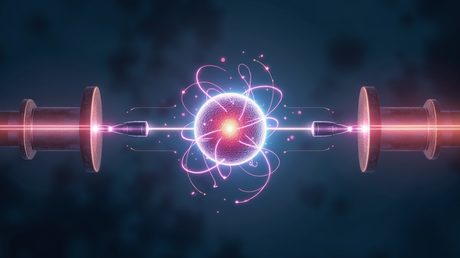}{FLUX.1 dev} \\[2.5pt]\qcell{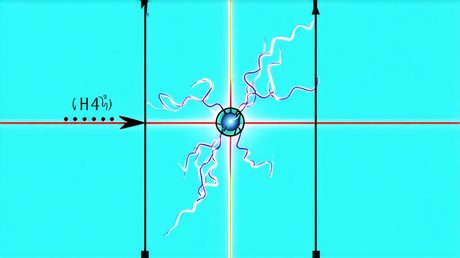}{SD 3.5 Large} & \qcell{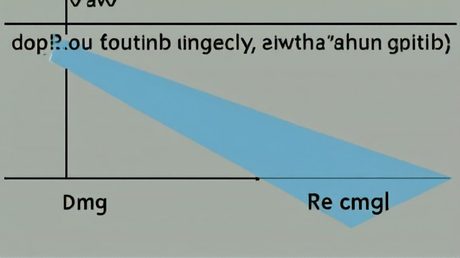}{BAGEL} &
\qcell{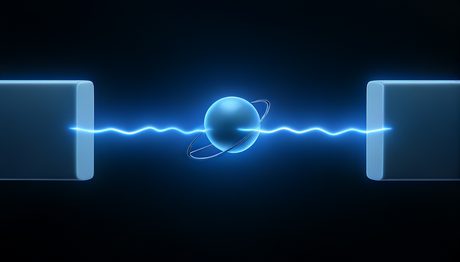}{DiMOO} & \qcell{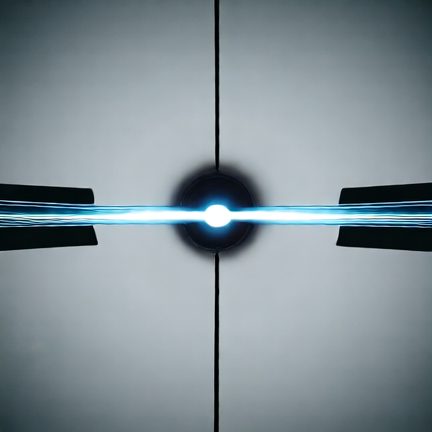}{Show-o2} & \qcell{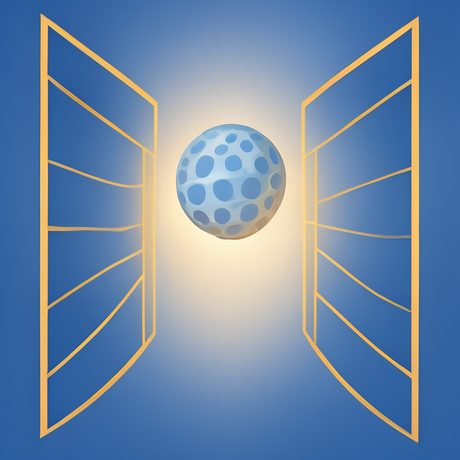}{BLIP3o} & \qcellO{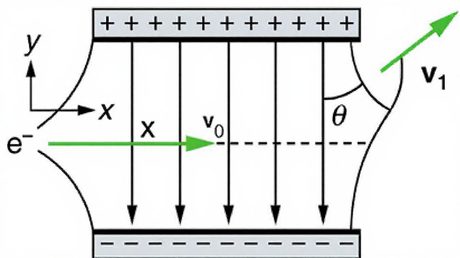}{Ours} \\
\end{tabular}
\caption{\textbf{Electromagnetism}: Qualitative comparison between our method and competing methods.}
\label{fig:qual-electromagnetism-168}
\end{figure}

\begin{figure}[H]\centering
\qcaseheader{Acoustics}{11}{A cylindrical tube is driven by a piston on the left moving with velocity $v_p(t)$. The tube has length $L$ and diameter $2a$, and an arrow indicates the direction of the piston's motion.}{Ours \textbf{86.5\%} \textperiodcentered\ Qwen 25.0\% \textperiodcentered\ FLUX 26.9\% \textperiodcentered\ 52~QA}
\setlength{\tabcolsep}{2pt}
\begin{tabular}{@{}cccccc@{}}
\qcell{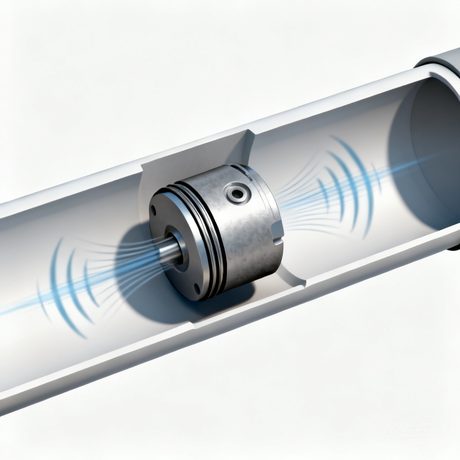}{Seedream 4.0} & \qcell{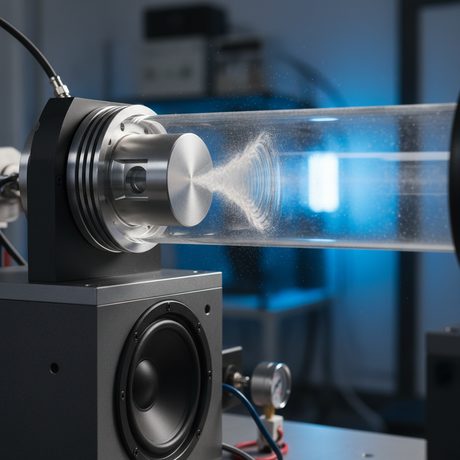}{Gemini 2.5 Flash} & \qcell{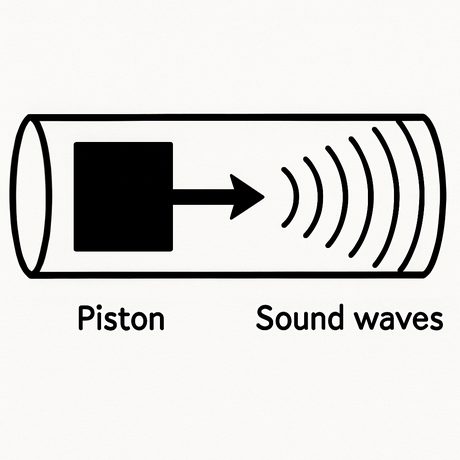}{GPT-Image-1} & \qcell{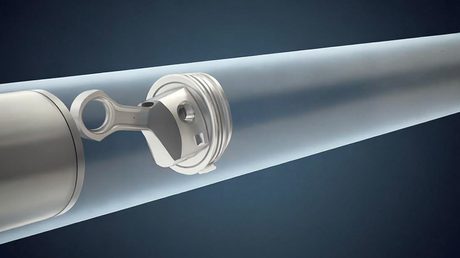}{Qwen-Image} &
\qcell{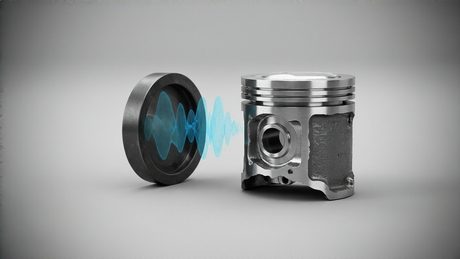}{HiDream-I1} & \qcell{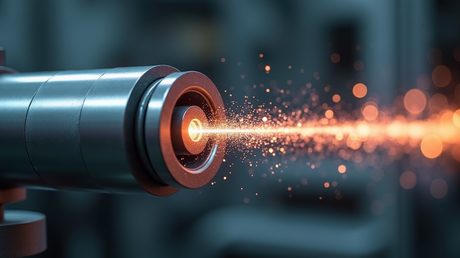}{FLUX.1 dev} \\[2.5pt]\qcell{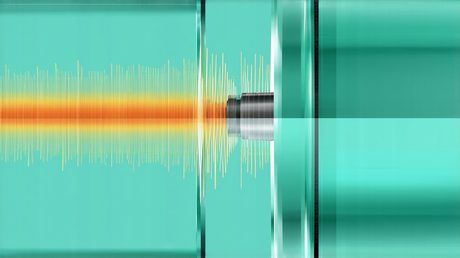}{SD 3.5 Large} & \qcell{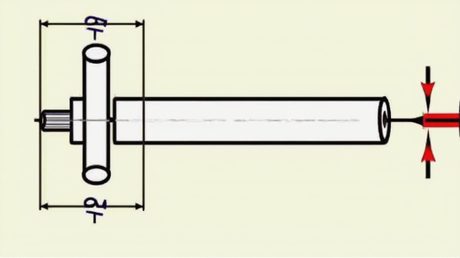}{BAGEL} &
\qcell{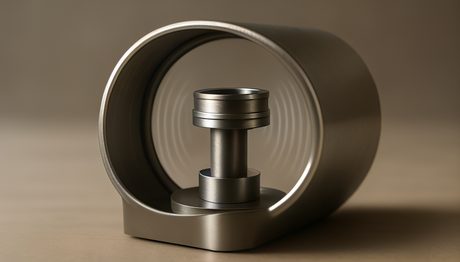}{DiMOO} & \qcell{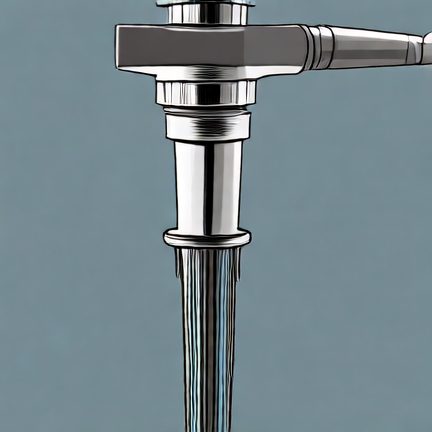}{Show-o2} & \qcell{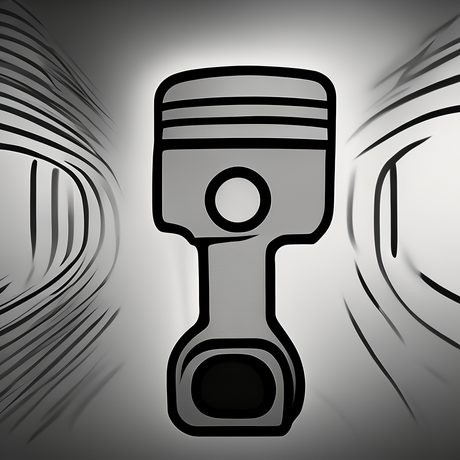}{BLIP3o} & \qcellO{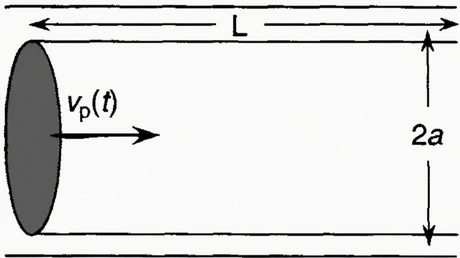}{Ours} \\
\end{tabular}
\caption{\textbf{Acoustics}: Qualitative comparison between our method and competing methods.}
\label{fig:qual-physicalacoustics-11}
\end{figure}

\begin{figure}[H]\centering
\qcaseheader{Quantum mechanics}{5}{A one-dimensional potential well has three regions: regions~I and~III are infinite potential barriers and region~II has zero potential. The particle's total energy is $E_{\mathrm{tot}}$, the potential $V(x)$ is infinite in~I and~III and zero in~II, and the well width is $L$.}{Ours \textbf{83.3\%} \textperiodcentered\ Qwen 29.2\% \textperiodcentered\ FLUX 29.2\% \textperiodcentered\ 48~QA}
\setlength{\tabcolsep}{2pt}
\begin{tabular}{@{}cccccc@{}}
\qcell{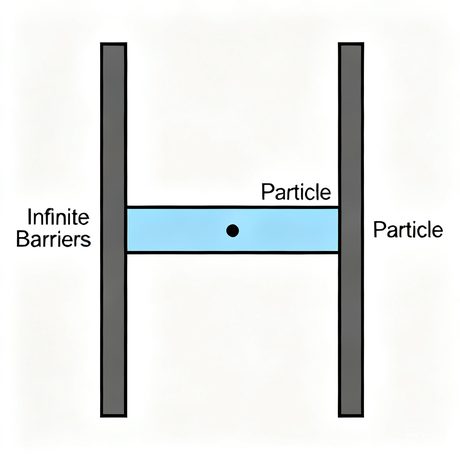}{Seedream 4.0} & \qcell{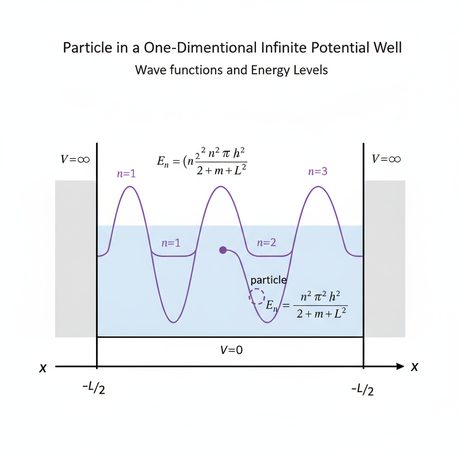}{Gemini 2.5 Flash} & \qcell{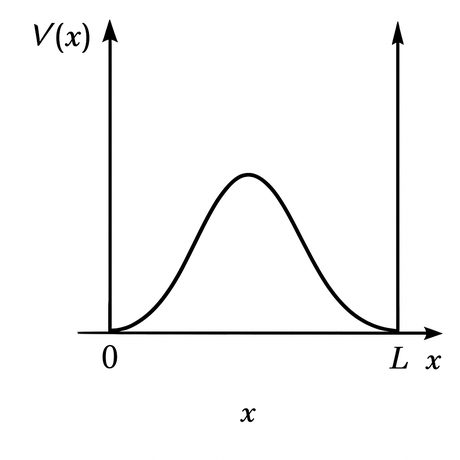}{GPT-Image-1} & \qcell{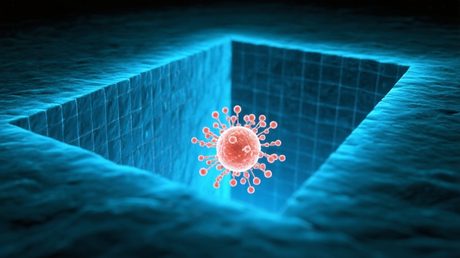}{Qwen-Image} &
\qcell{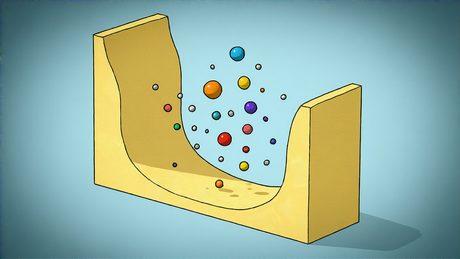}{HiDream-I1} & \qcell{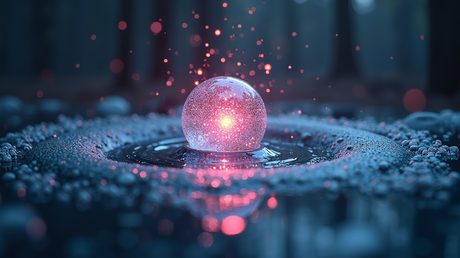}{FLUX.1 dev} \\[2.5pt]\qcell{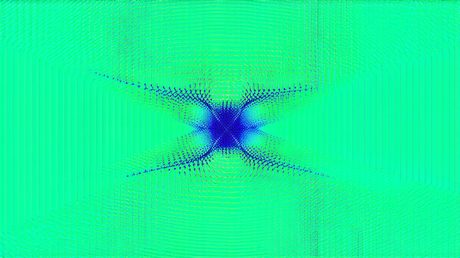}{SD 3.5 Large} & \qcell{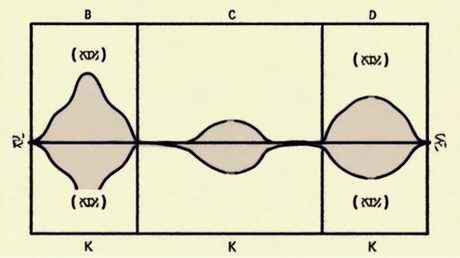}{BAGEL} &
\qcell{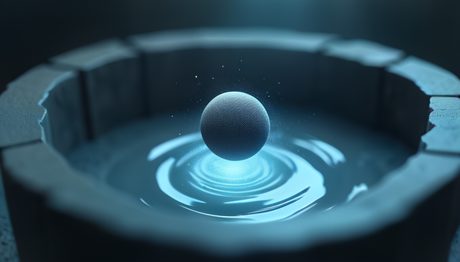}{DiMOO} & \qcell{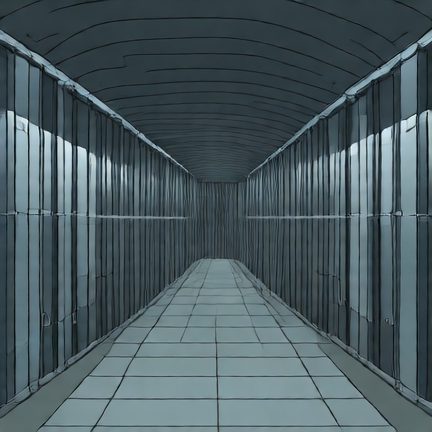}{Show-o2} & \qcell{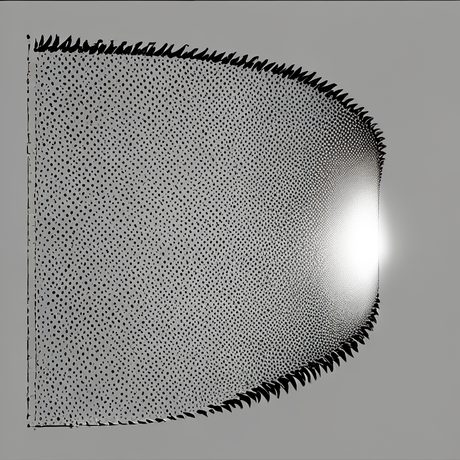}{BLIP3o} & \qcellO{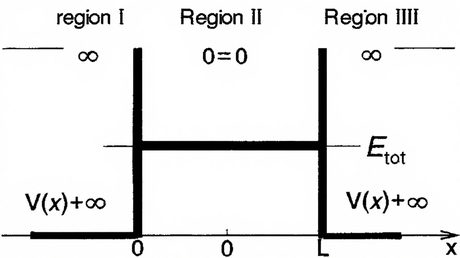}{Ours} \\
\end{tabular}
\caption{\textbf{Quantum mechanics}: Qualitative comparison between our method and competing methods.}
\label{fig:qual-quantummechanics-5}
\end{figure}

\begin{figure}[H]\centering
\qcaseheader{Thermodynamics}{0}{A closed system consisting of a cylindrical container exchanges heat with its surroundings: $Q_{\mathrm{in}} = 15$~kJ enters and $Q_{\mathrm{out}} = 3$~kJ leaves. The net energy change is $\Delta E = Q_{\mathrm{net}} = 12$~kJ.}{Ours \textbf{91.2\%} \textperiodcentered\ Qwen 23.5\% \textperiodcentered\ FLUX 23.5\% \textperiodcentered\ 34~QA}
\setlength{\tabcolsep}{2pt}
\begin{tabular}{@{}cccccc@{}}
\qcell{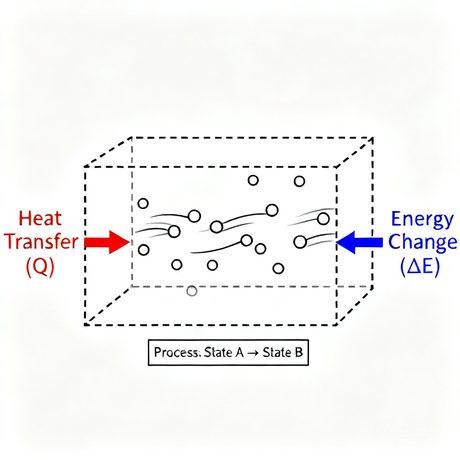}{Seedream 4.0} & \qcell{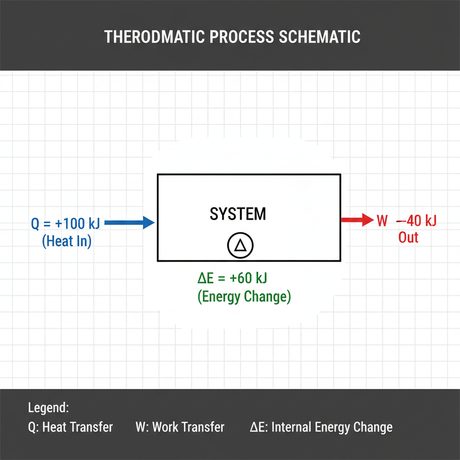}{Gemini 2.5 Flash} & \qcell{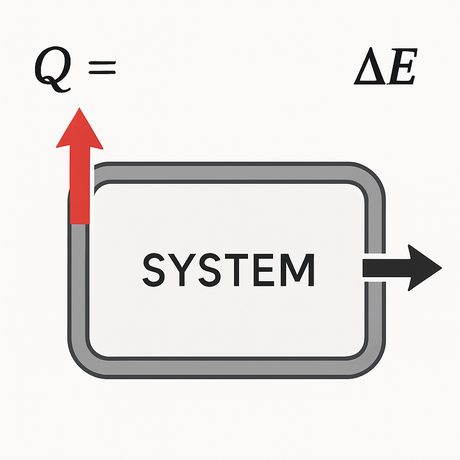}{GPT-Image-1} & \qcell{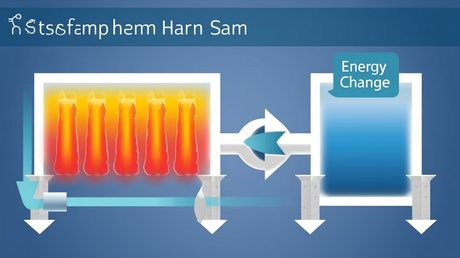}{Qwen-Image} &
\qcell{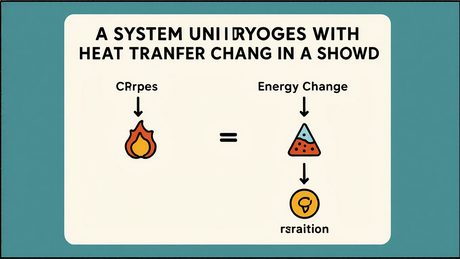}{HiDream-I1} & \qcell{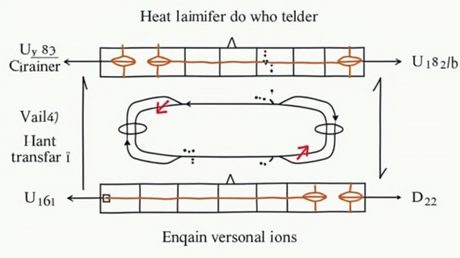}{FLUX.1 dev} \\[2.5pt]\qcell{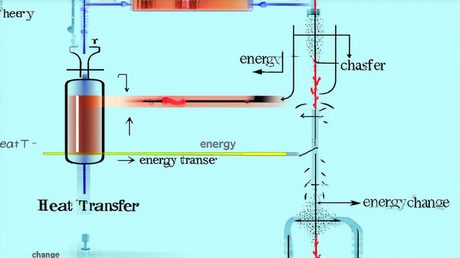}{SD 3.5 Large} & \qcell{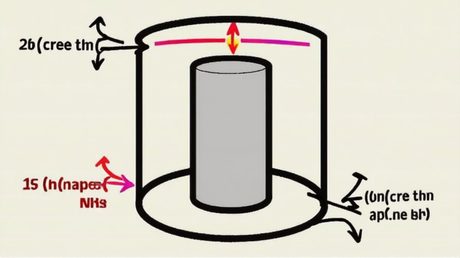}{BAGEL} &
\qcell{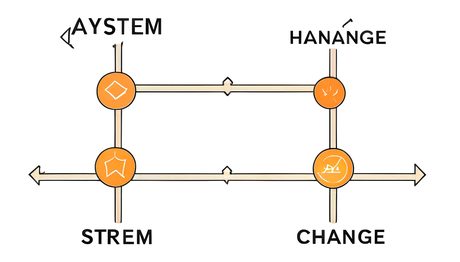}{DiMOO} & \qcell{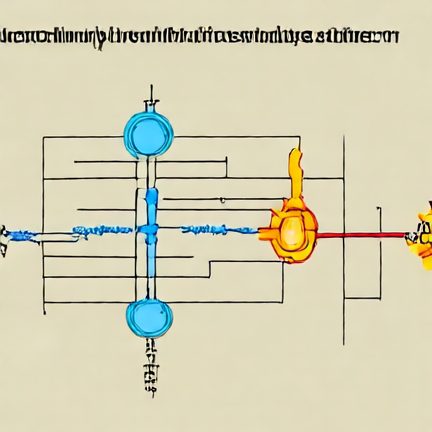}{Show-o2} & \qcell{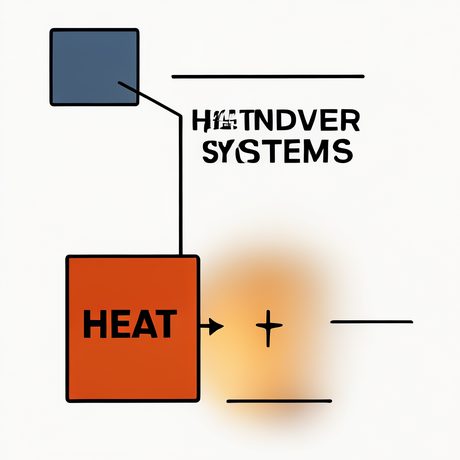}{BLIP3o} & \qcellO{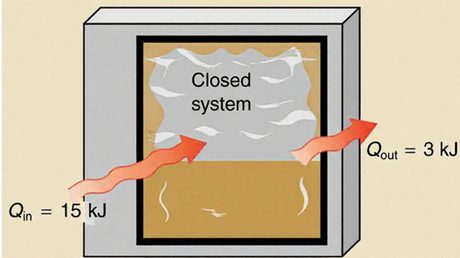}{Ours} \\
\end{tabular}
\caption{\textbf{Thermodynamics}: Qualitative comparison between our method and competing methods.}
\label{fig:qual-thermodynamics-0}
\end{figure}

\subsection{Checklist-grounded per-attribute evaluation}
\label{sec:qual-checklist}

\begin{figure}[!t]
\centering
\includegraphics[width=0.9\linewidth]{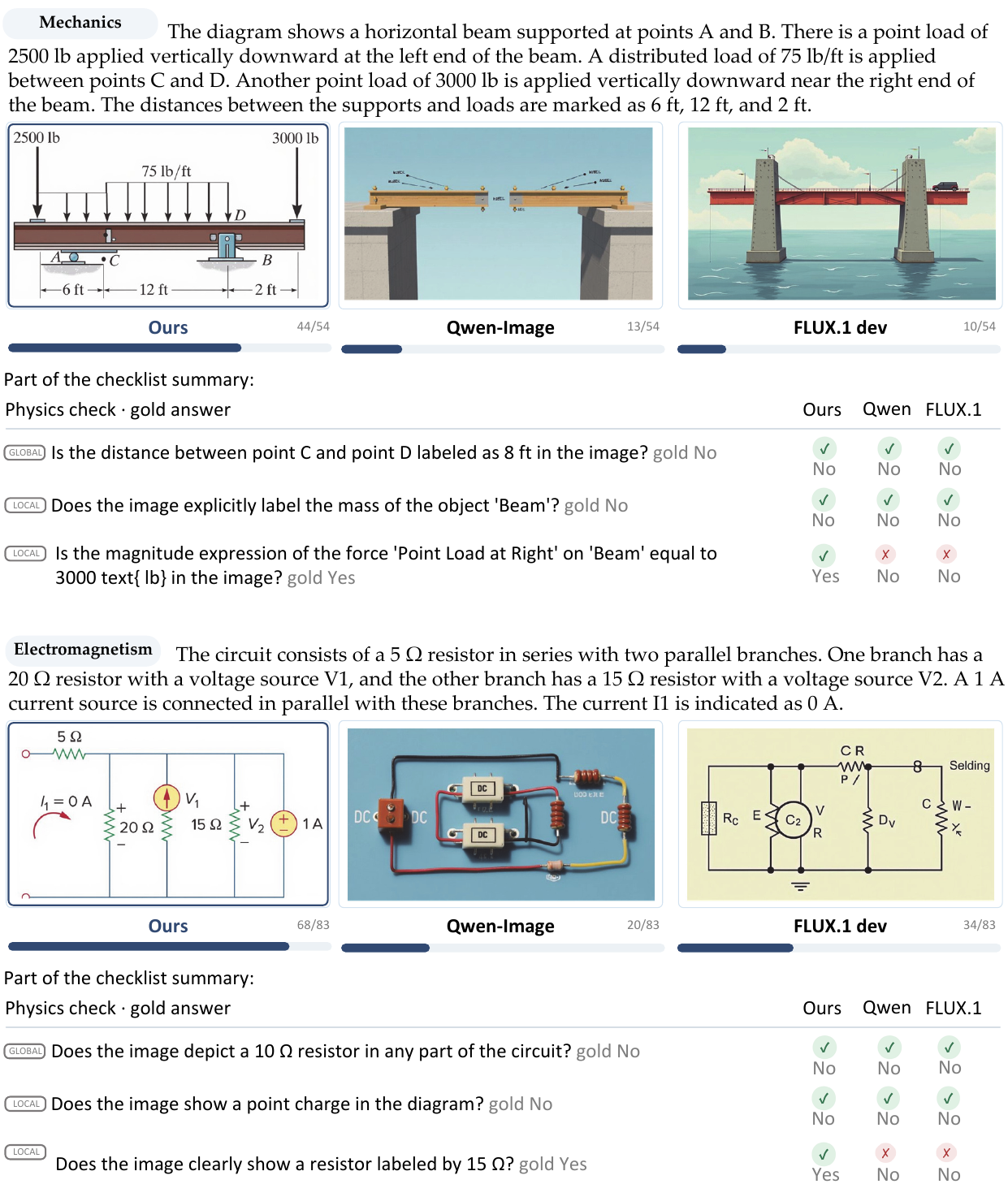}
\caption{\textbf{Checklist-grounded evaluation.} Generated diagrams are assessed using expert-verified \BenchName{} criteria. Checks and crosses indicate criterion satisfaction, while fractions report local-criterion accuracy.}
\label{fig:qual-checklist}
\end{figure}

Beyond the side-by-side images of \Cref{sec:qual-analysis}, \Cref{fig:qual-checklist} shows how \BenchName{} turns each generation into a quantitative, per-attribute verdict. For two representative prompts (a mechanics retaining-wall diagram and an electromagnetism circuit), it pairs the diagrams produced by \ProjectName{}, Qwen-Image, and FLUX.1~dev with an excerpt of the item's compiled checklist: each row is a binary physics question, tagged \textsc{global} or \textsc{local} and carrying its expert-verified gold answer, followed by the judge's Yes/No response for every model, marked correct or incorrect. The per-image counts (for the mechanics prompt, $44/54$ for \ProjectName{} against $13/54$ and $10/54$) are the local-bank scores of \Cref{eq:m-scores} made concrete: the same visible errors seen in the images translate directly into failed checklist items, which is what the aggregate scores of \Cref{sec:res-inhouse} summarise.

\subsection{Ablating the structured annotation}
\label{sec:qual-ablation}

The quantitative results attribute \ProjectName{}'s gains to the structured supervision rather than to the backbone or the image data; visually the effect is stark. \Cref{fig:qual-ablation} isolates \SPCoT{} as a single ingredient: the same model and pipeline are trained \emph{with} and \emph{without} the structured physical annotation (everything else held fixed), and both regimes render three mechanics prompts. Stripped of the structured labels, the model degenerates into plausible but physically meaningless artwork: the scene drifts (a child on a staircase for a pulley prompt), the geometry and topology are wrong, and every on-figure symbol collapses into illegible gibberish. Trained with \SPCoT{}, the same architecture places each mass, pulley, angle, and velocity where the prompt demands and renders every symbol ($m_1$, $m_2$, $\theta$, $8.00$~kg, $6.00$~kg, $v_1$) legibly. Structured physical labeling, not photorealism, is what makes the generations faithful.

\definecolor{abred}{HTML}{B24A38}
\definecolor{abteal}{HTML}{137F71}
\definecolor{abink}{HTML}{132029}
\definecolor{abnote}{HTML}{42535B}
\newtcbox{\abchip}{on line, colback=abink, colframe=abink, boxrule=0pt, arc=6pt, boxsep=0pt, left=7pt, right=7pt, top=1.5pt, bottom=1.5pt, nobeforeafter, fontupper=\sffamily\bfseries\footnotesize\color{white}}
\newlength{\abpw}\setlength{\abpw}{0.46\linewidth}
\newcommand{\abih}{4.1cm}
\newcommand{\abtag}[2]{{\sffamily\small\bfseries\color{#1}#2}}
\newcommand{\abframe}[2]{\setlength{\fboxsep}{0pt}\setlength{\fboxrule}{1.3pt}\fcolorbox{#1}{white}{\parbox[c][\abih][c]{\dimexpr\abpw-3pt\relax}{\centering\includegraphics[width=\dimexpr\abpw-6pt\relax,height=\dimexpr\abih-4pt\relax,keepaspectratio]{#2}}}}
\newcommand{\abx}[1]{\leavevmode{\color{abred}\ding{55}}\,#1\par}
\newcommand{\abck}[1]{\leavevmode{\color{abteal}\ding{51}}\,#1\par}
\newcommand{\abnotes}[1]{\begin{minipage}[t]{\abpw}\scriptsize\raggedright\color{abnote}#1\end{minipage}}

\begin{figure}[!t]\centering
{\abchip{Case 1}\hspace{0.6em}\small A pulley system with two masses: mass $m_1$ rests on an inclined plane at angle $\theta$, connected by a rope over a pulley to a hanging mass $m_2$.\par}
\smallskip
\begin{tabular}{@{}c@{\hspace{0.03\linewidth}}c@{}}
\abtag{abred}{Without SP-CoT} & \abtag{abteal}{With SP-CoT} \\[2pt]
\abframe{abred}{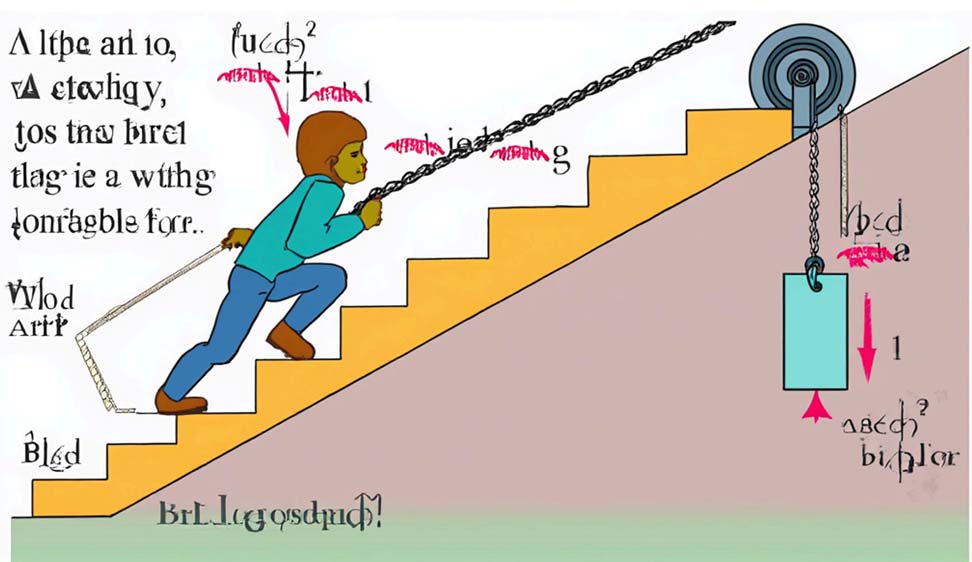} & \abframe{abteal}{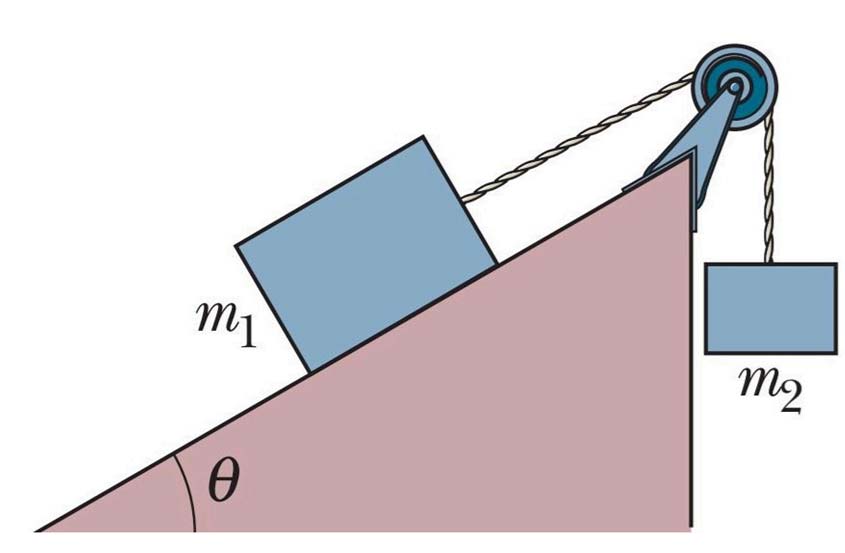} \\[3pt]
\abnotes{\abx{Scene drifts to a child on a staircase}\abx{Annotations are illegible gibberish}\abx{Pulley and hanging mass misplaced}} &
\abnotes{\abck{Correct incline with block $m_1$}\abck{Edge pulley \& hanging $m_2$}\abck{Angle $\theta$ labelled at the base}}
\end{tabular}

\medskip
{\abchip{Case 2}\hspace{0.6em}\small An $8.00$~kg block on a horizontal table, connected by a rope over a pulley at the table's edge to a $6.00$~kg block hanging vertically.\par}
\smallskip
\begin{tabular}{@{}c@{\hspace{0.03\linewidth}}c@{}}
\abtag{abred}{Without SP-CoT} & \abtag{abteal}{With SP-CoT} \\[2pt]
\abframe{abred}{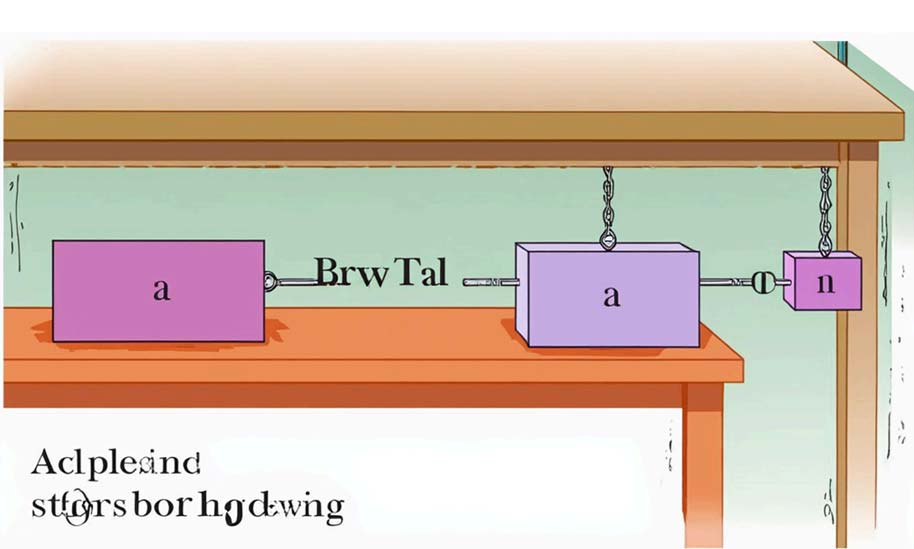} & \abframe{abteal}{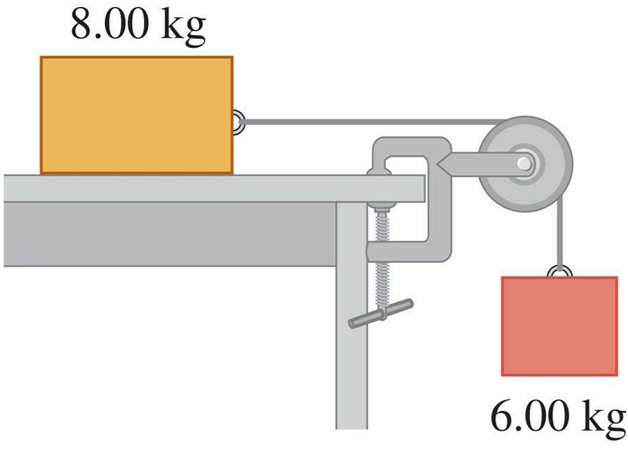} \\[3pt]
\abnotes{\abx{Meaningless labels ``Brw Tal'', ``a'', ``n''}\abx{Numeric masses absent}\abx{Pulley/hanging topology wrong}} &
\abnotes{\abck{$8.00$~kg block on the table}\abck{Edge-mounted pulley}\abck{$6.00$~kg mass hanging, masses correct}}
\end{tabular}

\medskip
{\abchip{Case 3}\hspace{0.6em}\small Two blocks~1 and~2 on a horizontal surface: block~1 moves toward block~2 with initial velocity $v_1$, while block~2 connects through a spring to a fixed wall.\par}
\smallskip
\begin{tabular}{@{}c@{\hspace{0.03\linewidth}}c@{}}
\abtag{abred}{Without SP-CoT} & \abtag{abteal}{With SP-CoT} \\[2pt]
\abframe{abred}{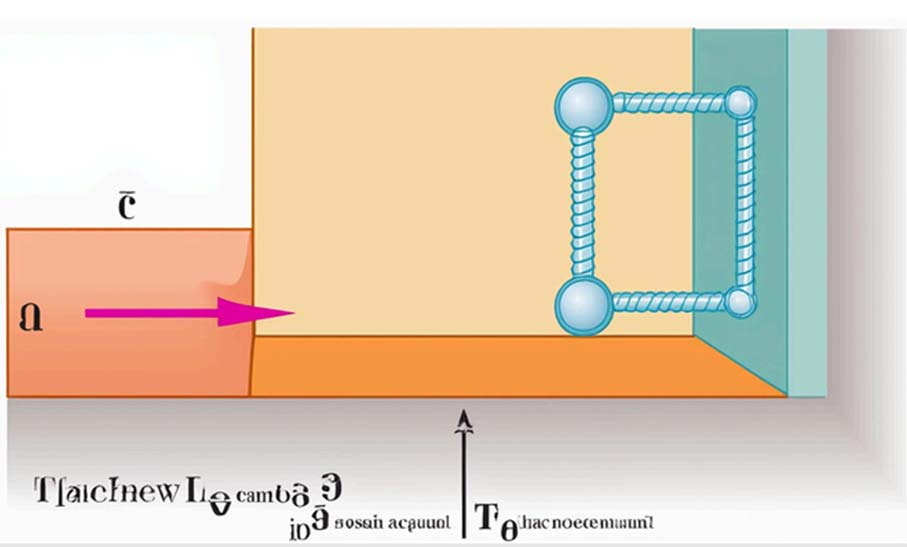} & \abframe{abteal}{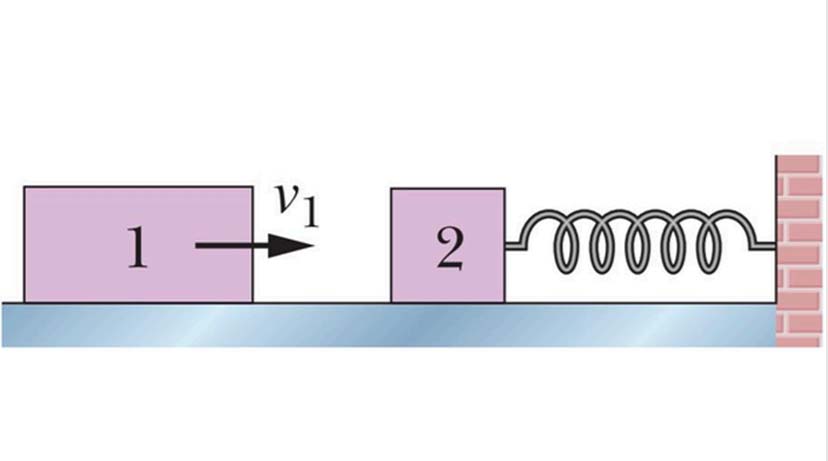} \\[3pt]
\abnotes{\abx{Layout corrupted into an L-shaped wall}\abx{Spring drawn as a detached loop}\abx{Blocks 1/2 and $v_1$ missing, captions nonsense}} &
\abnotes{\abck{Block~1 with velocity arrow $v_1$}\abck{Approaches block~2}\abck{Spring links block~2 to the fixed wall}}
\end{tabular}
\caption{\textbf{Ablation of the structured annotation.} Without the structured annotation the model produces figures that look plausible but whose geometry and topology are wrong and whose on-figure symbols are illegible, with it, the same architecture generates the entities, relations and values the prompt specifies. Checks and crosses list the per-panel observations.}
\label{fig:qual-ablation}
\end{figure}

\end{document}